\documentclass[journal]{IEEEtran}
\usepackage{booktabs}
\usepackage{amsfonts}
\usepackage{tabularx}
\usepackage{siunitx} 
\usepackage[linesnumbered,ruled]{algorithm2e}
\usepackage{algpseudocode}
\usepackage{romannum}
\usepackage{multirow}
\usepackage{graphicx}
\usepackage{float} 
\usepackage{dblfloatfix} 
\usepackage{colortbl}
\usepackage{caption}
\usepackage{subcaption}

\usepackage[cmex10]{amsmath}
\usepackage{array}

\usepackage{color}

\newcolumntype{C}{>{\centering\arraybackslash}p{1cm}}

\begin{document}

%
\title{MIMA: MAPPER-Induced Manifold Alignment for Semi-Supervised Fusion of Optical Image and Polarimetric SAR Data}
%
%
%

\author{Jingliang~Hu,~\IEEEmembership{Student Member,~IEEE,} Danfeng~Hong,~\IEEEmembership{Student Member,~IEEE,} and~Xiao~Xiang~Zhu,~\IEEEmembership{Senior~Member,~IEEE}
\thanks{The authors are with the Remote Sensing Technology Institute (IMF), German Aerospace Center (DLR), 82234 Wessling, Germany, and also with Signal Processing in Earth Observation (SiPEO), Technical University of Munich (TUM), 80333 Munich, Germany (e-mails: jingliang.hu@dlr.de; danfeng.hong@dlr.de; xiaoxiang.zhu@dlr.de).}
\thanks{This work is jointly supported by the European Research Council (ERC) under the European Union's Horizon 2020 research and innovation programme (grant agreement No. [ERC-2016-StG-714087], Acronym: \textit{So2Sat}), Helmholtz Association under the framework of the Young Investigators Group ``SiPEO'' (VH-NG-1018, www.sipeo.bgu.tum.de), and the Bavarian Academy of Sciences and Humanities in the framework of Junges Kolleg.}
}

\makeatletter
\def\hlinew#1{%
  \noalign{\ifnum0=`}\fi\hrule \@height #1 \futurelet
   \reserved@a\@xhline}

\maketitle

\begin{abstract}
\textcolor{blue}{This is the pre-acceptance version, to read the final version please go to IEEE Transactions on Geoscience and Remote Sensing on IEEE Xplore.} Multi-modal data fusion has recently been shown promise in classification tasks in remote sensing. Optical data and radar data, two important yet intrinsically different data sources, are attracting more and more attention for potential data fusion. It is already widely known that, a machine learning based methodology often yields excellent performance. However, the methodology relies on a large training set, which is very expensive to achieve in remote sensing. The semi-supervised manifold alignment (SSMA), a multi-modal data fusion algorithm, has been designed to amplify the impact of an existing training set by linking labeled data to unlabeled data via unsupervised techniques. In this paper, we explore the potential of SSMA in fusing optical data and polarimetric SAR data, which are multi-sensory data sources. Furthermore, we propose a MAPPER-induced manifold alignment (MIMA) for semi-supervised fusion of multi-sensory data sources. Our proposed method unites SSMA with MAPPER, which is developed from the emerging topological data analysis (TDA) field. To our best knowledge, this is the first time that SSMA has been applied on fusing optical data and SAR data, and also the first time that TDA has been applied in remote sensing. The conventional SSMA derives a topological structure using k-nearest-neighbor (\textit{kNN}), while MIMA employs MAPPER, which considers the field knowledge and derives a novel topological structure through the spectral clustering in a data-driven fashion. Experiment results on data fusion with respect to land cover land use classification and local climate zone classification suggest superior performance of MIMA.

\end{abstract}

\begin{IEEEkeywords}
Hyperspectral image, MAPPER, multi-modal data fusion, multi-sensory data fusion, multispectral image, PolSAR, semi-supervised manifold alignment (SSMA), topological data analysis (TDA).
\end{IEEEkeywords}

%
\IEEEpeerreviewmaketitle

\section{Introduction}
\IEEEPARstart{I}{n} recent decades, data fusion has attracted a lot of attention in the remote sensing community~\cite{zhang2010multi,dalla2015challenges,gomez2015multimodal,liu2019stfnet}, motivated by the simple fact that multiple data sources reveal complementary physical properties of observed scenes. For example, optical RGB data normally possesses high spatial resolution \cite{wu2019orsim}, while multi/hyperspectral data contains spectral information \cite{hong2018sulora}, and synthetic aperture radar (SAR) data gives dialectic and geometric properties. Thus, it is valuable to develop algorithms that are able to take advantage of different data sources for applications. In this regard, machine learning techniques are becoming increasingly important due to their excellent performance \cite{belgiu2016random,yokoya2017multimodal,hong2019augmented}. As is generally known in machine learning, the training data set is of great importance \cite{goodfellow2016deep}. Most successful techniques require a large set of training data \cite{zhou2017brief}. However, accessing a large training data set is very expensive, especially in remote sensing, because labeling a training data set in this field requires expertise that is more complicated than identifying dogs and cats. Therefore, a semi-supervised learning technique is a good option for remote sensing tasks, as the unlabeled data set is linked to the training data set by unsupervised approaches in the learning of the technique. It amplifies the effect of the existing training data set. Considering the importance of data fusion and precious training data, this paper studies a semi-supervised learning technique, named manifold alignment, to fuse optical image and polarimetric SAR (PolSAR) data for the purpose of classification.\\

\subsection{Fusion of optical and SAR data}
Due to the rapid development of remote sensing missions such as LandSat-8, Sentinel-2, EnMAP for optical remote sensing and TerreSAR-X, Tandem-X, Sentinel-1 for radar remote sensing, a huge amount of optical data and SAR data have been collected; the data volume can be expected to increase over time. The fusion of the two data sets hold great potential for use in various applications \cite{hu2019comparative}. Besides the data availability, the other reason to fuse them is that dialectic and geometric properties provided by SAR data are complementary to the spectral information of optical data. However, fusing them in practice is not as straightforward as the argument for doing it. The difficulty lies in the intrinsic differences in their imaging geometry. Because of the slanted looking angle of the SAR sensor, SAR images have an oblique appearance, with distortions of foreshortening, shadowing, and layover. This results in image geometry that is severely dissimilar to the nadir looking optical data. The extent of SAR distortions is positively correlated to height. This will pose substantial challenges when fusing these two data sets, especially in urban areas with large height fluctuations. To date, some studies have explored fusing these two data sources. We categorize those studies into three types based on their purposes: (1) registration oriented, (2) detection oriented, and (3) classification oriented.
\renewcommand{\@IEEEsectpunct}{\ \,}

\subsubsection{Registration} is actually a prerequisite of any further fusion. However, precise registration of SAR and optical image is very challenging due to geometric differences. A conditional generative adversarial network \cite{ninaCGAN} was trained to generate an artificial SAR image given a real-world optical image, and the optical image and SAR data are then registered by matching the artificial SAR data with the real-world SAR data. This technique was shown to be effective in a suburban area. A 3D registration is introduced in \cite{auer2017simulation} to align optical and SAR data by imitating the physical procedure of optical and SAR imaging based on a digital surface model. A pseudo-siamese convolutional neural network architecture \cite{hughes2018identifying} was trained to identify corresponding optical and SAR data in image patches and showed promising preliminary results. By far, although progress has been made in recent years, precise SAR and optical data registration has not achieved a robust solution yet, especially for complex urban areas. Thus, for other purposes of fusing these two data sources, the straightforward approach is registration by geographic coordinates. 
\subsubsection{Detection} tasks have been proven successful by using optical and SAR data for the purpose of detecting building outlines \cite{tupin2003detection}, crops \cite{campos2017exploitation}, water \cite{irwin2017fusion,lamarche2017compilation} and urban areas \cite{urbanDetect}. Since the detection task focuses on specific targets, studies extract representation of those targets from each of the two data sources so that they can work together to identify targets. For example, for detecting crops, optical data provides spectral signatures and SAR provides scattering mechanisms of interested targets; these characteristics are extracted and used together to identify the target under detection.
\subsubsection{Classification} is more challenging than detection tasks for more than one class of interest is under consideration. This paper focuses on this challenges \cite{hang2019cascaded}. Recently, a number of studies \cite{dimov2017sar,gaetano2017fusion,laurin2013optical,zhu2012assessment} have tried to solve classification tasks by using both optical and SAR data. In general, these fusing strategies all extracts features from the individual data set, then concatenate all the features and feed them into various classifiers. The most important part of this procedure is to extract hand-crafted informative features \cite{hong2014improved} regarding classification. A two-stream convolutional neural network (CNN) \cite{hu2017fusionet} derives high level features of individual data sets by utilizing the power of CNN and then concatenates those features for classification. In brief, concatenation is the main strategy for fusing SAR and optical data so far, which is an effective and straightforward approach.
\subsection{Semi-supervised manifold alignment (SSMA)}
SSMA pursues a projection for each input data source, and maps corresponding data into a shared latent space \cite{ham2005semisupervised,wang2009general}. These properties hold within this space: \textbf{\textit{(a)}} data of the same classes locate close to each other; \textbf{\textit{(b)}} data of different classes locate far from each other; \textbf{\textit{(c)}} the topological property of individual data is preserved. These three properties make SSMA to be a promising candidate for our task from a methodological perspective for two reasons. First, the first two properties promote classification-wise advantageous information from any data source to be used. Second, the final property implicitly connects unlabeled data to training data, which amplifies the functionality of the training data. These two factors meet our need for an algorithm that fuses data sets with the maximum usage of the training data. 

In the remote sensing community, SSMA has been investigated for various applications. It was applied to fuse an RGB image and hyperspectral image so that visualization of hyperspectral image could be achieved in the latent space, exhibiting more spectral information than conventional visualization methods in \cite{liao2016manifold}. A kernel manifold alignment was introduced in \cite{tuia2016multi} to fuse multiple optical remote sensing data into a latent space by nonlinear projections for a classification task. Manifold alignment was also used in \cite{volpi2015spectral} to align spectral signatures from different optical data sets by projecting them into a latent space so that object detection was achieved.

In regard to remote sensing data fusion, different data sources observe the same region of interest. Essentially, the observed target is a single object that appears differently in data sources due to sensor specifications. Thus, this question arises. Although, theoretically, SSMA is a good choice, does one latent space of observed objects, where data sources can be aligned? If it exists, can we find that space by using SSMA? Tuia et al. \cite{tuia2014semisupervised} applied SSMA to find the underlying space of multiple optical data sets under three scenarios: different looking angles, multi-temporal, and different sensors. In this work, we aim to fuse multi-sensory data sets, namely optical image and polarimetric SAR data, by SSMA. 

\subsection{Topology and MAPPER}
One important feature of SSMA resides its exploration of the topological structure of data. The conventional manifold based method~\cite{ham2005semisupervised,wang2009general,hong2017DR,he2004locality,roweis2000nonlinear} approximates topological properties by using the \textit{kNN}. They essentially assume that the underlying manifold of a data is a Riemannian manifold which can be locally approximated by Euclidean measurement~\cite{hatcher2005algebraic,lin2008riemannian,hong2018joint}. Recently, topological data analysis (TDA) has emerged as a new mathematical sub-field of big data analysis, by means of studying topological properties in the data~\cite{7472911,chazal2017introduction,zomorodian2005computing,edelsbrunner2000topological}. One TDA tool, named MAPPER, resolves a computable approximation of the Reeb graph which represents the topological structure of a data with respect to one interested intrinsic property of the data~\cite{singh2007topological,carriere2018statistical}.

A general explanation of topology is that it is an art of simplification. It ignores complex information of the object under studying, and rather focuses on one meaningful aspect of it. On this regard, conventional manifold methods focus on the aspect of the local connection or the local structure. On the side of MAPPER (Reeb graph), it focuses on the topological structure of data related to the interested intrinsic property. 

In real applications, the MAPPER has been proven capable of revealing unknown knowledge in medical studies, by interpreting topological structures of data sets. The tool was applied to analyze breast cancer transcriptional data and uncovered a sub-group of Estrogen Receptor-positive (ER+) breast cancers. Patients suffering this kind of cancer exhibit 100\% survival and no metastasis. This finding was previously unknown and is invaluable for future treatment \cite{nicolau2011topology}. MAPPER was also applied to analyze the data of preclinical traumatic brain injury (TBI) and spinal cord injury (SCI). It revealed a previously unknown pattern of co-occurring TBI and SCI, as well as a previously unknown harmful effect of an experimental drug treatment  \cite{nielson2015topological}. With the help of MAPPER, Li et al. \cite{li2015identification} explored complex medical records of type 2 diabetes (T2D) patients and revealed previously unknown sub-groups within T2D. All the above discoveries are invaluable and contribute to greater precision in the practice of medicine.

Besides the inspiration of these successful studies in medicine and the sound theoretical foundation of the MAPPER \cite{singh2007topological}, the other reasons which motivated the authors to utilize the MAPPER to explore the topological structure of the remote sensing data are listed as following:

\subsubsection{The field knowledge}
 The MAPPER focuses on the topological structure of data related to an intrinsic property. In practice, the intrinsic property is quantitatively derived from the data by an expert-designed filter function. The quantified property operates as a lens through which the MAPPER observes the data and extracts the topological structure of the data. Therefore, the choice of the lens, equally the filter function, introduces a field knowledge into the procedure of extracting topological structure. To our best knowledge, the ability of extracting topological structure from a field-knowledge perspective is unprecedented for the manifold-related technique in remote sensing.
\subsubsection{The regional-to-global topological structure}
  Instead of focusing on local structures of data points in conventional manifold-based techniques, the MAPPER focuses on an intrinsic property introduced by the filter function. Under the guidance of the filtered values, MAPPER divides a data into several bins, derives topological structure of each bin, and collects those structures together as a global one. This results in a regional-to-global topological structure. For the complex remote sensing data, especially SAR data, the regional derived structure is more robust to outliers than the local derived one.
\subsubsection{The data-driven and optimized topology }
  The spectral clustering is embedded into MAPPER in this work, leading to a data-driven and optimized topological structure.
  \begin{itemize}
  \item A data-driven topology.
  The eigen-gap concept in the spectral clustering detects the number of clusters~\cite{hong2016k}. This ensures the derived topological structure suits the distribution of the data. Rather, conventional techniques derive the topological structure of the whole data set, with the \textit{kNN} of a fixed $k$ \cite{hong2015novel}.
  
  \item An optimized topology.
  The spectral clustering is an optimized graph-cut algorithm, which is capable of unbiased grouping \cite{ng2002spectral,shi2000normalized}. However, a conventional manifold technique directly relies on the precision of the similarity measurement. Although sophisticated similarity measurements have been developed in remote sensing, the high dimensionality and the complexity of the data still pose challenges on the measurement.
\end{itemize}

\subsection{Summary}
The contributions of this paper are three-fold.
\begin{itemize}
  \item This work studies the fusion of heterogeneous remote sensing data sources, namely, the optical data and the polarimetric SAR data, with the semi-supervised manifold alignment technology.
  \item To our best knowledge, this is the first time that the topological data analysis (TDA) technique has been applied in the remote sensing community.
  \item A novel MAPPER-induced manifold alignment is proposed for semi-supervised data fusion. Its performance on the fusion of polarimetric data and optical data regarding classifications is quantitatively analyzed.
\end{itemize}

The remainder of this paper is organized as follows. In Section \Romannum{2}, MAPPER and SSMA are reviewed, and MIMA is introduced. The experimental setup, results, and comparisons are provided in Section \Romannum{3}. Finally, Section \Romannum{4} provides conclusions and remarks on the work.

\section{Methodology}
In this section, we first introduce the background of the topological data analysis tool called MAPPER. Then, we review the basics of semi-supervised manifold alignment. Finally, the novel MIMA is introduced.

\subsection{MAPPER}
\label{sec:mapper}
\begin{figure}
\includegraphics[width=\linewidth]{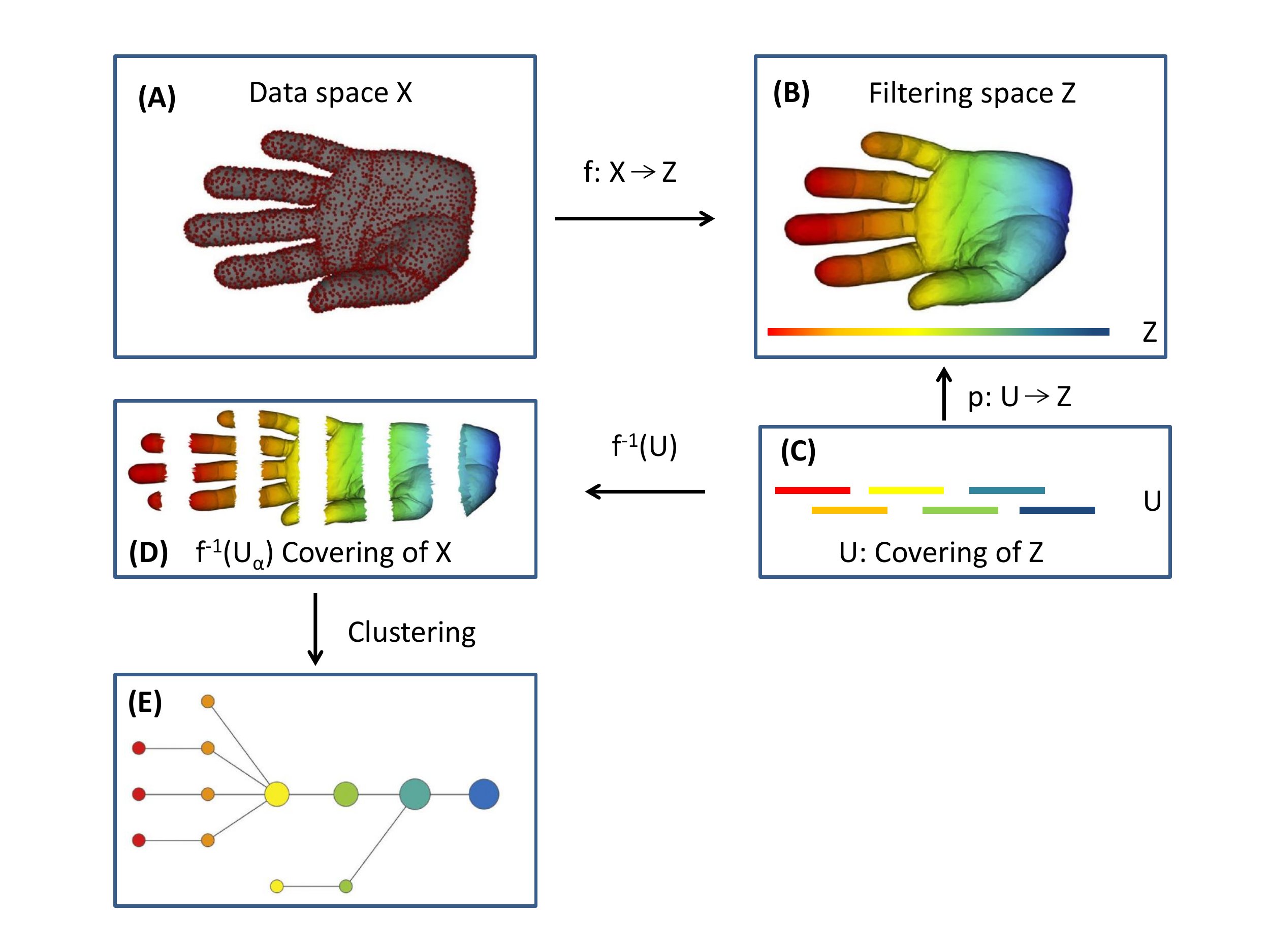} 
\caption{Example of MAPPER approach to derive the topological structure of the point cloud of a human hand. (A): Data space $X$, point cloud data of a human hand; (B): Filtered space $Z$, points colorized by the filter value; Filter function $f$: assigning data points with their horizontal distances to the right end; (C): U covering of Z, overlapped intervals of the filtered value; (D): $f^{-1}(U_{\alpha})$ covering of $X$, separating original data into bins according to intervals in (C), data in bins remain their original dimension; (E) $f^{-1}(U_{\alpha})$ covering of $X$, achieved by clustering bins of data. Modified from \cite{lum2013extracting}.}
\label{fig:mapper_concept}
\end{figure}

In order to introduce MAPPER \cite{singh2007topological} in a comprehensive and understandable way, we first provide an intuitive example, shown in Fig.~\ref{fig:mapper_concept}. The theoretical foundation of MAPPER is then introduced from the perspective of applied topology. Due to heavy reliance on mathematical concepts for this paper, we note that notations in section~\ref{sec:mapper} represent separate meaning from the other notations in the rest of the paper.

\subsubsection{Intuitive explanation}
MAPPER is a mathematical tool developed from applied topology to analyze and visualize big data sets \cite{singh2007topological}. The algorithm essentially consists of three components:

\begin{itemize}
  \item Filter function selection. MAPPER first requires a filter function which derives a filtering space where the interested intrinsic property is quantified. The chosen filter function should reveal physical meaning or geometric property of the data. It allows a specialist to introduce field knowledge into data analysis. For the example shown in Fig.~\ref{fig:mapper_concept}, the filter function is chosen as the distance to the wrist, so that a filtering space in Fig.~\ref{fig:mapper_concept} (B) is derived from the data point cloud.
    
  \item Data separation. In the filtering space, the continuous value range is sliced into overlapped intervals with a given overlap percentage and number of intervals, shown in Fig.~\ref{fig:mapper_concept} (C). Guided by the overlapped intervals \cite{hong2016robust}, the original input data can be separated into overlapped data bins accordingly, as shown in Fig.~\ref{fig:mapper_concept} (D). The separated data in bins have the same dimension as the original data.
  
  \item Clustering and visualization construction. Clustering is applied on each data bin. Clusters of adjacent data bins might include common data points. MAPPER constructs a graph where a node represents a cluster, and an edge represents a link of two clusters. The link is generated for two clusters if they share common data points. Therefore, the graph serves as a simplified visualization of the topological structure of a data set. For example, the graph in Fig.~\ref{fig:mapper_concept} (E) is derived by MAPPER to represent the topological structure of the point cloud data of a human hand. 
  
\end{itemize}

It is worth to highlight that the filter function is not seen as a dimension reduction, but quantifies a filtered space which guides the separation of the original data. As mentioned above, topology is an art of simplification. The conventional manifold learning focuses on the local structure of individual data points. On the other hand, MAPPER derives the topological structure of data while focusing on the property quantified by the filter function.

\subsubsection{Theoretical foundation} 
First, it is necessary to introduce the concept of covering in topology. In \cite{munkres2000topology}, it is explained as: let $p: U\,\to\,Z$ be continuous and surjective. If every point $z$ of $Z$ has a neighborhood $C$ that is evenly covered by $p$, then $p$ is called a covering map, and $U$ is defined to be a covering space of $Z$, then $p$ is a local homeomorphism of $U$ with $Z$. It means that, in terms of function $p$, the preimage in $U$ and the image in $Z$ share the same topological properties locally. 

The rest of the theoretical foundation is introduced in blocks corresponding to the three components of the MAPPER.

\begin{itemize}
  \item Filter function selection. According to MAPPER \cite{singh2007topological}, data is situated in a topological space $X$ as illustrated in Fig.~\ref{fig:mapper_concept} (A). A continuous function $f: X\,\to\,Z$ projects space $X$ to another space $Z$, as shown in Fig.~\ref{fig:mapper_concept} (B).
    
  \item Data separation. The space $Z$ is equipped with a covering space $U$, as shown in Fig.~\ref{fig:mapper_concept} (C). Assuming that covering space $U$ is a k-simplex spanned by a set $\{\alpha_{1},\alpha_{2},...,\alpha_{k}\}$ so that $U=\{U_{\alpha}\}$, since $f$ is continuous, $f^{-1}(U_{\alpha})$ forms a covering of space $X$ and could be used to represent topological space $X$ of given data , as shown in Fig.~\ref{fig:mapper_concept} (D).
    
  \item Clustering and visualization construction. The set $\{\alpha_{1},\alpha_{2},...,\alpha_{k}\}$, as the vertices of k-simplex, are k connected components in topological space $X$ which can be achieved by clustering. Thus, $f^{-1}(U_{\alpha})$ is achieved to represent data space $X$, as shown in Fig~\ref{fig:mapper_concept} (E).
  
\end{itemize}

\subsection{Semi-supervised manifold alignment (SSMA)}
Let $\mathbf{X}_{i}=[\mathbf{x}_{i}^{1},..., \mathbf{x}_{i}^{k},...,\mathbf{x}_{i}^{n_{i}}]\in \mathbb{R}^{m_{i}\times n_{i}}$ be a matrix representing the $i^{th}$ data source, with $m_{i}$ dimensions by $n_{i}$ instances. The term $\mathbf{x}_{i}^{k}$ denotes the $k^{th}$ instance of the $i^{th}$ data source. Let $K$ denote the total number of data sources. SSMA learns a set of $K$ projections $\{f_{1},...,f_{K}\}$. The $i^{th}$ projection $f_{i}$ maps the $i^{th}$ data source $\mathbf{X}_{i}$ into the latent space, where all the $K$ data sources are aligned in terms of the three desired properties discussed in the Introduction. The properties are formulated by three matrices, called the similarity matrix, dissimilarity matrix, and topology matrix. More specifically, the similarity matrix (\ref{eq:smat}) is computed by labeled information to pursue property \textbf{\textit{(a)}}: the data of same class located close to each other.
\begin{equation}
\label{eq:smat}
W_{s} = 
\begin{pmatrix}
W_{s}^{1,1}	& ... & W_{s}^{1,K} \\
	...		& ... & ... 		\\
W_{s}^{K,1} & ... & W_{s}^{K,K} \\
\end{pmatrix}
\end{equation}
The dissimilarity matrix is formed as (\ref{eq:dmat}) to accomplish property \textbf{\textit{(b)}}: data of different classes located far from one another.
\begin{equation}
\label{eq:dmat}
W_{d} = 
\begin{pmatrix}
W_{d}^{1,1}	& ... & W_{d}^{1,K} \\
	...		& ... & ... 		\\
W_{d}^{K,1} & ... & W_{d}^{K,K} \\
\end{pmatrix}
\end{equation}
The topology matrix (\ref{eq:tmat}) describes the topological structure of the data, which aims at the property \textbf{\textit{(c)}}: the topological property of individual data is preserved.
\begin{equation}
\label{eq:tmat}
W_{t} = 
\begin{pmatrix}
W_{t}^{1,1}	&  0  & 0 			\\
	 0 		& ... & 0 			\\
	 0 		&  0  & W_{t}^{K,K} \\
\end{pmatrix}
\end{equation}

Each of the matrices (\ref{eq:smat}), (\ref{eq:dmat}), and (\ref{eq:tmat}) is a matrix with the size $(n_{1}+n_{2}+...+n_{k})\times(n_{1}+n_{2}+...+n_{k})$. In each matrix, the $W^{i,j}$ is a matrix representing the relationship between the $i^{th}$ and $j^{th}$ data sources on the individual property.

Similarity matrix $W_{s}$ and dissimilarity matrix $W_{d}$ are generated based on label information. If $x_{i}^{p}$ and $x_{j}^{q}$ share a same label, then $W_{s}^{i,j}(p,q) = 1$, otherwise $W_{s}^{i,j}(p,q) = 0$. If $x_{i}^{p}$ and $x_{j}^{q}$ belong to different classes,  then $W_{d}^{i,j}(p,q) = 1$, otherwise $W_{s}^{i,j}(p,q) = 0$. 

Since the topological structure of the individual data set is preserved, the matrix $W_{t}$ is a block-wise diagonal matrix. The topological structure is conventionally given by the \textit{kNN}, which means $W_{t}^{i,i}(p,q) = 1$ if $x_{i}^{p}$ and $x_{i}^{q}$ are neighbors in a given \textit{kNN} neighborhood. Otherwise, $W_{t}^{i,i}(p,q) = 0$.

In order to simultaneously model the three properties of the latent space, three terms are formulated for the cost function:
\begin{equation}
\label{eq:cf_sterm}
A = \sum_{i=1}^{K}\sum_{j=1}^{K}\sum_{p=1}^{n_{i}}\sum_{q=1}^{n_{j}} \|f_{i}^{T}x_{i}^{p} - f_{j}^{T}x_{j}^{q}\|^{2}W_{s}^{i,j}(p,q).
\end{equation}
Minimizing Eq. (\ref{eq:cf_sterm}) has the effect of pulling data of the same class together in the latent space, which meets property \textbf{\textit{(a)}}.
\begin{equation}
\label{eq:cf_dterm}
B = \sum_{i=1}^{K}\sum_{j=1}^{K}\sum_{p=1}^{n_{i}}\sum_{q=1}^{n_{j}} \|f_{i}^{T}x_{i}^{p} - f_{j}^{T}x_{j}^{q}\|^{2}W_{d}^{i,j}(p,q).
\end{equation}
Maximizing Eq. (\ref{eq:cf_dterm}) tends to push data of different classes away, which is consistent with property \textbf{\textit{(b)}}.

\begin{equation}
\label{eq:cf_cterm}
C = \sum_{i=1}^{K} \sum_{p=1}^{n_{i}} \sum_{q=1}^{n_{i}} \|f_{i}^{T}x_{i}^{p} - f_{i}^{T}x_{i}^{q}\|^{2} W_{t}^{i,i}(p,q).
\end{equation}

\begin{figure*}[!h]
\centering
\includegraphics[height=0.7\linewidth]{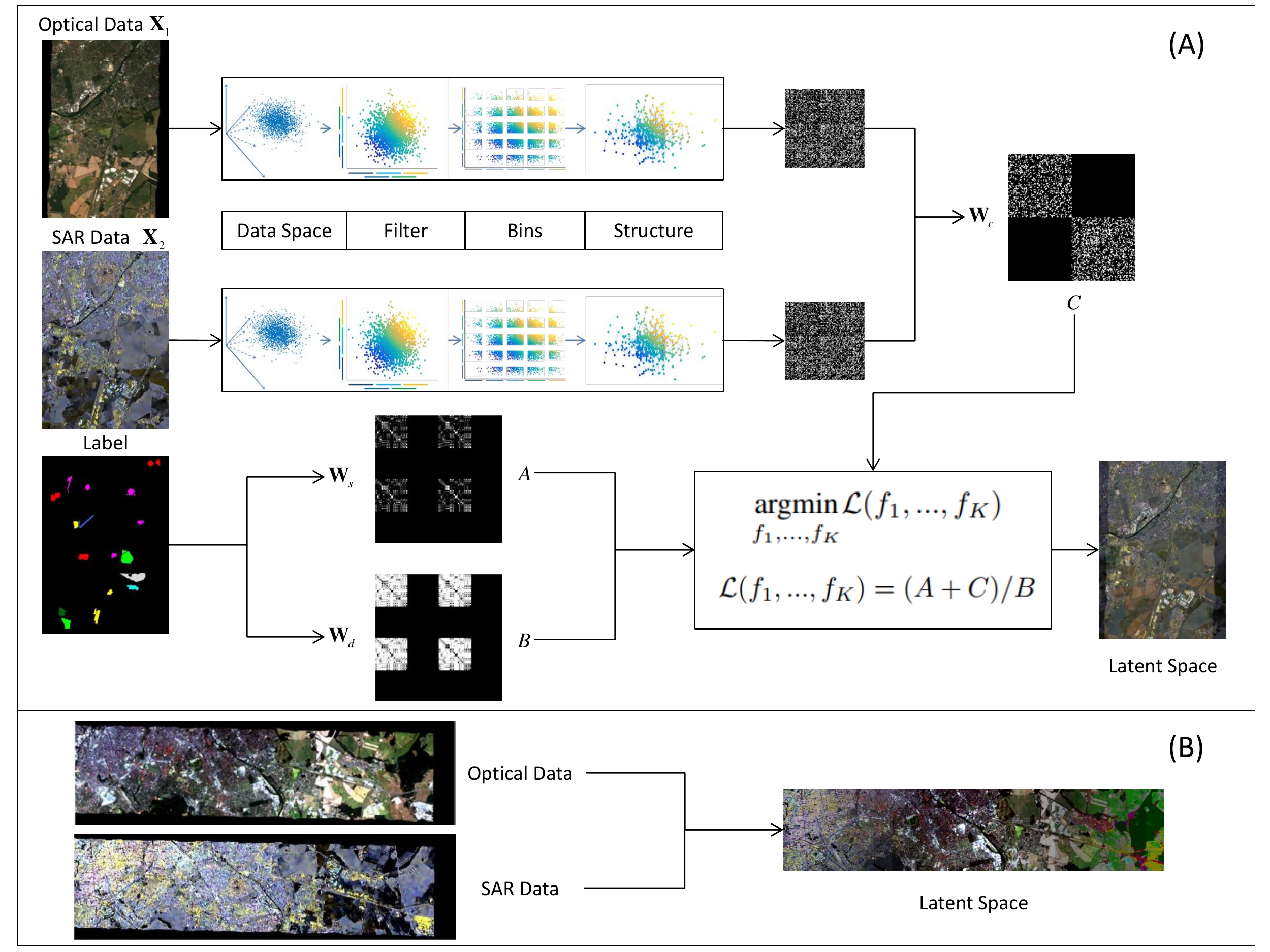} \\
\caption{The flowchart of the algorithm MIMA. (A) Training phase: a topological graph ($W_{c}$) is derived from the optical data and the SAR data by MAPPER. A similarity graph ($W_{s}$) and a dissimilarity graph ($W_{d}$) are formed by using the label information. Therefore, three regularization terms $A$, $B$, and $C$ are formulated as Eq.~\ref{eq:cf_sterm}, Eq.~\ref{eq:cf_dterm}, and Eq.~\ref{eq:cf_cterm}, respectively. Lastly, the projection to the latent space is learned by optimizing $\underset{f_{1},...,f_{K}}{\text{argmin}} \mathcal{L}(f_{1},...,f_{K})$, where $\mathcal{L}(f_{1},...,f_{K}) = (A+C)/B$. (B) Testing phase: the out-of-sample optical data and SAR data are projected into the latent space to accomplish fusion.} 
\label{fig:mapperssma}
\end{figure*}

Minimizing Eq. (\ref{eq:cf_cterm}) preserves the topological structure of individual data set, corresponding to property \textbf{\textit{(c)}}. Eqs. (\ref{eq:cf_sterm} -~\ref{eq:cf_cterm}) can be combined into the final cost function, which is formulated as (\ref{eq:cost_func}):
\begin{equation}
\label{eq:cost_func}
\mathcal{L}(f_{1},...,f_{K}) = (A+C)/B,
\end{equation}
and hence an optimization problem (\ref{eq:argmin}) can be written as
\begin{equation}
\label{eq:argmin}
\underset{f_{1},...,f_{K}}{\text{argmin}} \mathcal{L}(f_{1},...,f_{K})
\end{equation}


Proven in \cite{wang2009general}, the solution $f_{1},...,f_{K}$ that minimizing the cost function $\mathcal{L}(f_{1},...,f_{K})$ is given by the smallest non-zero eigenvectors of the generalized eigenvalue decomposition of (\ref{eq:GED}). And the matrix $D$ and the matrix $L$ in (\ref{eq:GED}) are the degree matrix and the Laplacian matrix, respectively.

\begin{equation}
\label{eq:GED}
Z(\mu L_{t}+ L_{s})Z^{T}x = \lambda ZL_{d}Z^{T}x,
\end{equation}
where \\

$Z = 
\begin{pmatrix}
X_{1}	&  0  & ... & 0 	  \\
...	    & ... & ... & ...     \\
0       & ... &  0  & X_{K} \\
\end{pmatrix}$, \\

\vspace{0.4cm}
$L_{a} = W_{a} - D_{a}$, $a \in \{s, d, t\}$

\vspace{0.4cm}
$ D_{a}(p,q) = 
\begin{cases}
\sum_{q=1}^{m_{1}+...+m_{k}}W_{a}(p,q) & p = q\\
0 & p\neq q
\end{cases}
$.

\subsection[MAPPER-induced manifold alignment for semi-supervised data fusion (MIMA)]{MAPPER-induced manifold alignment for semi-supervised\\ \hspace*{0.4 cm} data fusion (MIMA)}
\label{sec:MIMAtheory}

\begin{algorithm}
\caption{\textbf{MAPPER}($\mathbf{X}_{i}$,$\mathbf{b}$,$\mathbf{c}$,$\mathbf{F}$)}
\label{al:mapper}
\KwIn{$\mathbf{X}_{i}\in \mathbb{R}^{m_{i}\times n_{i}}$: the $i^{th}$ data source with $n_{i}$ instances and $m_{i}$ dimensions,
\hspace{0.2cm} $\mathbf{b}$: the number of bins,
\hspace{0.2cm} $\mathbf{c}$: overlapping percentage of adjacent bins,
\hspace{0.2cm} $\mathbf{F}$: filter function.}
\KwOut{ $\mathbf{W}_{c}^{i,i}$: adjacent matrix with the size of $n_{i}\times n_{i}$.}
\vspace{0.2cm}
calculate the parameter space $\mathbf{X}_{i} \mathbf{F}$ \\

divide $\mathbf{X}_{i} \mathbf{F}$ into $\mathbf{b}$ intervals with $\mathbf{c}\%$ overlap of adjacent intervals\\

divide data $\mathbf{X}_{i}$ into $\mathbf{b}$ data bins corresponding to intervals achieved in 2\\

\textbf{for} (each data bin):\\
\hspace{0.2cm} Spectral clustering\\
\textbf{end for}\\

Construct topological matrix
     $\mathbf{W}_{c}^{i,i}(p,q)=
      \begin{cases}
        1, & \text{if $p$ and $q$ in the same cluster;}\\
        1, & \text{if $p$ and $q$ in the linked clusters;}\\
        0, & \text{otherwise.}
      \end{cases}$\\
\vspace{0.2cm}
\textbf{Return} $\mathbf{W}_{c}^{i,i}$
\end{algorithm}

\begin{algorithm}
\caption{MIMA (\{$\mathbf{X}_{i}$,$\mathbf{Y}_i$\}, $\mathbf{b}$, $\mathbf{c}$, $\mathbf{F}$)}
\label{al:mapperssma}
\KwIn{ \{$\mathbf{X}_{i}$,$\mathbf{Y}_i$\}, $i\in \{1,...,K\}$: $K$ data sources and label,
\hspace{0.2cm} $\mathbf{b}$: the number of bins,
\hspace{0.2cm} $\mathbf{c}$: overlapping percentage of adjacent bins,
\hspace{0.2cm} $\mathbf{F}$: filter function.}
\KwOut{ $\hat{\mathbf{X}}_{i}$, $i\in \{1,...,K\}$: the learned latent features of the $K$ data sources.}
\vspace{0.2cm}
Construct $\mathbf{W}_s$ by labeled data $\mathbf{Y}_i$, $i\in \{1,...,K\}$;\\

Construct $\mathbf{W}_d$ by labeled data $\mathbf{Y}_i$, $i\in \{1,...,K\}$;\\

\textbf{for} (i = 1:K):\\

\hspace{0.4cm} $\mathbf{W}_{c}^{i,i}=\textbf{MAPPE}(\mathbf{X}_i, \mathbf{b}, \mathbf{c}, \mathbf{F})$\\

\textbf{end for}\\

Construct $\mathbf{W}_{c} = 
\begin{pmatrix}
W_{c}^{1,1}	&  0  & 0 			\\
	 0 		& ... & 0 			\\
	 0 		&  0  & W_{c}^{K,K} \\
\end{pmatrix}$\\

Compute the projections $\{f_{1},...,f_{K}\}$ by solving Eq (\ref{eq:GED})\\

\textbf{for} (i = 1:K):\\

\hspace{0.4cm} $\hat{\mathbf{X}}_{i} = \mathbf{X}_{i} f_{i}$ \\

\textbf{end for}\\

\vspace{0.2cm}
\textbf{Return} $\hat{\mathbf{X}}_{i}$, $i\in \{1,...,K\}$

\end{algorithm}

As introduced in the last section, three properties are pursued in SSMA while projections are being learned. Essentially, the first two properties seek to minimize intra-class variance and maximize inter-class variance for the projected data by using label information. This is a goal commonly pursued by many classification strategies, such as linear discriminant analysis \cite{mclachlan2004discriminant}. The third property, preserving topological structure, brings two powerful characteristics to SSMA. First, the topological structure is extracted from data, both with and without a label. Thus, SSMA builds up connections among them, which implicitly propagates the label information to unlabeled data. This would amplify the usage of existing labels. Since the label is valuable, the propagation property of the topological term is highly valued. Second, topology emphasizes a notion of nearness, but can distort or even ignore large distances \cite{singh2007topological}. This is a desirable property for the purpose of classification. For instance, data of one class located in a certain extent of feature space, and locations with large distance to the extent are meaningless for classifying the specific class. This is also proven truth in classification using topology \cite{roweis2000nonlinear,he2004locality}.

In order to achieve the topological term, \textit{kNN} commonly serves as the tool to approximate topological structure in conventional methods \cite{wang2009general, hong2016local}, chosen for its simplicity. In our proposed MIMA, we utilize MAPPER to extract topological structure. There are two reasons to do so. First, when applying MAPPER, field knowledge could be introduced by choosing the filtering function $\mathbf{F}$. In remote sensing classification, field knowledge is essential for the complicity of data. Second, when using \textit{kNN}, nearness is decided solely by the parameter K, which is manually given. Once the K is determined, it is applied to all data without any adaptation. However, nearness is achieved by clustering in MAPPER, which is a more robust approach than deciding nearness by giving a threshold value K. Furthermore, in order to empower MAPPER to decide the nearness in an adaptive manner, the original single-linkage clustering \cite{jain1988algorithms, singh2007topological} is replaced by the spectral clustering \cite{ng2002spectral} in MIMA. The reason is that the spectral clustering is able to detect the number of clusters by the concept of the eigen-gap \cite{ng2002spectral}. Thus, when clustering each data bin, the number of clusters is decided based on the data itself, meaning that the nearness is derived in a data-driven manner. For different data bins, the numbers of clusters are different, meaning the nearness is derived for different parts of data in adaptive fashion. Thus, our improved version of MAPPER is capable of deriving topological structure in an automatic and adaptive fashion.

\begin{figure*}[t]
\centering
\begin{tabular}{cccc}
\includegraphics[height=0.6\linewidth]{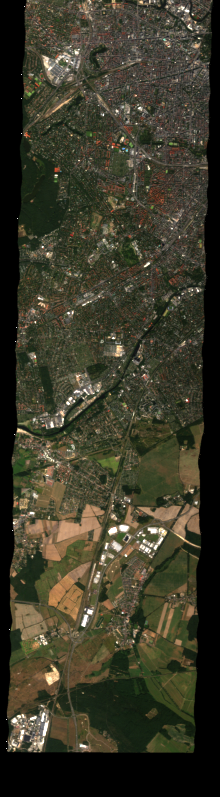}&
\includegraphics[height=0.6\linewidth]{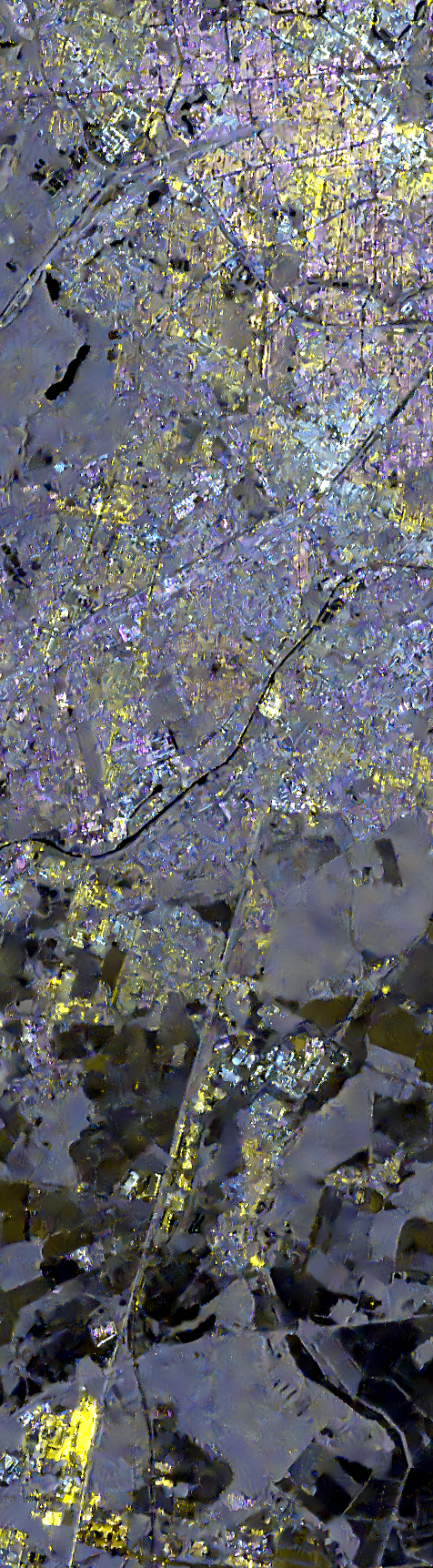} &
\includegraphics[height=0.6\linewidth]{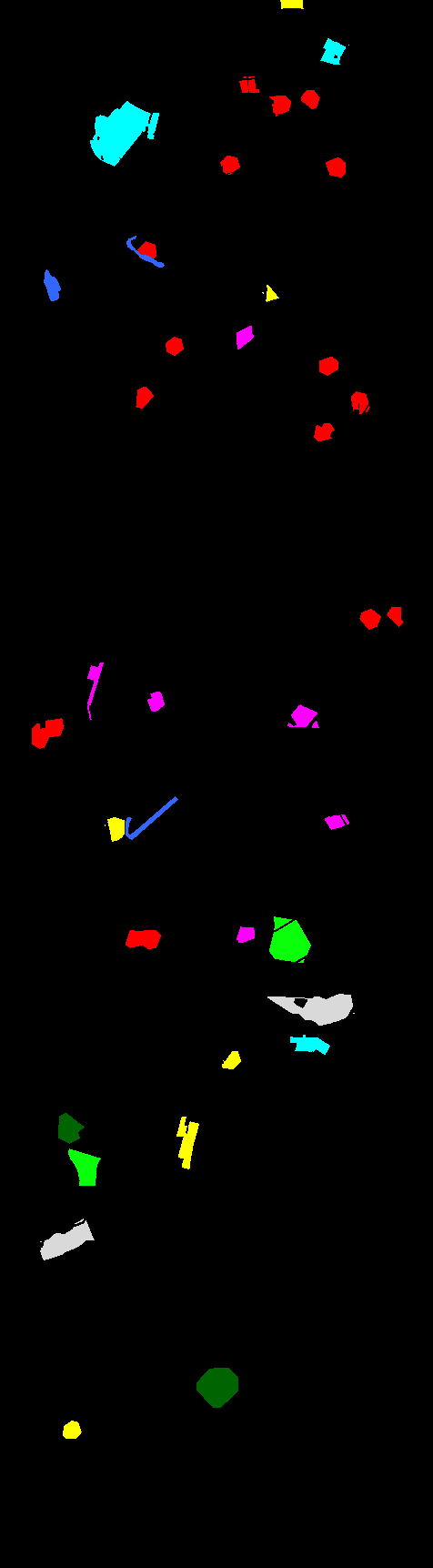} &
\includegraphics[height=0.6\linewidth]{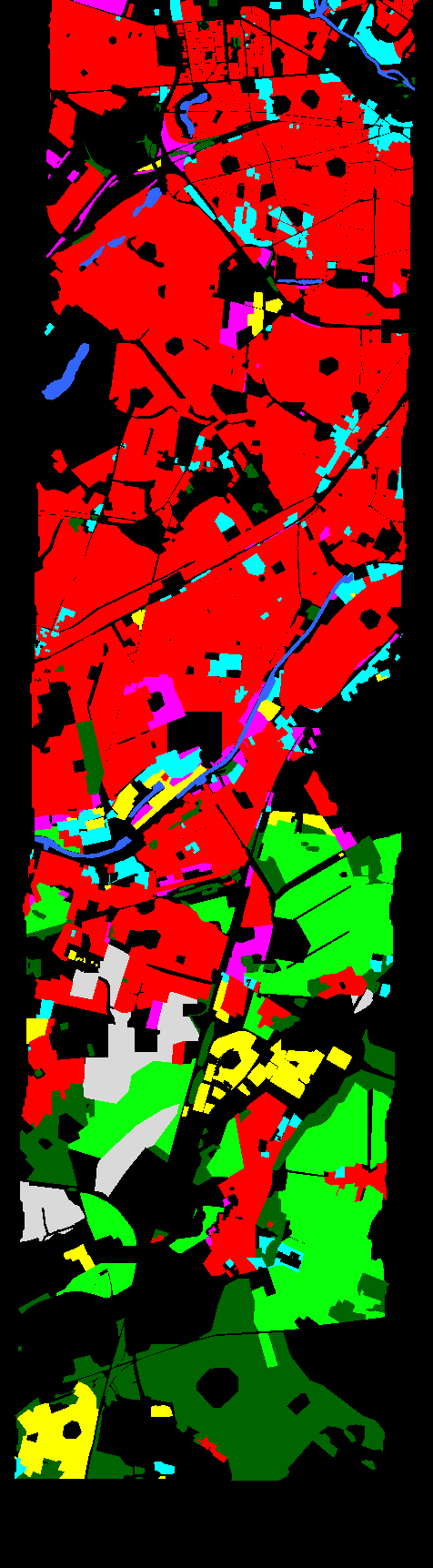} \\
\end{tabular}
\caption{LCLU data set. From left to right: RGB components of simulated EnMAP data; RGB component of Sentinel-1 dual-Pol data; LCLU training set; LCLU testing set.}
\label{fig:berlinLCLU}
\end{figure*}

Although the original goal of MAPPER is to provide a simplified visualization of a complicated data set, as shown in Fig~\ref{fig:mapper_concept}, one can also derive the comprehensive topological structure of all data points using MAPPER. The topological structure of data source $\mathbf{X}_{i}$ could be represented as an $n_{i}\times n_{i}$ matrix $W_{c}^{i}$, where $n_{i}$ is the number of instances: $W_{c}^{i}(p,q) = 1$, when data instances $p$ and $q$ are in the same cluster or in linked clusters, otherwise, $W_{c}^{i}(p,q) = 0$. In MIMA, the topological matrix $W_{t}$ in equation (\ref{eq:tmat}) is replaced by $W_{c}$ (\ref{eq:mappermat}). 

\begin{equation}
\label{eq:mappermat}
W_{c} = 
\begin{pmatrix}
W_{c}^{1,1}	&  0  & 0 			\\
	 0 		& ... & 0 			\\
	 0 		&  0  & W_{c}^{K,K} \\
\end{pmatrix}.
\end{equation}

The detailed steps of MIMA are summarized in Algorithm~\ref{al:mapper} and Algorithm~\ref{al:mapperssma}.

\section{Experiments and Discussion}

\subsection{Data set and feature design}
\subsubsection{Land cover land use data set (LCLU data set)} 
As shown in Fig.~\ref{fig:berlinLCLU}, the LCLU data set consists of three data sources: a hyperspectral image, dual-Pol SAR data, and ground truth data. The hyperspectral image is a simulated spaceborne EnMAP scene with a size of 817 by 220, a 30-meter ground sampling distance (GSD), and 244 spectral bands ranging from \SI{400}{\nm} to \SI{2500}{\nm} \cite{okujeni2016berlin}. The dual-Pol SAR data is a VH-VV polarized Sentinel-1 single look complex (SLC) data collected by interferometric wide swath mode.\footnote{https://sentinel.esa.int/web/sentinel/user-guides/sentinel-1-sar/acquisition-modes/interferometric-wide-swath} The Sentinel-1 SLC data is preprocessed by the ESA SNAP toolbox.\footnote{http://step.esa.int/main/toolboxes/snap/} The processed dual-Pol SAR data has a GSD of 13 meters and a size of 1723 by 476. It is organized as the commonly used PolSAR covariance matrix. The ground truth is a land cover land use data set derived from an Open Street Map data.\footnote{http://download.geofabrik.de/}

\subsubsection{Local climate zone data set (LCZ data set)}
\begin{figure*}[!t]
\centering
\begin{tabular}{cc}
\includegraphics[width=0.3\linewidth,height=0.3\linewidth]{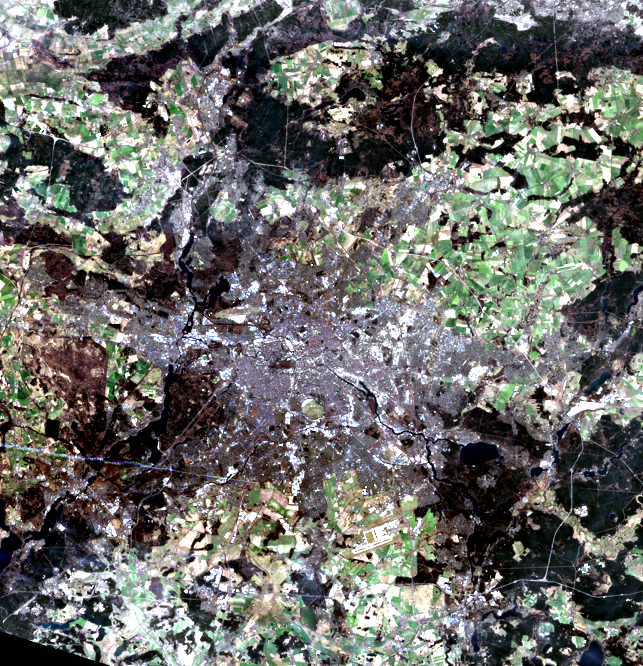} &
\includegraphics[width=0.3\linewidth,height=0.3\linewidth]{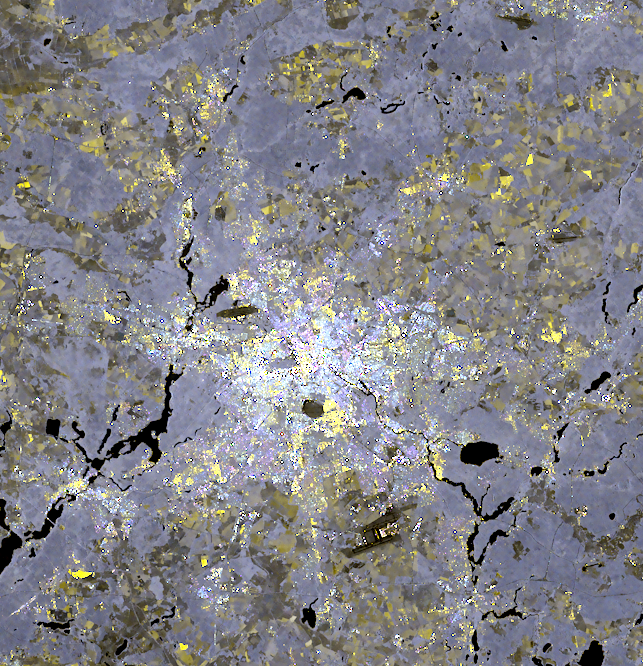} \\
\includegraphics[width=0.3\linewidth,height=0.3\linewidth]{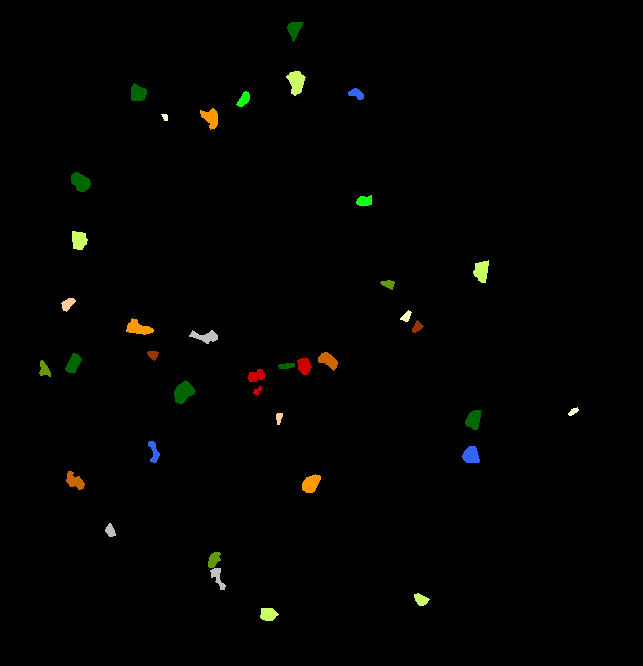} &
\includegraphics[width=0.3\linewidth,height=0.3\linewidth]{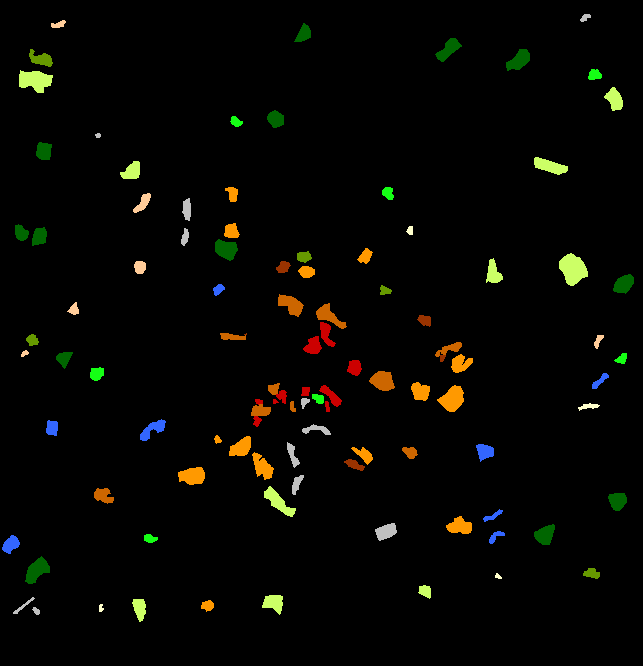} \\
\end{tabular}
\caption{LCZ data set. From left to right, first row: RGB components of LandSat-8 data; RGB component of Sentinel-1 dual-Pol data; second row: LCZ training set; LCZ testing set.}
\label{fig:berlinLCZ}
\end{figure*}

The local climate zone data set is demonstrated in Fig.~\ref{fig:berlinLCZ}. It consists of a multispectral image, a dual-Pol SAR data, and a ground truth data. The multispectral image is a scene of LandSat-8 data with a size of 2220 by 2143, a 30-meter GSD, and 11 bands. The dual-Pol SAR data is also a VV-VH polarized Sentinel-1 data processed by the ESA SNAP toolbox. It has a 13.9-meter GSD, a size of 4795 by 4632, and is organized as the commonly used PolSAR covariance matrix. The ground truth is a local climate zone label released by the IEEE GRSS IADF for the data fusion contest in 2017.\footnote{http://www.grss-ieee.org/}

\subsubsection{Label configuration} For both the LCLU data set and the LCZ data set, as shown in Fig.~\ref{fig:berlinLCLU} and Fig.~\ref{fig:berlinLCZ}, the training label and the testing label are block-wise separated so that the transferring ability of algorithms is under examination and the risk of implicitly including testing samples into the training procedure is avoided \cite{hansch2017correct}. The label information is detailed in Table~\ref{tb:labLCLU} and Table~\ref{tb:labLCZ}.

\subsubsection{Unlabeled data}
Regarding SSMA and MIMA, the training procedures involves both labeled data and unlabeled data. The unlabeled data was selected by the clustering strategy in \cite{tupin2003detection} so that cluster centers of unlabeled data were selected. In this work, for a more general case, the unlabeled data for training is randomly selected outside the extent of training set. For both the LCLU data set and the LCZ data set, 6000 unlabeled data instances are selected to be involved in training.

\begin{table}
\centering
\caption{Summary of training and testing for LCLU data set}
\label{tb:labLCLU}
\begin{tabular}{c|cc}
\toprule
\toprule
Class       & \multicolumn{2}{c}{Number of Samples}       \\
\midrule
 Name          &  Train Set  & Test Set \\
 \midrule
Forest           & 298            & 52455       \\
Residential area & 756            & 262903      \\
Industrial area  & 296            & 17462       \\
Low plants       & 344            & 56683       \\
Soil             & 428            & 14505       \\
Allotment        & 281            & 11322       \\
Commercial area  & 560            & 20909       \\
Water            & 153            & 5539        \\
\midrule
Total            & 3116           & 441778      \\
\bottomrule
\end{tabular}
\end{table}

\begin{table}
\centering
\caption{Summary of training and testing for LCZ data set}
\label{tb:labLCZ}
\begin{tabular}{c|cc}
\toprule
\toprule
Class       & \multicolumn{2}{c}{Number of Samples}       \\
\midrule
 Name         & Train Set  & Test Set \\
 \midrule
Compact mid-rise   & 198            & 1138        \\
Open high-rise     & 83             & 412         \\
Open mid-rise      & 213            & 2023        \\
Open low-rise      & 375            & 3260        \\
Large low-rise     & 233            & 1189        \\
Sparsely built     & 92             & 577         \\
Dense trees        & 769            & 3423        \\
Scattered trees    & 136            & 756         \\
Bush, scrub        & 181            & 689         \\
Low plants         & 560            & 3305        \\
Bare soil or sands & 84             & 192         \\
Water              & 246            & 1241        \\
\midrule
Total              & 3170           & 18205       \\
\bottomrule
\end{tabular}
\end{table}

\subsubsection{Feature design of the LCLU data set}
In order to conduct fair comparisons among algorithms, two principles are pursued on the design of input features of individual data sources. The first principle is simply that input features of each data source should be the same for all algorithms. The second principle is that, when an individual data source is used for classification, the input feature should enable reasonably good performance. This is to ensure that later improvements do not originate from the unexplored potential of one data source, but from the fusion or the fusion algorithms. For example, due to the well-known curse-of-dimensionality \cite{donoho2000high}, conducting classification on selected dimensions of hyperspectral images could result in better performance than using the data with all dimensions \cite{farrell2005impact}. If the original full dimensional data were used in our case, it would then be unclear later whether the improvement comes from the fusion or from the dimension reduction. 

Regarding the feature design of the simulated EnMAP data, the spectral-spatial feature concept is adopted by extracting morphological profiles from extracted informative sub-dimensions \cite{benediktsson2005classification,rasti2017hyperspectral}. Specifically, the first four principal components (PCs) are extracted, which accounts for 99\% of the variances of the simulated EnMAP data. The morphological profile is then extracted from these four PCs with radius equal to one, two, and three. In total, 28 features are extracted from the simulated EnMAP data set.

\begin{table*}[t]
\caption{The nine algorithms in experimental comparisons. Their fusion strategies are MA-fusion (manifold alignment fusion) and DR-fusion (joint dimension reduction fusion). The learning resource are the Label (annotated data-label records), the Pseudo-label (prediction from a classifier), and the data structure (the distribution of data in feature space). The parameters of these algorithms are, \textit{k}: the $k^{th}$ neighbor in \textit{kNN} for approximating topological structure; $dn$: the number of dimensions in the projected space; $\mu$: the importance weighting of topological structure; $\alpha$ and $\beta$: learning rates.} 
\centering
\label{tb:para}
\begin{tabular}{l|cc|ccccccc}
\toprule
\toprule
\multirow{2}{*}{Algorithm} & \multicolumn{2}{c|}{Principle} 			            & \multicolumn{5}{c}{Parameters} \\
                           &  Fusion Strategy       &  Learning Resource        & \textit{k}   &   $dn$       &     $\mu$    &  $\alpha$    &    $\beta$   &  $b$    &    $c$   \\
\midrule
(\textit{A}) POL           &  PolSAR only           &  Label 			        &      -	   &      -       &       -      &       -      &       -      &       -      &       -      \\
(\textit{B}) OPT           &  Optical only        	&  Label		            &      -	   &      -       &       -      &       -      &       -      &       -      &       -      \\
(\textit{C}) OPT-POL       &  Feature concatenation &  Label 			        &      -	   &      -       &       -      &       -      &       -      &       -      &       -      \\
(\textit{D}) COSPACE       &  MA-fusion	            &  Label                    &      -       &      -       &       -      & $\checkmark$ & $\checkmark$ &       -      &       -      \\
(\textit{E}) LeMA          &  MA-fusion	            &  Label \& Pseudo-label    &      -       &      -       &       -      & $\checkmark$ & $\checkmark$ &       -      &       -      \\
(\textit{F}) LPP           &  DR-fusion			    &  Data structure           & $\checkmark$ & $\checkmark$ &       -      &       -      &       -      &       -      &       -      \\
(\textit{G}) LPP-SE        &  DR-fusion			    &  Label \& Data structure  & $\checkmark$ & $\checkmark$ &       -      &       -      &       -      &       -      &       -      \\
(\textit{H}) SSMA          &  MA-fusion 	        &  Label \& Data structure  & $\checkmark$ & $\checkmark$ & $\checkmark$ &       -      &       -      &       -      &       -      \\
(\textit{I}) MIMA          &  MA-fusion 	        &  Label \& Data structure  &      -       & $\checkmark$ & $\checkmark$ &       -      &       -      &       $\checkmark$      &       $\checkmark$      \\
\bottomrule
\bottomrule
\end{tabular}
\end{table*}

Regarding the feature design of Sentinel-1 dual-Pol data, four polarimetric features are derived: intensity of the VH channel, intensity of the VV channel, the coherence of VV and VH, and the intensity ratio of VV and VH. Since the morphological profile was proven to promote classification of PolSAR \cite{zhu2012assessment,wurm2017slum,hu2018feature}, it is also used to extract spatial information from dual-Pol data here with radius equal to one, two, and three. In total, it results in 28 features from Sentinel-1 dual-Pol data.

\subsubsection{Feature design of the LCZ data set}
The feature design for the LCZ data set also follows principles described in the feature design of LCLU data set. 

Regarding the feature design of the LandSat-8 data, in order to achieve feature combination for reasonable good performance, the feature extraction and selection follows the strategy in first prize work from the GRSS IADF data fusion contest in 2017 \cite{yokoya2017multimodal}. Local statistical parameters (mean and standard deviations in a $100 \times 100$-meter neighborhood) and morphological profiles are extracted from original LandSat-8 data. For details, please refer to \cite{yokoya2017multimodal}. In total, 34 features are used in our work.

Regarding the feature design of Sentinel-1 dual-Pol data in the LCZ data set, the data source and the preprocessing are the same as those in the LCLU data set. The prepared fundamental features are the four polarimetric features. However, due to the local climate zone describes an urban local neighborhood at a grid with a $100 \times 100$-meter unit cell, feature extraction is different in the LCZ data set than in the in LCLU data set. Local statistical parameters, mean and standard deviation of local $100 \times 100$-meter cell, are derived from all four polarimetric features, resulting in eight features. Morphological profiles are therefore extracted from these eight features with radius equal to one and three. Thus, 40 features are prepared in total.

\subsection{Experiments setting}

In experiments, nine algorithms, which are listed in Table~\ref{tb:para}, are applied to extract features from optical and dual-Pol SAR data. Hereafter, three classifiers are used to test the performance of these algorithms in terms of classification accuracy. The seven algorithms are: (\textit{A}) dual-Pol SAR data ({POL}), (\textit{B}) optical data (OPT), (\textit{C}) fusing of optical and dual-Pol SAR data by feature concatenation ({OPT-POL}), (\textit{D}) fusing optical and dual-Pol SAR data by COSPACE~\cite{hong2019cospace}, (\textit{E}) fusing optical and dual-Pol SAR data by LeMA~\cite{hong2019learnable}, (\textit{F}) fusing optical and dual-Pol SAR data by unsupervised joint dimension reduction using locality preserving projection \cite{he2004locality} ({LPP}), (\textit{G}) fusing optical and dual-Pol SAR data by semi-supervised joint dimension reduction using locality preserving projection ({LPP-SE}), (\textit{H}) fusing optical and dual-Pol SAR data by {SSMA} \cite{tuia2014semisupervised}, and (\textit{I}) fusing optical and dual-Pol SAR data by the proposed {MIMA}.

Among these nine algorithms, parameter tuning is required by (\textit{D}) COSPACE, (\textit{E}) LeMA, (\textit{F}) LPP, (\textit{G}) LPP-SE, (\textit{H}) SSMA, and (\textit{I}) MIMA, as shown in Table~\ref{tb:para}. For the (\textit{D}) COSPACE and (\textit{E}) LeMA, two learning rates, $\alpha$ and $\beta$ need to be set for the optimization. They are tunned by searching in a grid of $\{10^{-1},10^{-2},10^{-3},10^{-4}\}$. Regarding the parameter \textit{k} in (\textit{F}) LPP, (\textit{G}) LPP-SE and (\textit{H}) SSMA, it has been reported in \cite{tuia2014semisupervised} that the parameter does not have significant influence on the result and is recommended to be nine. For the parameters $dn$ and $\mu$, they will be discussed later in our experiments. In MIMA, two more parameters have to be decided: (1) $\mathbf{b}:$ the number of intervals for dividing the data, and (2) $\mathbf{c}$: the overlapping rate. Since these two parameters have limited influences, as discussed in section \ref{sec:MAPPERpara}, specially when compared with the other parameters. Their values are chosen as 5 and 50\%, respectively, which is a result of consulting other studies \cite{nicolau2011topology,li2015identification} and summarizing practical experiences of the author.

\begin{figure*}[t]
	\centering	
    \begin{tabular}{ccc}
      \includegraphics[width=0.3\textwidth]{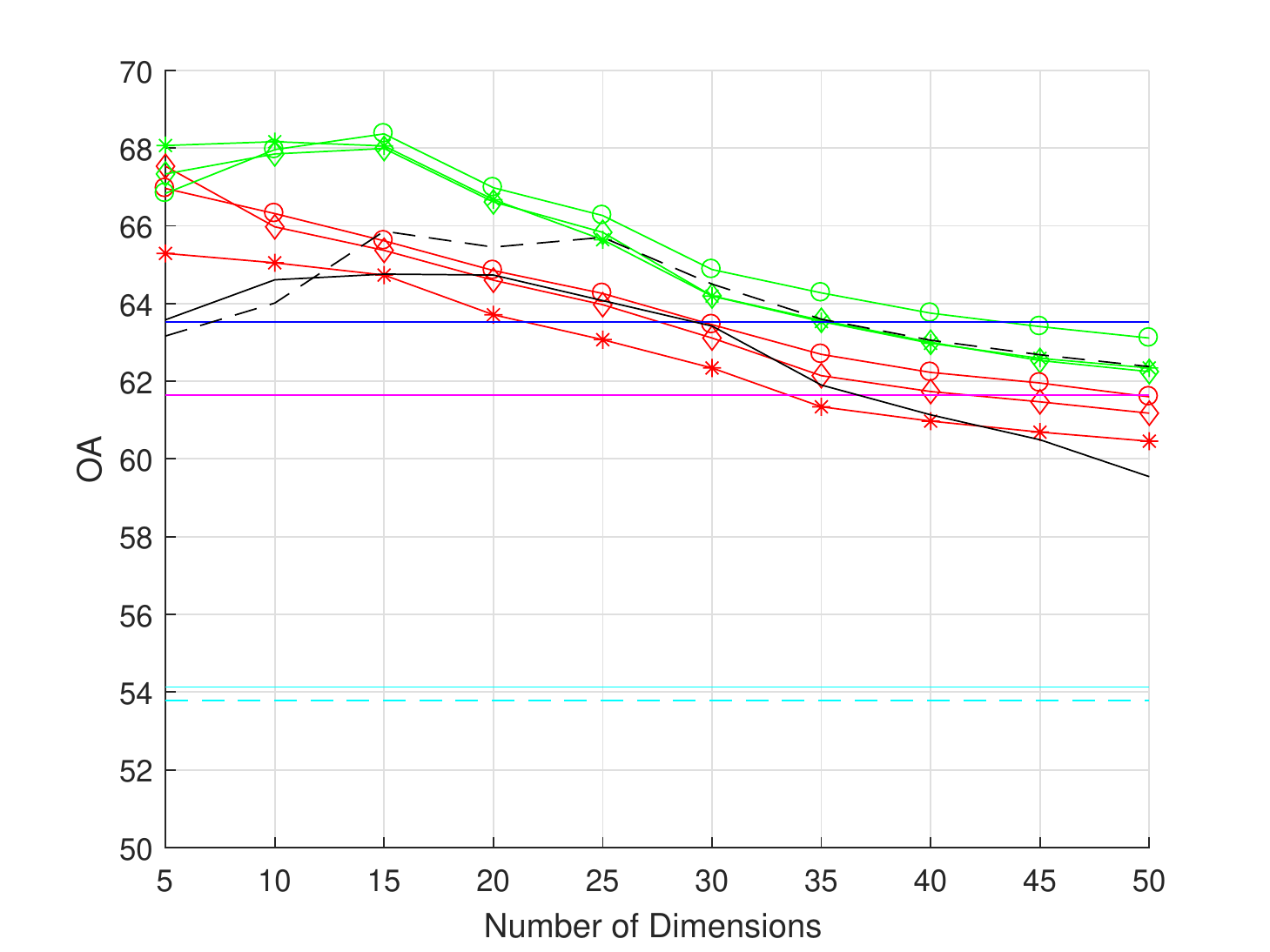}  &
      \includegraphics[width=0.3\textwidth]{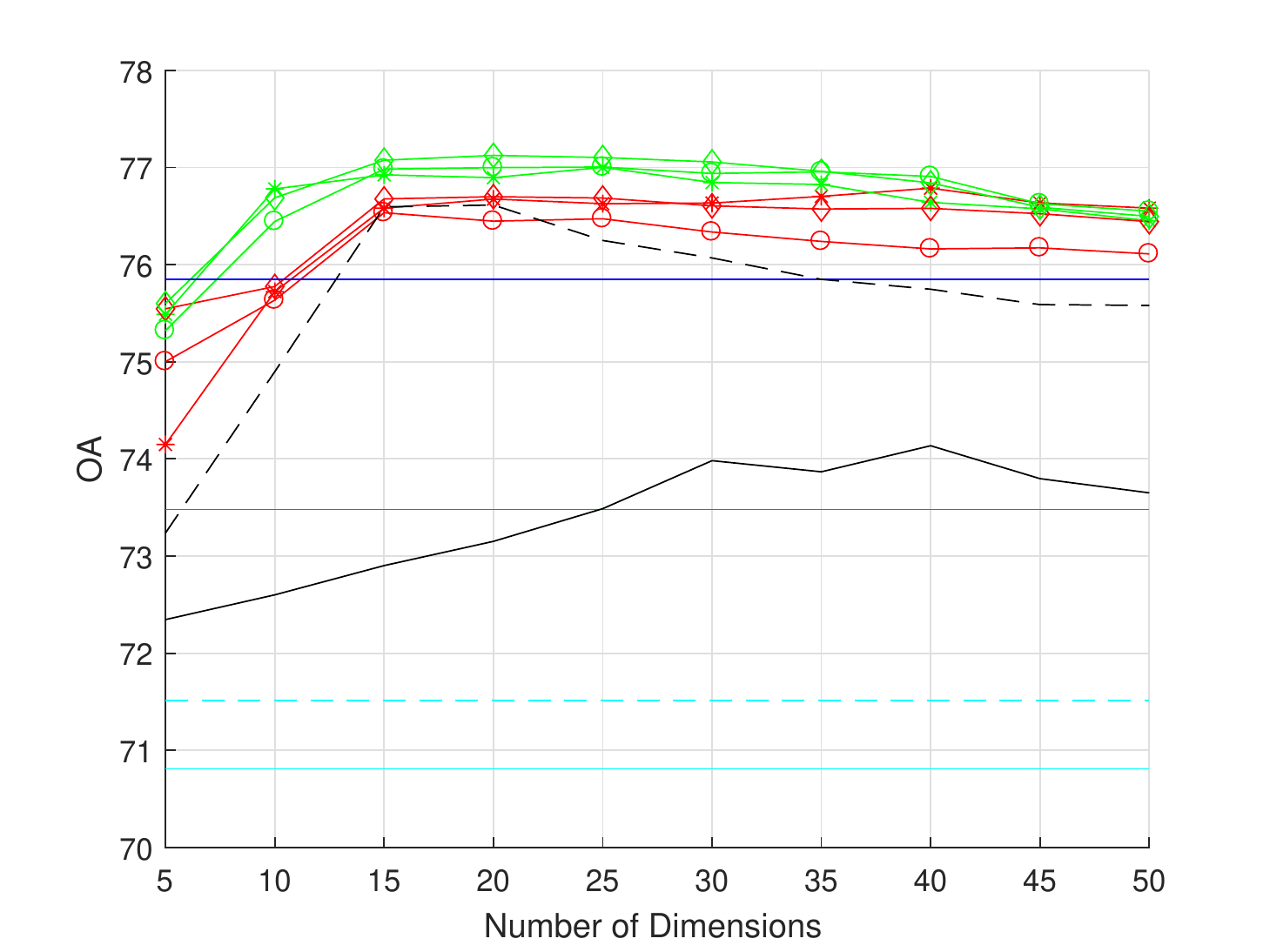} &
      \includegraphics[width=0.3\textwidth]{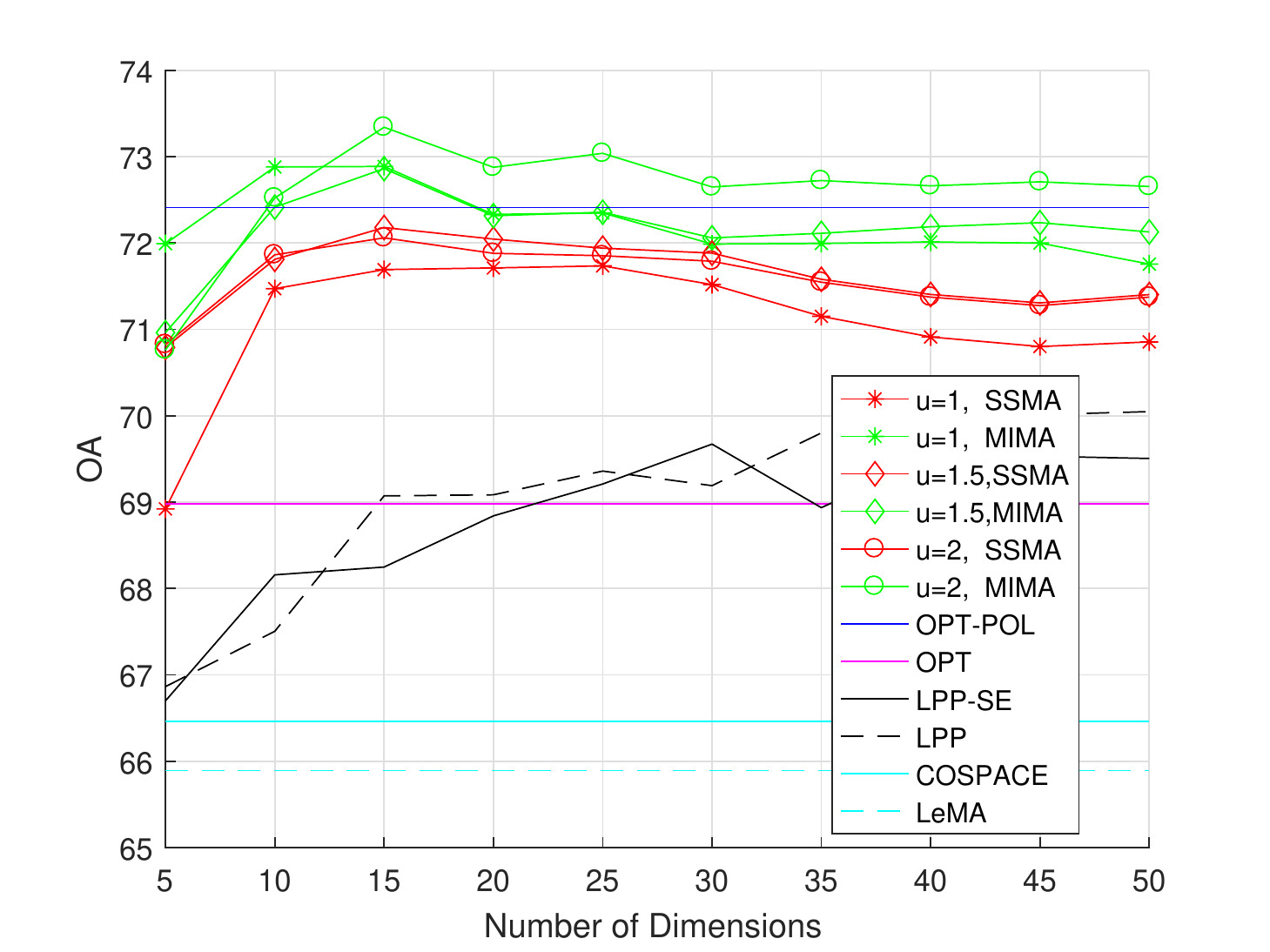} \\
    \end{tabular}        
    \caption{Classification performance in terms of overall accuracy (OA) for the experiments applied on the LCLU data set. The charts show the results of the three classifiers, from left to right, \textit{ONE-NN}, \textit{LSVM}, and \textit{KSVM}.}
\label{fig:LCLUexp}
\end{figure*}

\begin{table*}
\scriptsize
\centering
\caption{Quantitative performance comparison with the different algorithms on the LCLU data, in terms of class-specific accuracy, kappa coefficient, average accuracy, overall accuracy, and mean overall accuracy. The best performance achieved is shown in bold. Power $({a},{b})$ of learning rates $\alpha = 10^{a}$ and $\beta = 10^{b}$ are shown for COSPACE and LeMA in terms of the best performance. Number of dimensions ($dn$) is indicated as $(dn)$ for the best performance of LPP and LPP-SE. Parameter values of $\mu$ and number of dimensions ($dn$) are indicated as $(\mu, dn)$ for the best performance of SSMA and MIMA.}
\resizebox{\textwidth}{!}{
\begin{tabular}{C|C|C|CCCCCCCC|C|C|C|C}
\hline
\hline
&&&&&&&&&&&&&&\\
{Algorithm}         & {Parameter}         & {Classifiers} & {Forest} & {Residential Area} & Industrial Area & Low Plants  & Soil  & Allotment & Commercial Area & Water & KAPPA   & AA    &  OA  & Mean OA\\
&&&&&&&&&&&&&&\\[-1em]
\midrule
\multirow{3}{*}{\textit{POL}} 	& \multirow{3}{*}{-} & 1NN         & 35.19  & 51.03       & 21.54      & 25.43 & 62.1  & 29.24     & 29.22      & 28.36 & 0.2353 & 35.26 &  43.19 &\multirow{3}{*}{51.02}\\
              				    &                    & LSVM        & 30.32  & 75.65       & 7.93       & 26.18 & 85.17 & 39.99     & 28.92      & 28.42 & 0.3614 & 40.32 &  57.84 &\\
                 				&                    & KSVM        & 34.15  & 66.71       & 9.17       & 20.51 & 67.35 & 31.2      & 42.17      & 22.44 & 0.2999 & 36.71 &  52.04 &\\
\midrule
\multirow{3}{*}{\textit{OPT}}  & \multirow{3}{*}{-} & 1NN         & 68.78  & 63.87       & 30.01      & 57.58 & 90.73 & 55.76     & 32.86      & 73.89 & 0.4599 & 59.18 &  61.64 &\multirow{3}{*}{68.03}\\
                               &                    & LSVM        & 69.2   & 82.5        & 18.55      & 65.7  & 79.06 & 53.59     & 44.77      & 72.81 & 0.585  & 60.77 &  73.48 &\\
                               &                    & KSVM        & 66.54  & 77.68       & 26.18      & 58.33 & 65.16 & 54.94     & 39.64      & 72.74 & 0.5269 & 57.65 &  68.98 &\\
\midrule
\multirow{3}{*}{\textit{OPT-POL}} & \multirow{3}{*}{-} & 1NN         & 67.02  & 66.41       & 32.99      & 63.54 & 83.57 & 55.54     & 33.53      & 67.09 &  0.478 & 58.71 &  63.53 &\multirow{3}{*}{70.60}\\
                                  &                    & LSVM        & 70.13  & 84.63       & 30.43      & 71.58 & 79.22 & 56.27     & 37.67      & 75.03 & 0.6183 & 63.12 &  75.85 &\\
                                  &                    & KSVM        & 61.76  & 82.97       & 29.7       & 61.17 & 76.55 & 52.98     & 40.79      & 70.36 & 0.5667 & 59.53 &  72.41 &\\

\midrule
\multirow{3}{*}{COSPACE} & $(-1,-2)$ & 1NN  & 47.72 & 58.67 & 24.27 & 55.99 & 61.32 & 47.67 & 29.81 & 60.93 & 0.3596 & 48.3  & 54.13 & \multirow{3}{*}{63.80}\\
                                        & $(-1,-3)$ & LSVM & 55.04 & 82.24 & 27.48 & 66.28 & 75.98 & 57.63 & 23.66 & 51.49 & 0.5379 & 54.98 & 70.81 & \\
                                        & $(-1,-4)$ & KSVM & 34.14 & 85.08 & 17.03 & 52.93 & 48.37 & 46.87 & 22.51 & 36.52 & 0.4395 & 42.93 & 66.46 & \\

\midrule
\multirow{3}{*}{LeMA}    & $(-1,-1)$ & 1NN  & 50.93 & 57.07 & 26.88 & 56.86 & 63.67 & 49.68 & 26.92 & 61.53 & 0.3613 & 49.19 & 53.78 & \multirow{3}{*}{63.73}\\
                                        & $(-1,-1)$ & LSVM & 68.52 & 80.13 & 25.38 & 64.57 & 76.54 & 62.99 & 29.41 & 70.12 & 0.5601 & 59.71 & 71.51 &\\
                                        & $(-2,-2)$ & KSVM & 36.57 & 84.21 & 13.64 & 48.01 & 55.22 & 45.66 & 27.45 & 36.02 & 0.4345 & 43.35 & 65.89 &\\

\midrule
\multirow{3}{*}{\textit{LPP}}   &  $(15)$ & 1NN         & 70.81  & 68.84       & 34.57      & 64.82 & 89.64 & 51.91     & 33.22      & 76.17 & 0.5057 & 61.25 &  65.86 & \multirow{3}{*}{70.86}\\
                                &  $(20)$ & LSVM        & 69.39  & 85.56       & 32.84      & 71.41 & 81.94 & 57.46     & 38.23      & 76.67 & 0.6255 & 64.19 &  76.55 &\\
                                &  $(50)$ & KSVM        & 59.94  & 79.37       & 21.41      & 66.37 & 63.32 & 54.66     & 46.84      & 61.78 & 0.5394 & 56.71 &  70.18 &\\
\midrule
\multirow{3}{*}{\textit{LPP-SE}}    & $(15)$ & 1NN         & 67.21  & 68.2        & 37.33      & 65.06 & 79.01 & 55.61     & 29.5       & 75.57 & 0.4898 & 59.69 &  64.75 & \multirow{3}{*}{69.53}\\
                                    & $(40)$ & LSVM        & 66.19  & 83.62       & 36.88      & 66.44 & 81.5  & 60.61     & 28.31      & 75.43 & 0.5926 & 62.37 &  74.12 &\\
                                    & $(30)$ & KSVM        & 62.07  & 79.3        & 34.73      & 55.15 & 76.33 & 55.04     & 39.41      & 72.97 & 0.5326 & 59.38 &  69.71 &\\
\midrule
\multirow{3}{*}{\textit{SSMA}}  & $(1.5, 5)$  & 1NN         & 69.26  & 71.78       & 33.67      & 67.44 & 84.3  & 55.94     & 31.93      & 71.08 & 0.5222 & 60.67 &  67.53 & \multirow{3}{*}{72.25}\\
                                & $( 1, 40)$  & LSVM        & 66.07  & 85.31       & 34.67      & 80.05 & 78.83 & 57.13     & 34.59      & 70.1  & 0.632  & 63.34 &  76.82 &\\
                                & $(1.5, 15)$ & KSVM        & 59.65  & 83.6        & 30.21      & 67.64 & 43.58 & 55.47     & 41.93      & 68.44 & 0.5642 & 56.32 &  72.4  &\\
\midrule
\multirow{3}{*}{\textit{MIMA}}  & $( 2, 15)$    & 1NN         & 71.29  & 72.5        & 35.6       & 71.85 & 66.86 & 59.78     & 31.83      & 71.29 & 0.5329             & 60.13         &  68.36        & \multirow{3}{*}{\textbf{73.04}}\\
                                & $(1.5, 20)$   & LSVM        & 67.58  & 86.05       & 36.32      & 76.79 & 78.35 & 58.75     & 33.71      & 76.06 & \textbf{0.6367}    & \textbf{64.2} & \textbf{77.15}& \\
                                & $( 2, 15)$    & KSVM        & 64.3   & 84.21       & 34.46      & 65.73 & 54.74 & 57.45     & 40.54      & 69.71 & 0.5844             & 58.89         & 73.6          & \\
\bottomrule
\bottomrule
\end{tabular}}
\label{tb:LCLU}
\end{table*}

Besides the parameter tunning, one important part of MIMA is to select the filtering function. As discussed before, the filtering function provides a perspective of observing the data and introduces field knowledge. As principal components are widely used in classification of remote sensing data and have been proven to be effective, and this is the first attempt of applying MAPPER in remote sensing, the first and second principal components are chosen to serve as the field knowledge in this work.

As shown in Table~\ref{tb:para}, (\textit{D}) COSPACE, (\textit{E}) LeMA, (\textit{H}) SSMA, and (\textit{I}) MIMA are all fall into the manifold alignment fusion strategy. However, their learning resources are different. COSPACE is designed to learn a joint latent space via the existed labeled data. In addition to the labeled data, LeMA also uses the pseudo-labeled data, predictions of a trained classifier on unlabeled data, to include unlabeled data into the procedure of data fusion. For SSMA and MIMA, they utilize the labeled data and extract the data distribution under the guidance of mathematical assumptions, to achieve data fusion. Therefore, when a large amount of labeled data exists or the data distribution is not correlated with labels, LeMA would be more appropriate than SSMA and MIMA, and vice versa.

Since our goal is to assess the performance of fusion, three classical classifiers are chosen: the one-nearest-neighbor classifier (\textit{ONE-NN}), the linear support vector machine (\textit{LSVM}), and the Gaussian kernel support vector machine (\textit{KSVM}). In this work, parameter tuning of \textit{LSVM} and \textit{KSVM} are done in a heuristic procedure \cite{scholkopf2002learning}.

\subsection{Classification on the LCLU data set}

This section demonstrates and discusses the experimental results obtained on the LCLU data set.

\subsubsection{Fusion vs. non-fusion.} As shown in Fig.~\ref{fig:LCLUexp} and Table~\ref{tb:LCLU}, classification on fused hyperspectral imagery and dual-Pol SAR data outperforms classification on the individual data source, in terms of classification accuracies. Among the fusion algorithms, our proposed MIMA provides the best classification performance, which, in terms of overall accuracy, exceeds classifications on dual-Pol SAR data by 25\%, 20\%, and 21\% and exceeds classifications on hyperspectral imagery by 7\%, 4\%, and 5\%, using \textit{ONE-NN}, \textit{LSVM}, and \textit{KSVM}, respectively. This proves that fusion of hyperspectral imagery and dual-Pol SAR data is advantageous to LCLU classification.

\begin{figure*}[t]
	\centering	
	{\footnotesize
    \begin{tabular}{ccccc}
      \includegraphics[width=0.18\textwidth]{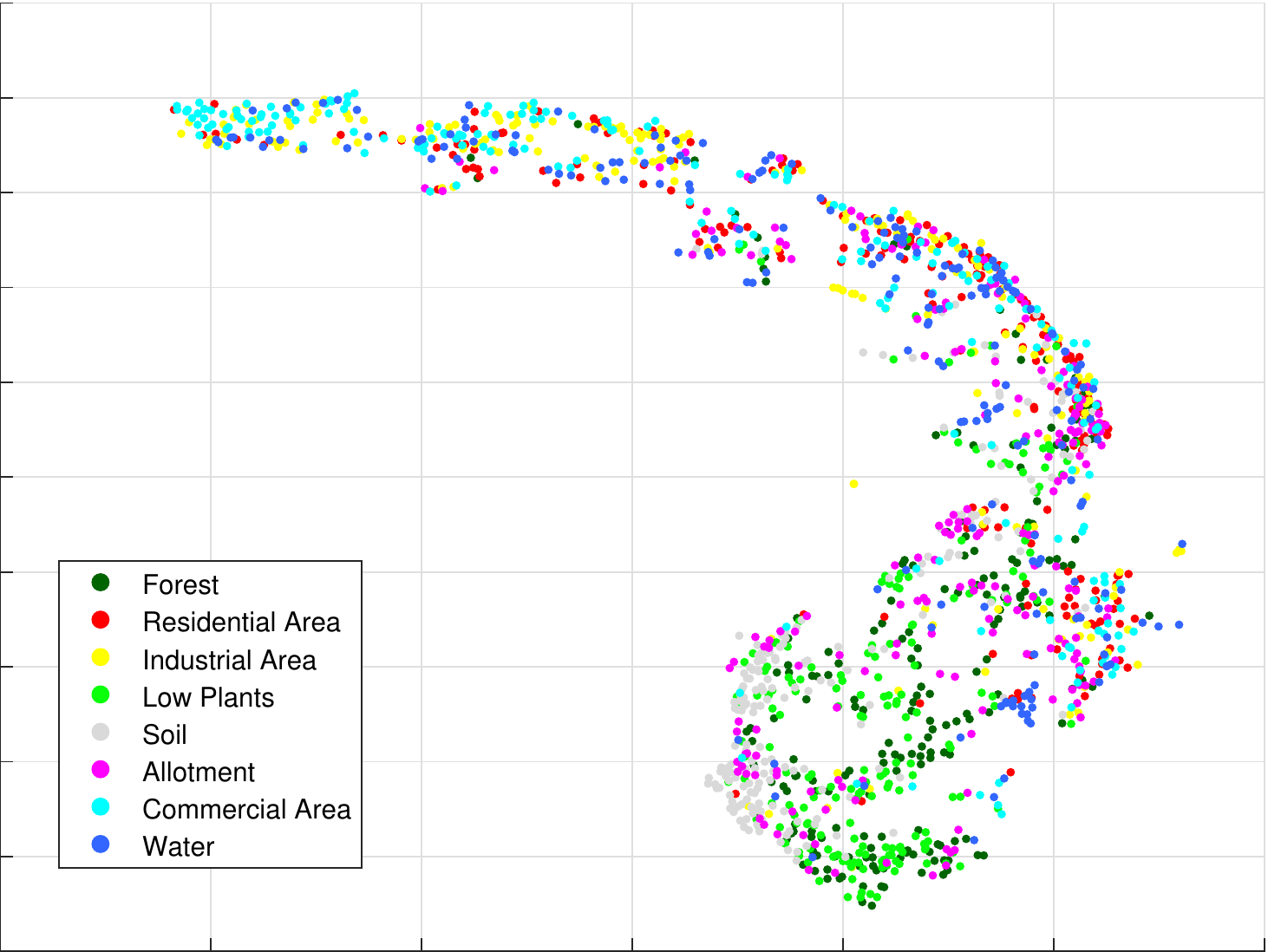}      &
      \includegraphics[width=0.18\textwidth]{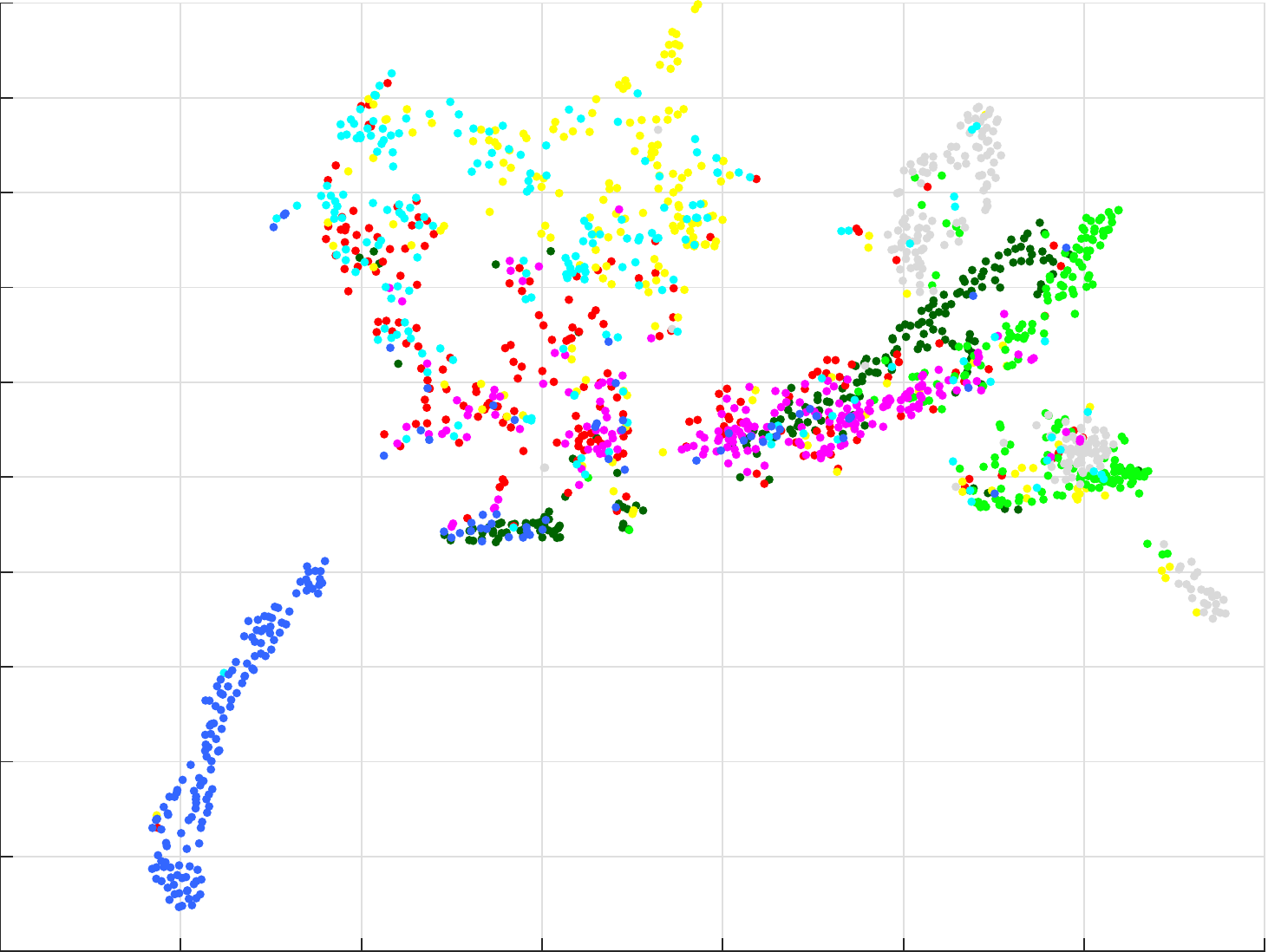}      &
      \includegraphics[width=0.18\textwidth]{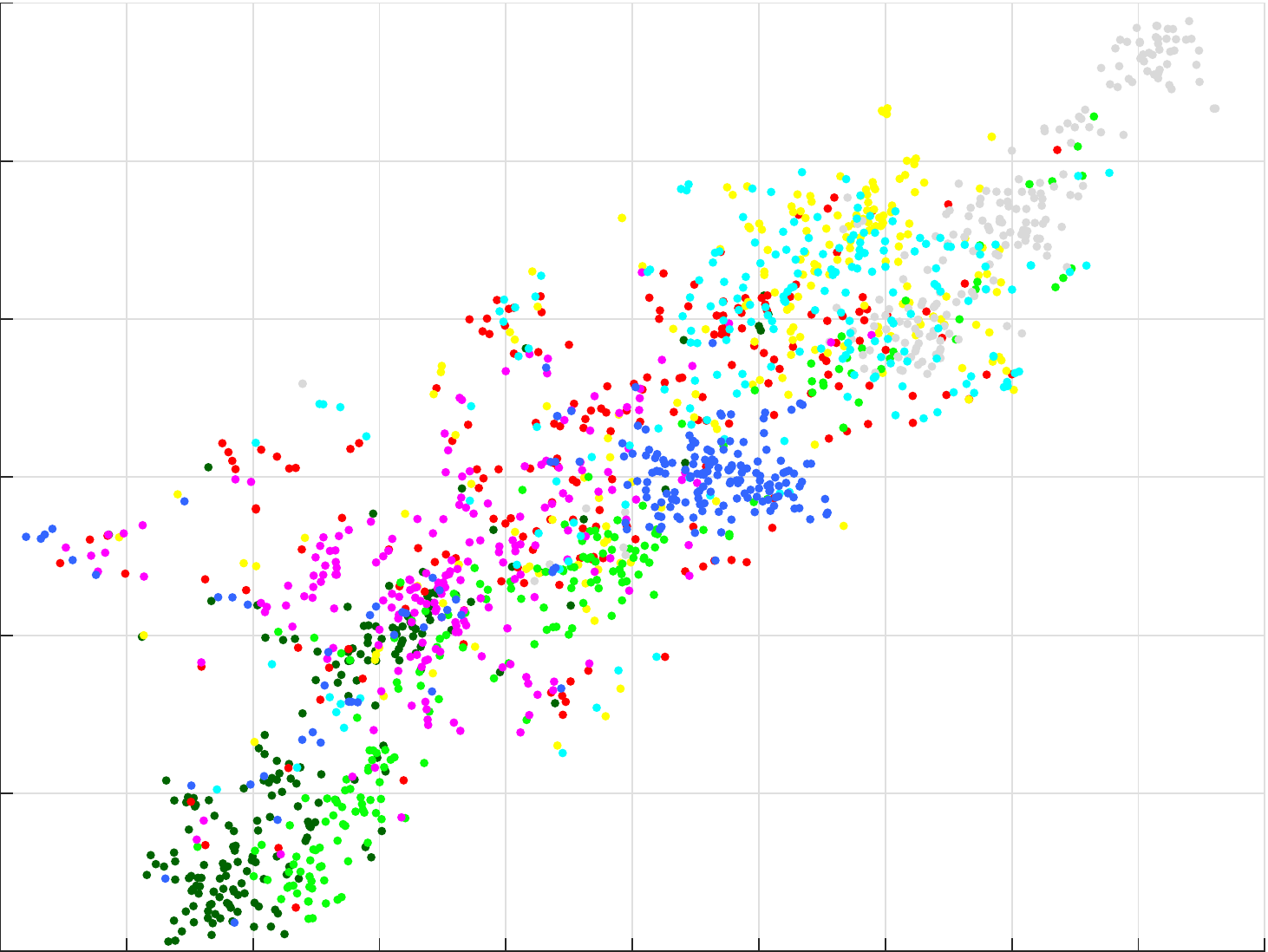}      &
      \includegraphics[width=0.18\textwidth]{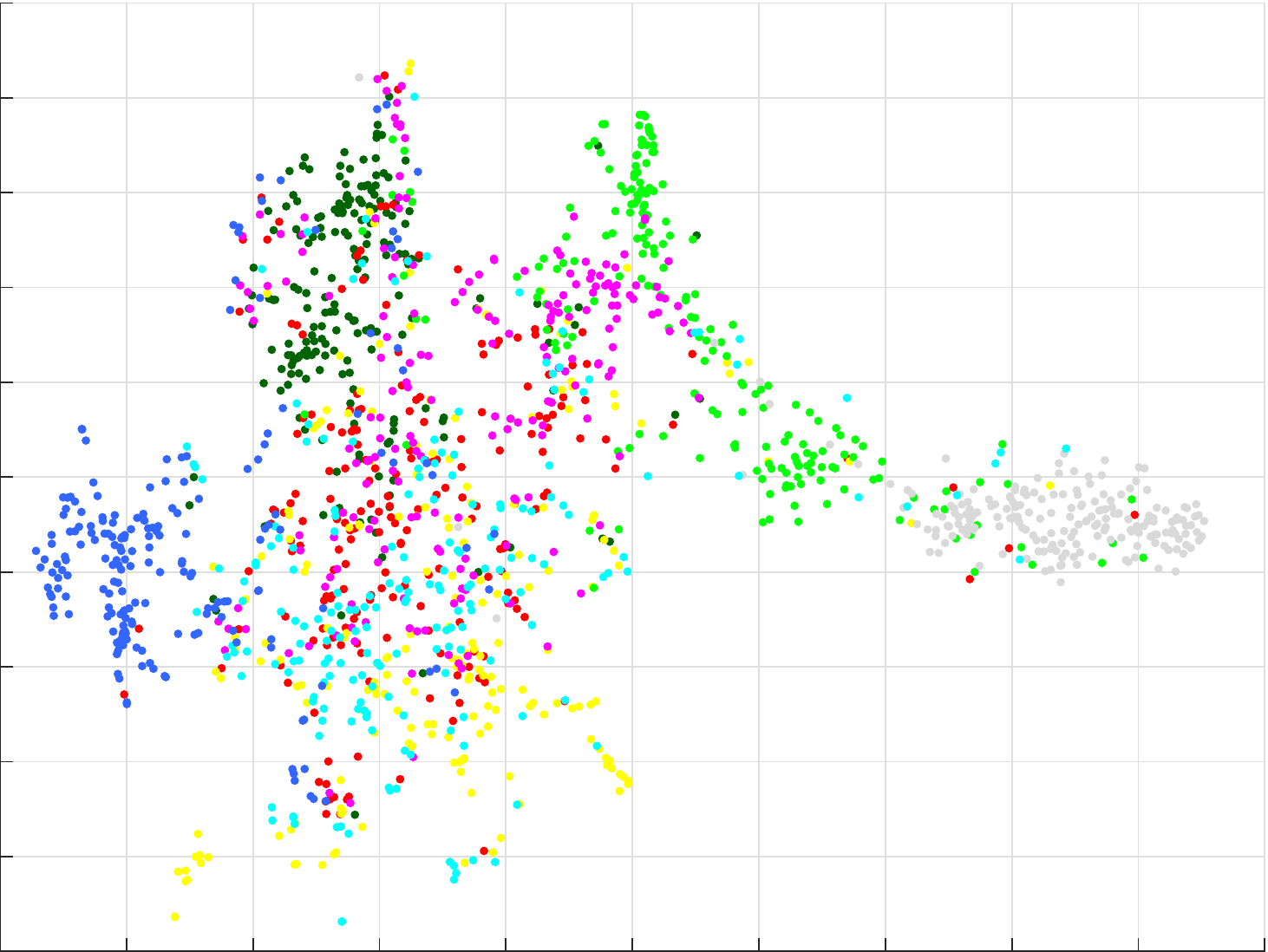}   &
      \includegraphics[width=0.18\textwidth]{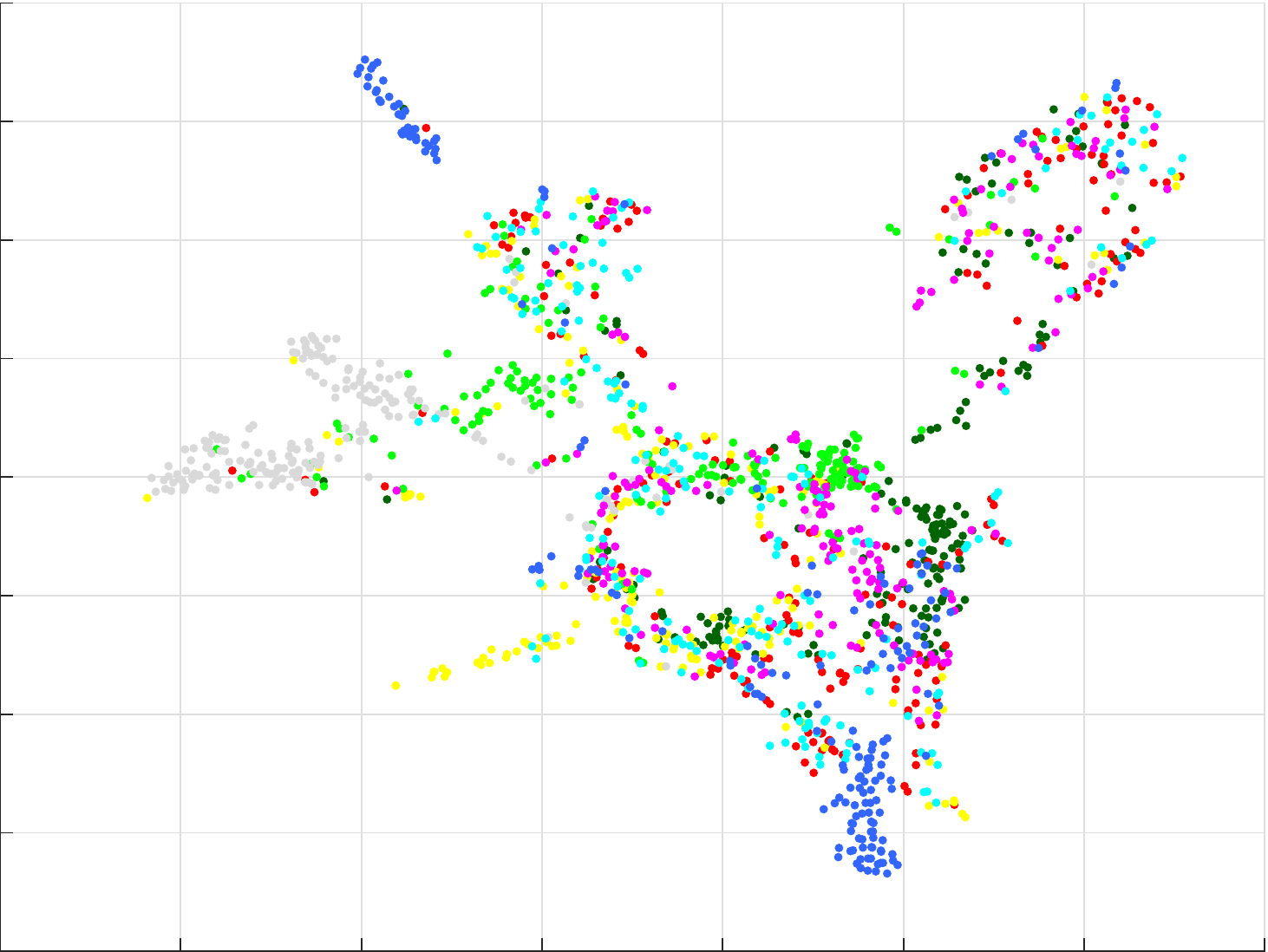}   \\

      PolSAR data space         &
      Optical data space        &
      LPP projected space       &
      LPP-SE projected space    &
      COSPACE projected space   \\
      
      \includegraphics[width=0.18\textwidth]{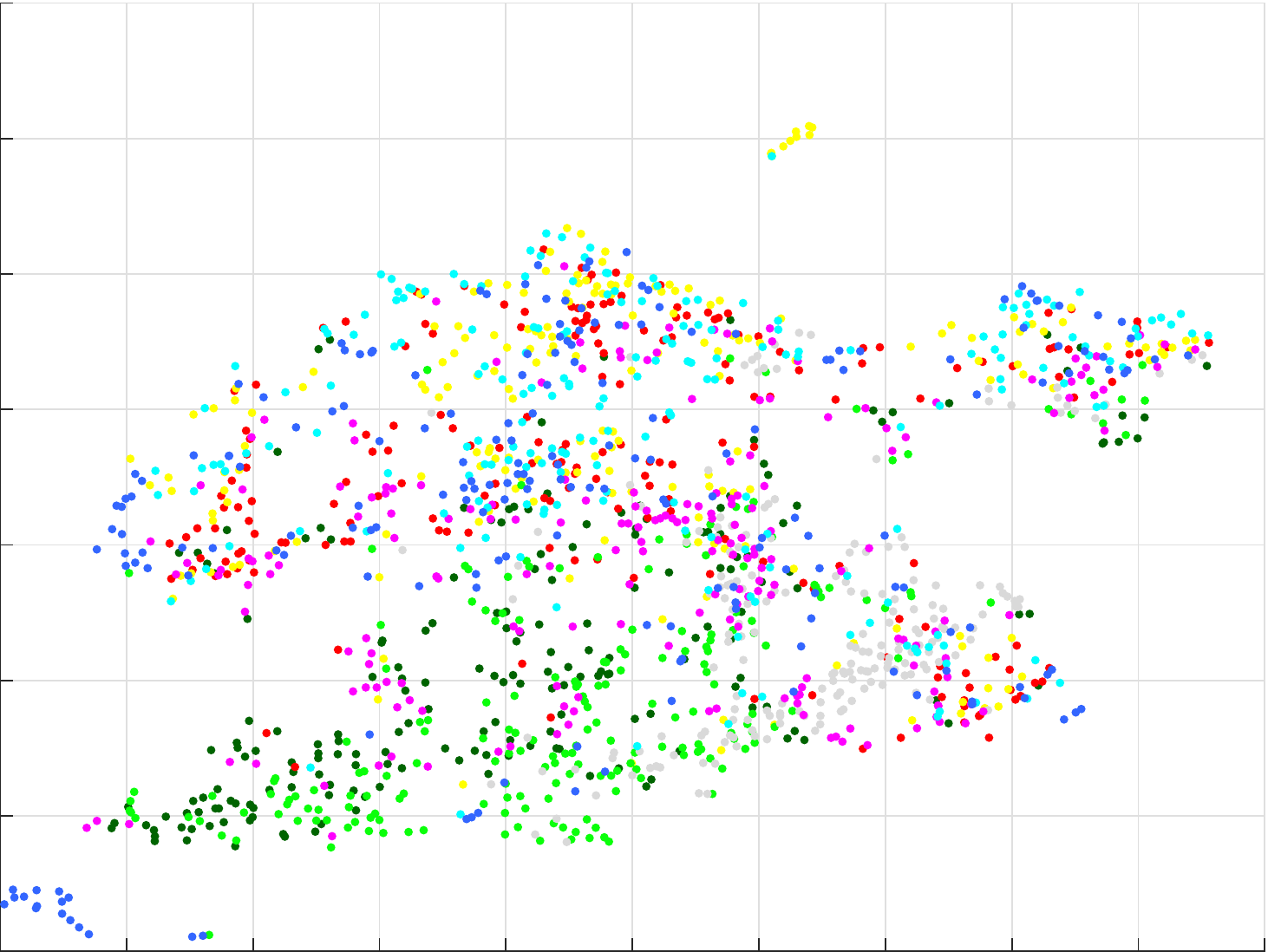}  &
      \includegraphics[width=0.18\textwidth]{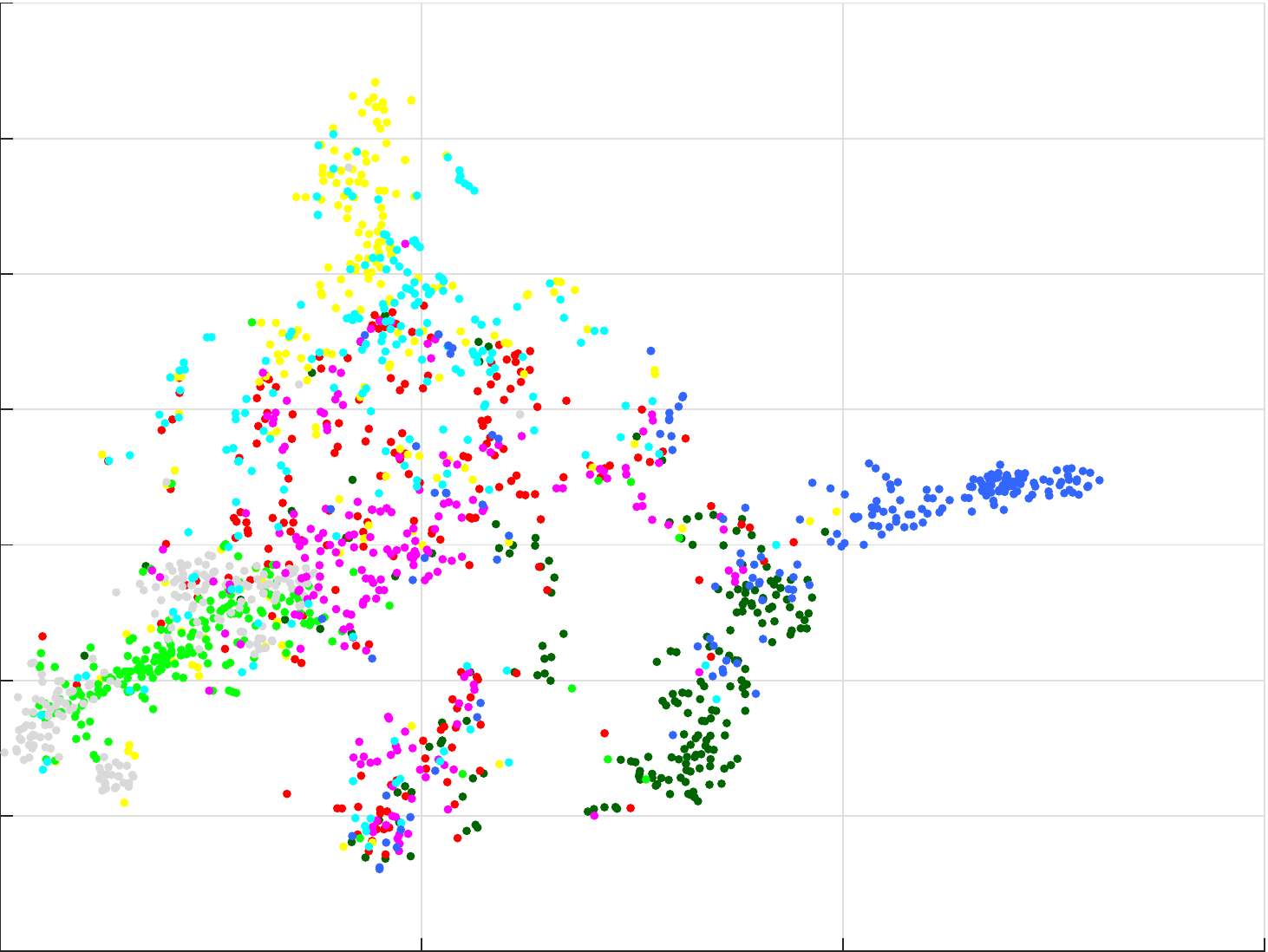}  &
      \includegraphics[width=0.18\textwidth]{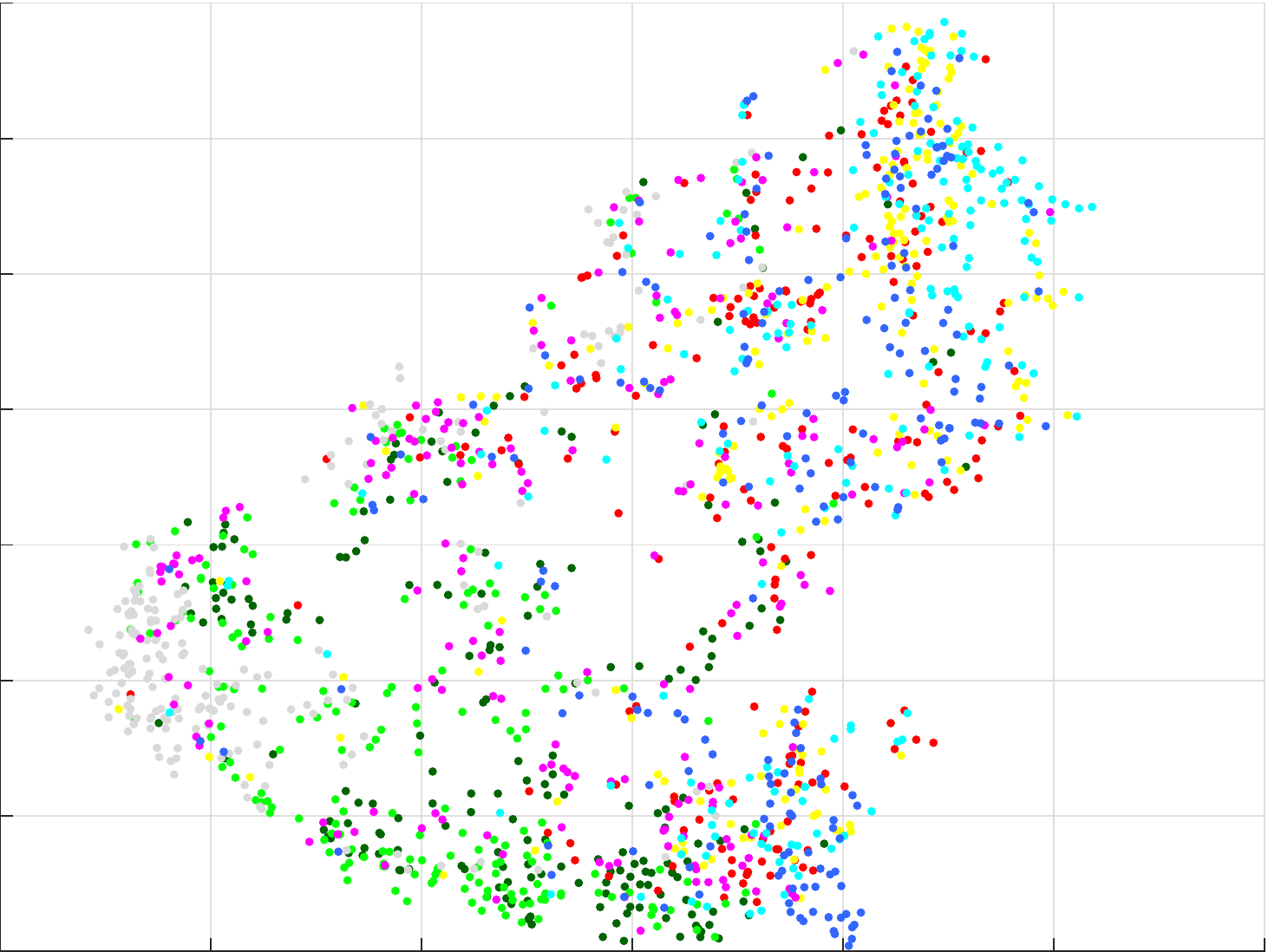}  &
      \includegraphics[width=0.18\textwidth]{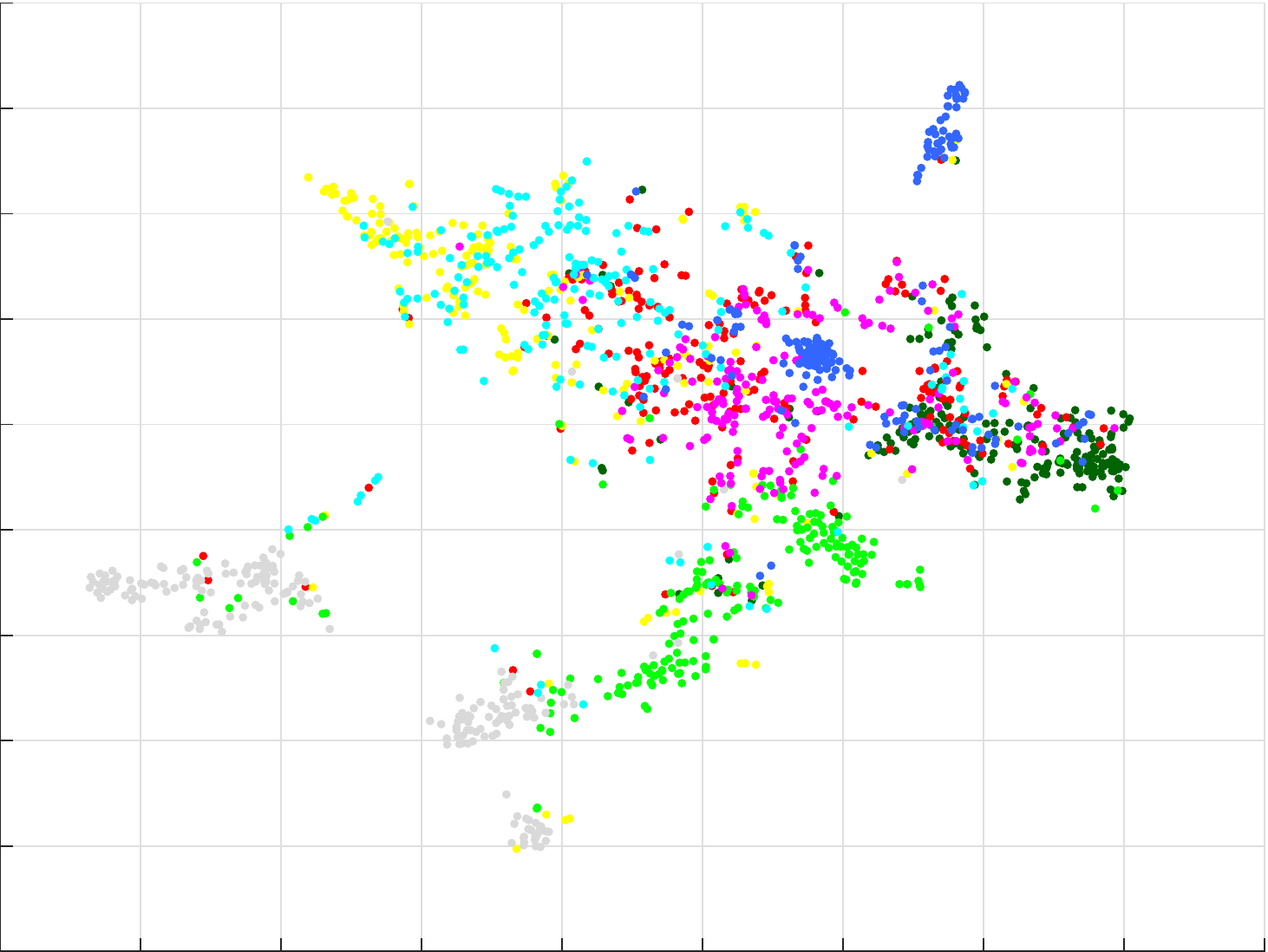}  &
      \includegraphics[width=0.18\textwidth]{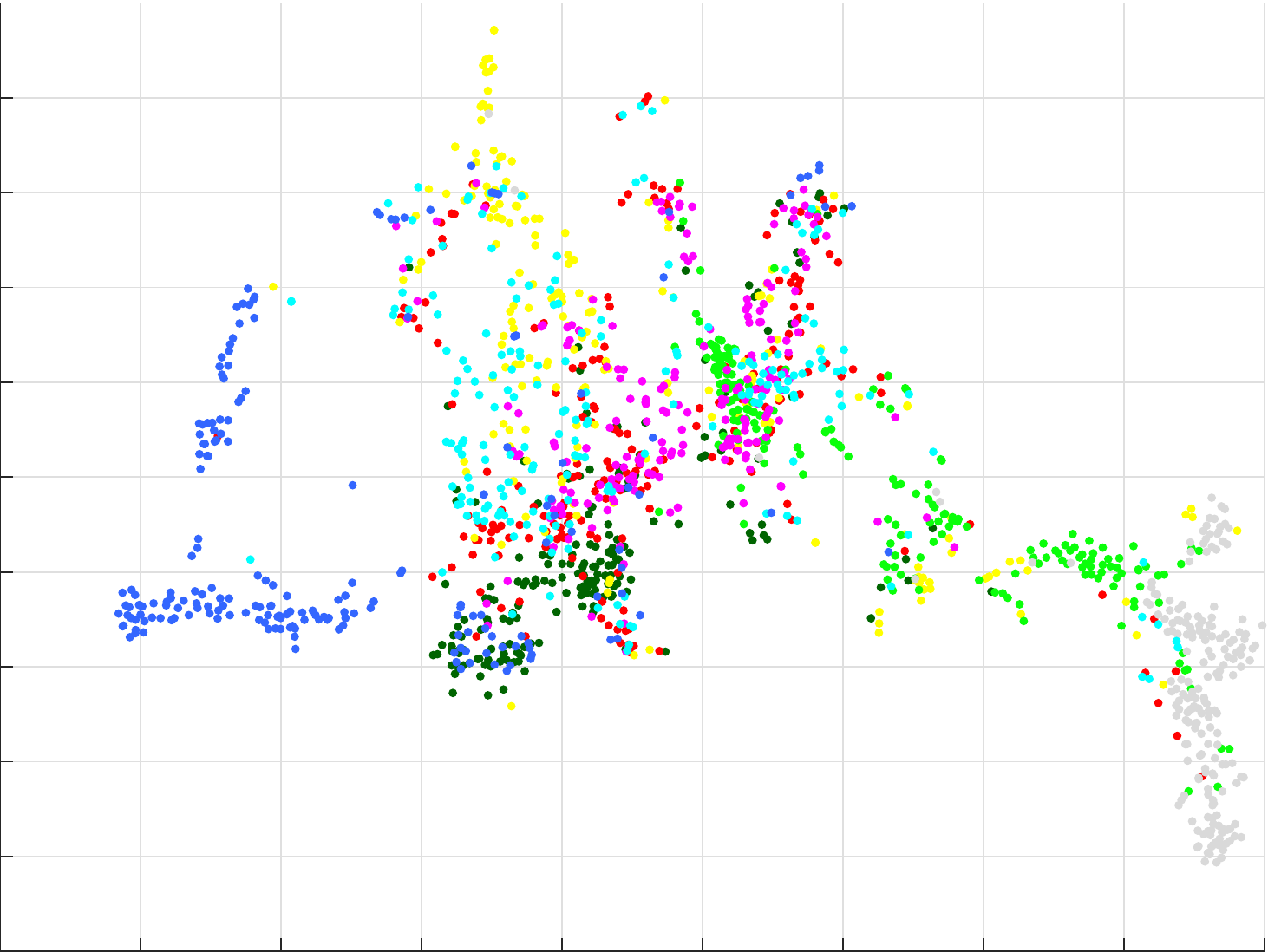}      \\

      SSMA projected & SSMA projected & MIMA projected & MIMA projected & \multirow{2}{*}{LeMA projected space} \\ 
      PolSAR data space & optical data space & PolSAR data space & optical data space &

    \end{tabular}} 
    \caption{Visualization of the optical data and the PolSAR data of the LCLU data set, using t-SNE \cite{maaten2008visualizing} in their original and projected spaces. The x and y axis are the first and second dimensions resluted from the t-SNE. The first row are: the PolSAR data in the original space, the optical data in the original space, LPP jointly projected space, LPP-SE jointly projected space, and COSPACE projected space, respectively. The second row are: the PolSAR data in SSMA projected space, the optical data in SSMA projected space, the PolSAR data in MIMA projected space, the optical data in MIMA projected space, and LeMA projected space, respectively.}
	\label{fig:LCLUVis}
\end{figure*}

\subsubsection{Fusion categories.} Based on properties of fusion algorithms listed in Table~\ref{tb:para}, to simplify the discussion, we would like to divide these seven fusion algorithms into four categories: (1) feature concatenation (OPT-POL); (2) joint dimension reduction fusion (LPP and LPP-SE); (3) label-driven manifold alignment (COSPACE and LeMA); and (4) data-driven manifold alignment fusion (SSMA and MIMA). According to the classification accuracy in Table~\ref{tb:labLCLU}, with the feature concatenation (OPT-POL) serves as the benchmark, it is obvious to find that: (a) joint dimension reduction fusion algorithms achieve similar classification accuracy to the feature concatenation fusion; (b) the overall accuracy provided by label-driven manifold alignment are around 7\% lower than the accuracy achieved by the feature concatenation; (c) the data-driven manifold alignment fusion outperforms the feature concatenation by 2\%. Discussions regarding the three findings are detailed as follows.

It is well known that the dimension reduction technique is capable of boosting the classification accuracy, due to the curse-of-dimensionality \cite{donoho2000high}. However, according to the finding (a) above, this doesn't suit the feature concatenation fusion in out experiment. Because the curse-of-dimensionality has been tackled in our feature design. The finding (a) also validates that the improvement of our proposed method is not a side effect of dimension reduction.

The label-driven manifold alignment fusion learns projections that map original data sources to a latent space purely based on the label, and applies learned projections on the unlabeled data to accomplish fusion. The finding (b) gives a clear clue that this type of fusion can not provide a proper fusion result for the LCLU data set. This could because the label-driven learned latent space is not applicable to a general case, namely the unlabeled data. Thus, the label-driven manifold alignment fusion might provide a destructive fusion when the label data can not represent the data distribution which is often the case in remote sensing.

The data-driven manifold alignment fusion also learns projections that map original data sources to a latent space. However, the latent space is jointly defined by the label and the data structure explored from the original data sources, including labeled and unlabeled data. The finding (c) suggests that the data-driven manifold alignment fusion is an effective fusion strategy, which improves the overall accuracy about 2\% by comparing to the feature concatenation fusion.

\begin{figure*}
	\centering	
    \begin{tabular}{ccc}
      \includegraphics[width=0.3\textwidth]{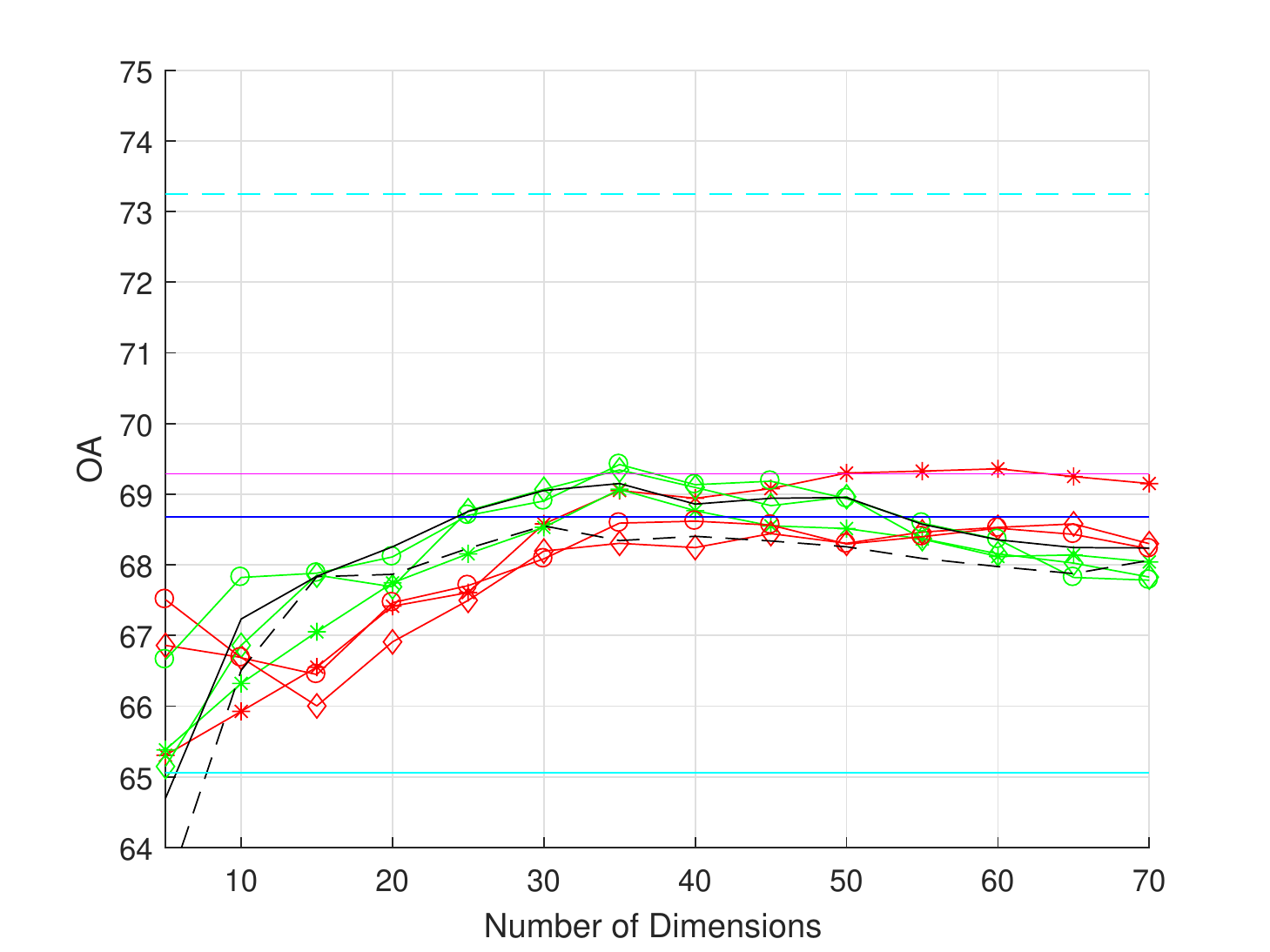} &

      \includegraphics[width=0.3\textwidth]{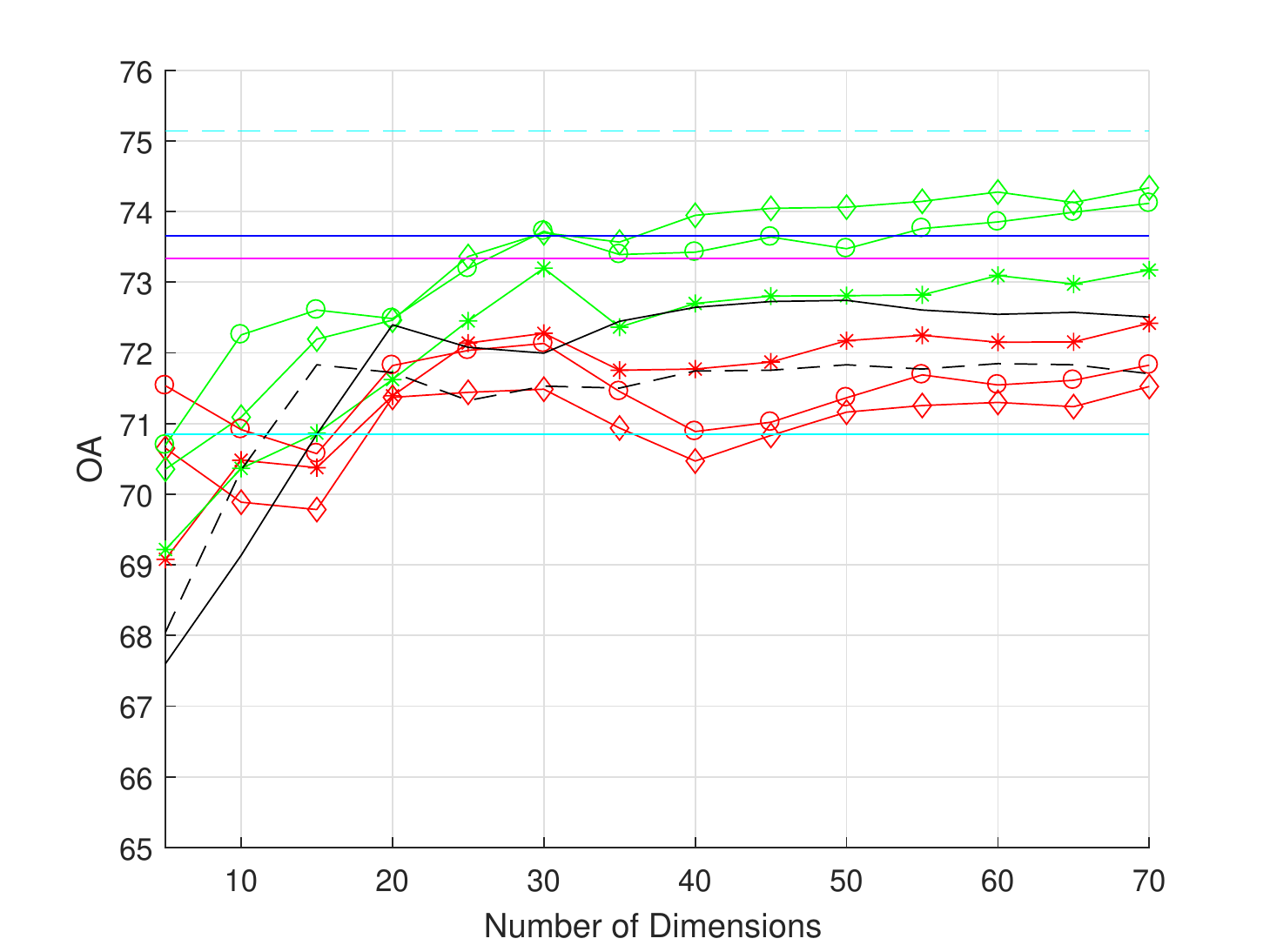} &

      \includegraphics[width=0.3\textwidth]{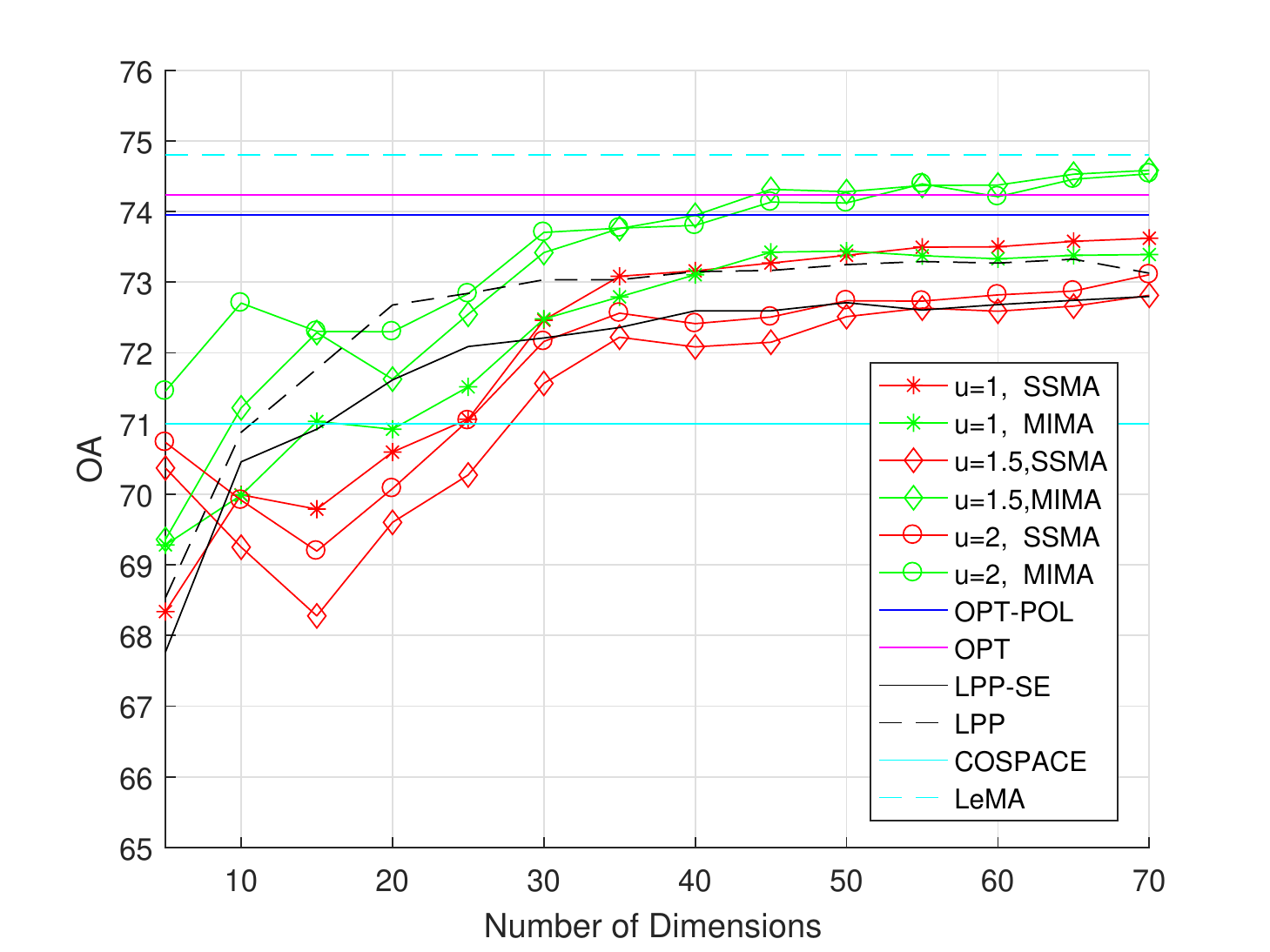} 
    \end{tabular}    
    \caption{Classification performance in terms of overall accuracy (OA) for the experiments applied on the LCZ data set. The charts show results for the three classifiers, from left to right, \textit{ONE-NN}, \textit{LSVM}, and \textit{KSVM}.}
    \label{fig:LCZexp}
\end{figure*}

\begin{table*}
\scriptsize
\centering
\caption{Quantitative performance comparison with the different algorithms on the LCZ data, in terms of class-specific accuracy, kappa coefficient, average accuracy, overall accuracy, and mean overall accuracy. The best performance achieved is shown in bold. Power $({a},{b})$ of learning rates $\alpha = 10^{a}$ and $\beta = 10^{b}$ are shown for COSPACE and LeMA in terms of the best performance. Number of dimensions ($dn$) is indicated as $(dn)$ for the best performance of LPP and LPP-SE. Parameter values of $\mu$ and number of dimensions ($dn$) are indicated as $(\mu, dn)$ for the best performance of SSMA and MIMA. }
\resizebox{\textwidth}{!}{
\begin{tabular}{C|C|C|CCCCCCCCCCCC|C|C|C|C}
\hline
\hline
&&&&&&&&&&&&&&\\
Algorithm         & Parameter         & Classifiers & Compact mid-rise & Open high-rise & Open mid-rise & Open low-rise & Large low-rise & Sparsely built & Dense trees & Scattered trees & Bush, scrub & Low plants & Bare soil or sands & Water & KAPPA     & AA   & OA  & Mean OA \\
&&&&&&&&&&&&&&\\[-1em]
\midrule
\multirow{3}{*}{POL}& \multirow{3}{*}{-} & 1NN  & 43.76 & 29.13 & 21.4 & 59.26 & 33.64 & 18.54 & 82.33 & 10.98 & 19.59 & 70.62 & 28.13 & 54.23 & 0.4566 & 39.3  & 52.66 &\multirow{3}{*}{59.47} \\
                    &                    & LSVM & 24.78 & 18.93 & 23.92& 71.1  & 48.95 & 7.11  & 96.23 & 2.25  & 14.51 & 94.19 & 32.29 & 88.88 & 0.5688 & 43.6  & 63.03 &\\
                    &                    & KSVM & 40.6  & 20.87 & 24.67& 75.46 & 46.76 & 10.75 & 95.3  & 0.13  & 14.37 & 92.04 & 33.85 & 66.4  & 0.5637 & 43.43 & 62.72 &\\
\midrule
\multirow{3}{*}{OPT}& \multirow{3}{*}{-} & 1NN  & 72.32 & 39.56 & 40.63& 67.24 & 53.99 & 20.97 & 98.63 & 33.99 & 32.66 & 79.46 & 72.4  & 98.95 & 0.648  & 59.23 & 69.29 &\multirow{3}{*}{72.29} \\
                    &                    & LSVM & 72.93 & 47.33 & 52.55& 74.66 & 53.66 & 20.45 & 99.5  & 37.7  & 28.01 & 84.87 & 79.69 & 99.19 & 0.6933 & 62.55 & 73.34 & \\
                    &                    & KSVM & 70.3  & 36.17 & 51.16& 73.19 & 65.85 & 23.92 & 99.77 & 34.79 & 38.9  & 90.23 & 65.63 & 94.36 & 0.7024 & 62.02 & 74.24 & \\
\midrule
\multirow{3}{*}{OPT-POL} & \multirow{3}{*}{-} & 1NN  & 61.42 & 35.44 & 26.3  & 73.83 & 50.55 & 23.92 & 98.71 & 37.3  & 35.27 & 82    & 75.52 & 98.39 & 0.6413 & 58.22 & 68.68 & \multirow{3}{*}{72.10}\\
                         &                    & LSVM & 49.3  & 33.01 & 47.45 & 80.4  & 61.14 & 22.18 & 99.42 & 46.16 & 23.66 & 89.62 & 89.06 & 99.03 & 0.6957 & 61.7  & 73.66 &\\
                         &                    & KSVM & 57.47 & 32.52 & 40.04 & 79.51 & 61.56 & 22.36 & 99.8  & 37.17 & 43.11 & 92.38 & 77.08 & 98.07 & 0.6991 & 61.75 & 73.95 &\\
\midrule
\multirow{3}{*}{COSPACE} & $(-1,-1)$ & 1NN  & 70.74 & 32.77 & 8.35  & 67.82 & 42.39 & 14.56 & 98.48 & 32.54 & 34.83 & 82.45 & 73.96 & 97.66 & 0.5961 & 54.71 & 65.06 & \multirow{3}{*}{68.97}\\
                                        & $(-1,-3)$ & LSVM & 69.86 & 63.83 & 25.41 & 78.44 & 54.58 & 18.02 & 99.15 & 34.39 & 44.56 & 83.27 & 77.6  & 93.07 & 0.6653 & 61.85 & 70.85 & \\
                                        & $(-1,-1)$ & KSVM & 68.63 & 45.39 & 25.75 & 75.74 & 71.83 & 10.05 & 99.39 & 31.48 & 42.09 & 87.5  & 60.94 & 90.01 & 0.6651 & 59.07 & 71    & \\

\midrule
\multirow{3}{*}{LeMA} & $(-1,-1)$ & 1NN  & 82.07 & 75.49 & 37.47 & 68.31 & 56.85 & 37.44 & 99.62 & 56.61 & 39.04 & 81.66 & 87.5  & 99.92 & 0.6953 & 68.5  & 73.25 & \multirow{3}{*}{\textbf{74.40}}\\
                      & $(-3,-2)$ & LSVM & 82.07 & 78.64 & 46.27 & 76.69 & 55    & 34.14 & 99.24 & 51.59 & 40.93 & 81.24 & 75.52 & 99.6  & \textbf{0.7166} & 68.41 & \textbf{75.14} &\\
                      & $(-1,-1)$ & KSVM & 75.22 & 74.27 & 42.26 & 72.73 & 67.45 & 35.88 & 99.74 & 50    & 47.46 & 86.35 & 86.98 & 87.11 & 0.7119 & \textbf{68.79} & 74.8 &\\ 
                      
\midrule
\multirow{3}{*}{LPP}  & $(30)$ & 1NN  & 60.98 & 62.62 & 33.22 & 68.65 & 51.3  & 26.69 & 99.77 & 44.31 & 33.96 & 75.22 & 85.94 & 98.23 & 0.6407 & 61.74 & 68.55 & \multirow{3}{*}{71.19} \\
                      & $(60)$ & LSVM & 49.12 & 62.86 & 45.13 & 68.77 & 60.22 & 25.3  & 99.8  & 44.31 & 50.8  & 83.06 & 79.17 & 99.19 & 0.6761 & 63.98 & 71.76 &\\
                      & $(65)$ & KSVM & 60.37 & 54.85 & 43.45 & 71.01 & 81.5  & 24.09 & 99.8  & 35.71 & 45.43 & 86.02 & 69.27 & 92.51 & 0.6927 & 63.67 & 73.27 &\\
\midrule
 \multirow{3}{*}{LPP-SE} & $(35)$ & 1NN  & 58.88 & 41.99 & 19.67 & 76.35 & 44.41 & 19.76 & 99.36 & 52.25 & 42.38 & 83    & 77.6  & 99.68 & 0.6453 & 59.61 & 69.15 & \multirow{3}{*}{71.57}\\
                         & $(50)$ & LSVM & 48.59 & 50    & 22.24 & 79.33 & 64.51 & 30.68 & 99.68 & 53.97 & 50.22 & 89.23 & 83.33 & 99.44 & 0.6867 & 64.27 & 72.77 &\\
                         & $(70)$ & KSVM & 43.23 & 44.17 & 25.46 & 77.88 & 81.92 & 20.45 & 99.77 & 48.68 & 51.38 & 89.77 & 70.31 & 96.21 & 0.6873 & 62.44 & 72.8  &\\
\midrule
\multirow{3}{*}{SSMA}    & $(1, 60)$ & 1NN  & 80.49 & 53.16 & 17.55 & 75.25 & 49.29 & 23.4  & 99.3  & 37.04 & 45.86 & 78.52 & 74.48 & 99.11 & 0.6496 & 61.12 & 69.36 & \multirow{3}{*}{71.70}\\
                         & $(1, 70)$ & LSVM & 72.06 & 65.53 & 36.38 & 71.81 & 58.37 & 27.73 & 99.59 & 37.17 & 53.85 & 82.09 & 78.13 & 98.55 & 0.6838 & 65.1  & 72.33 &\\
                         & $(1, 70)$ & KSVM & 71.53 & 61.17 & 32.43 & 78.22 & 80.07 & 23.74 & 99.8  & 33.6  & 50.8  & 80.94 & 56.25 & 96.54 & 0.6949 & 63.76 & 73.4  &\\
\midrule
\multirow{3}{*}{MIMA}    & $(2, 35)$    & 1NN  & 75.04 & 49.51 & 23.18 & 75.03 & 48.53 & 25.48 & 99.15 & 40.61 & 38.17 & 78.82 & 75.52 & 98.87 & 0.6496 & 60.66 & 69.42 & \multirow{3}{*}{72.78}\\
                         & $(1.5, 70)$  & LSVM & 73.73 & 53.64 & 43.85 & 78.22 & 60.39 & 26    & 99.36 & 40.34 & 49.06 & 83.69 & 74.48 & 98.31 & 0.7055 & 65.09 & 74.36 &\\
                         & $(1.5, 70)$  & KSVM & 70.83 & 47.09 & 44.34 & 80.09 & 76.87 & 22.88 & 99.65 & 35.19 & 45.57 & 84.96 & 54.69 & 90.09 & 0.7072 & 62.69 & 74.57 &\\
\bottomrule
\bottomrule
\end{tabular}}
\label{tb:LCZ}
\end{table*}

\subsubsection{MIMA vs. SSMA.} As shown in Fig.~\ref{fig:LCLUexp} and Table~\ref{tb:LCLU}, the proposed MIMA has superior performance to SSMA. In Fig.~\ref{fig:LCLUexp}, verified with three different classifiers, classifications on MIMA-fused data outperform classifications on SSMA-fused data, when parameter $\mu$ and the number of dimensions are the same for both fusion strategies. The classification performance of the best parameter combinations is shown in Table~\ref{tb:LCLU}. It is clear that the novel MIMA strategy still outperforms SSMA strategy, not only verifying the superior performance of the proposed novel MIMA algorithm, but also proving that a MAPPER-derived topological structure is more effective than a \textit{kNN}-derived structure regarding LCLU classification.

\subsubsection{Parameter $\mu$.} As shown in Fig.~\ref{fig:LCLUexp}, with \textit{ONE-NN} and \textit{KSVM} classifiers, a higher value of $\mu$ results in better classification performance for both SSMA and MIMA algorithms. Recalling that a higher value of $\mu$ assigns stronger weight on topological structure of data in the fusing phase, this is solid evidence that topological structure benefits our classification task. We also find that the way MIMA derives the structure is more beneficial to this LCLU classification than the way SSMA accomplishes it.

\subsubsection{Fusion visualization.} In Fig.~\ref{fig:LCLUVis}, we visualize the fused features of different algorithms using the t-SNE algorithm \cite{maaten2008visualizing}. It is obvious that the joint dimension reduction technique results a set of features which is less discriminative than the original feature. This is also reflected on the classification results, shown in Fig.~\ref{fig:LCLUexp}. On the other side, when using manifold alignment techniques, it is clear that the derived feature is more discriminative than the original ones.

\subsection{Classification on the LCZ data set}

\begin{figure*}[!t]
	\centering	
	{\small
    \begin{tabular}{ccccc}
      \includegraphics[width=0.18\textwidth]{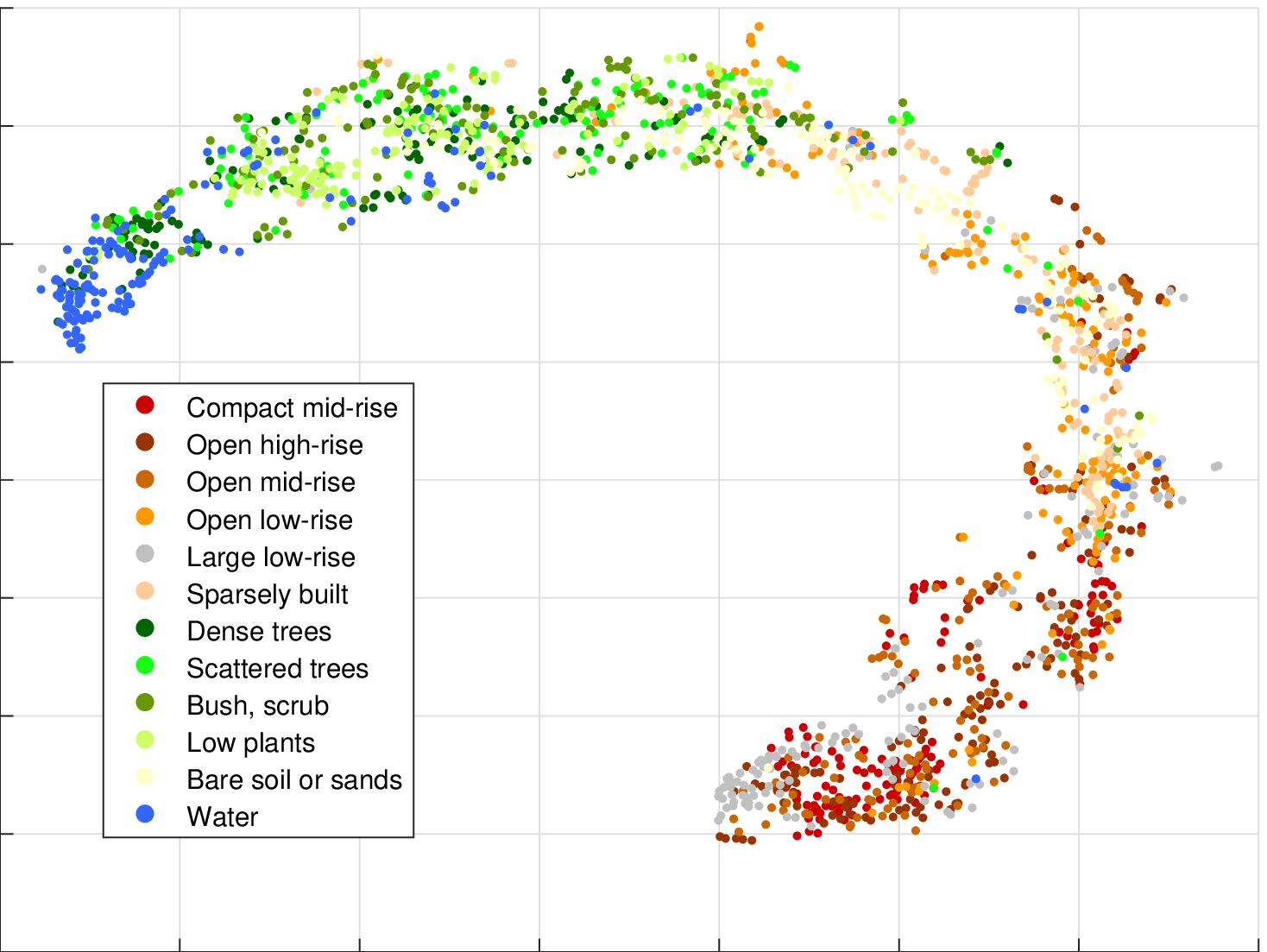}       &
      \includegraphics[width=0.18\textwidth]{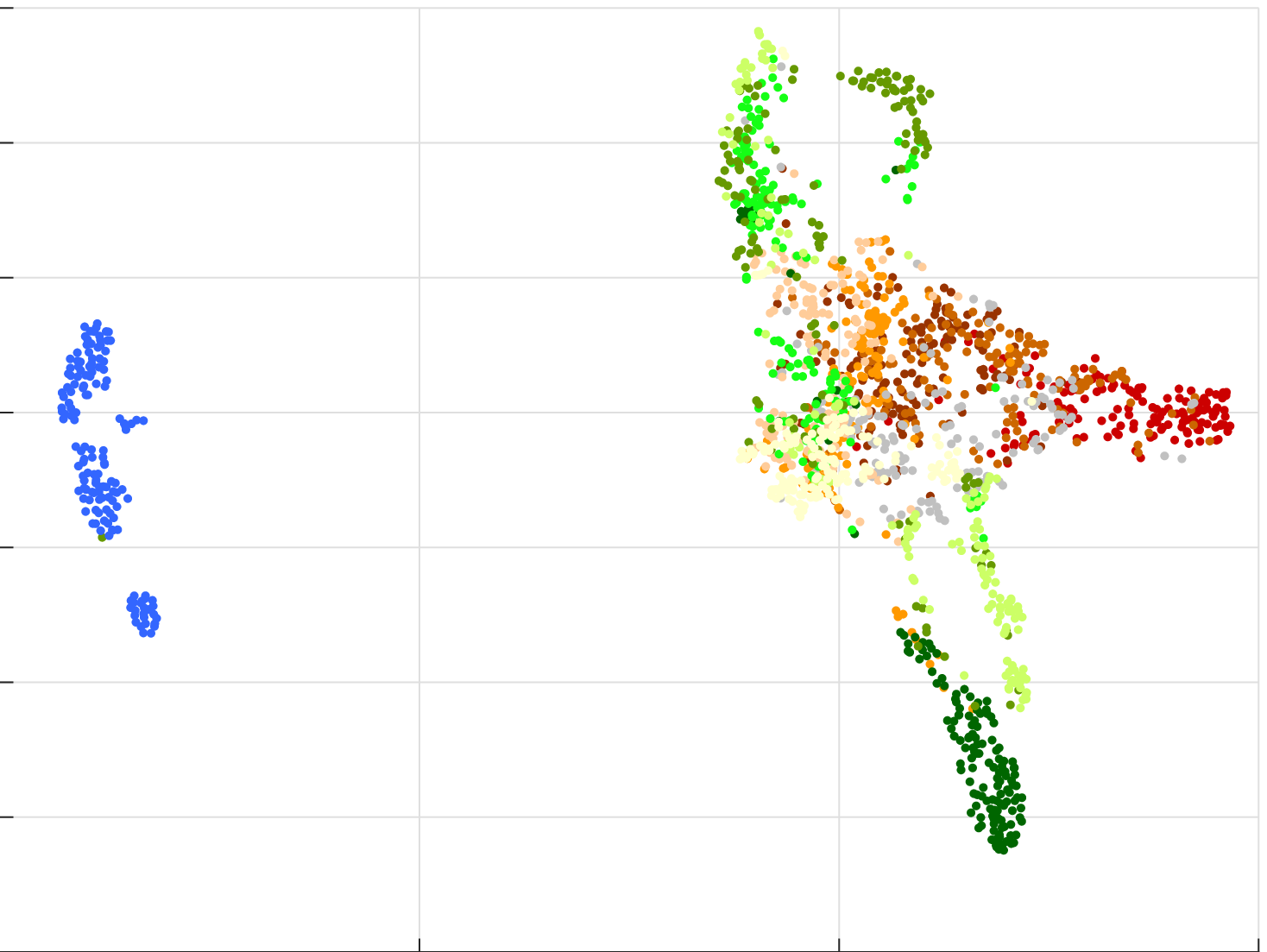}       &
      \includegraphics[width=0.18\textwidth]{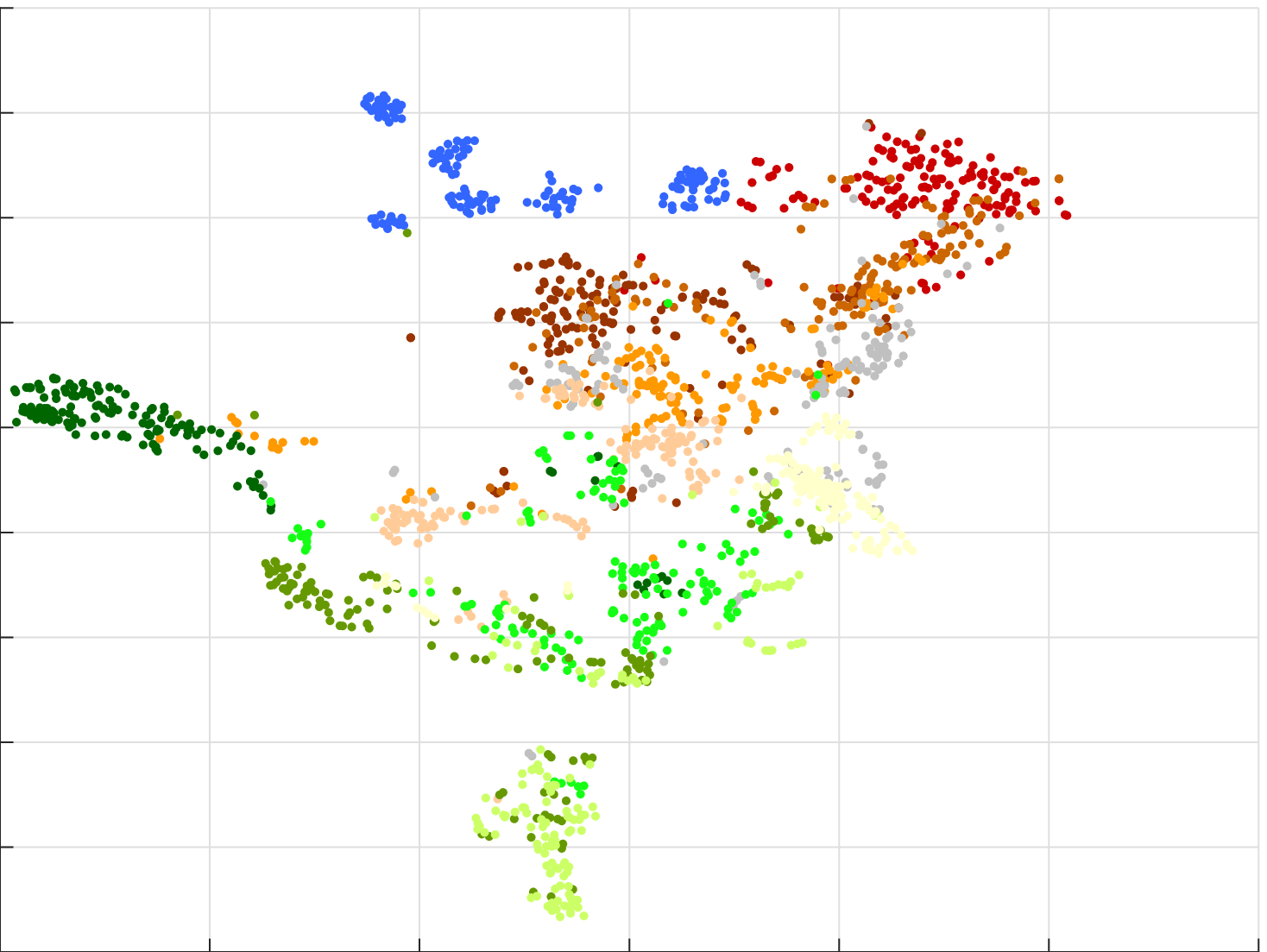}       &
      \includegraphics[width=0.18\textwidth]{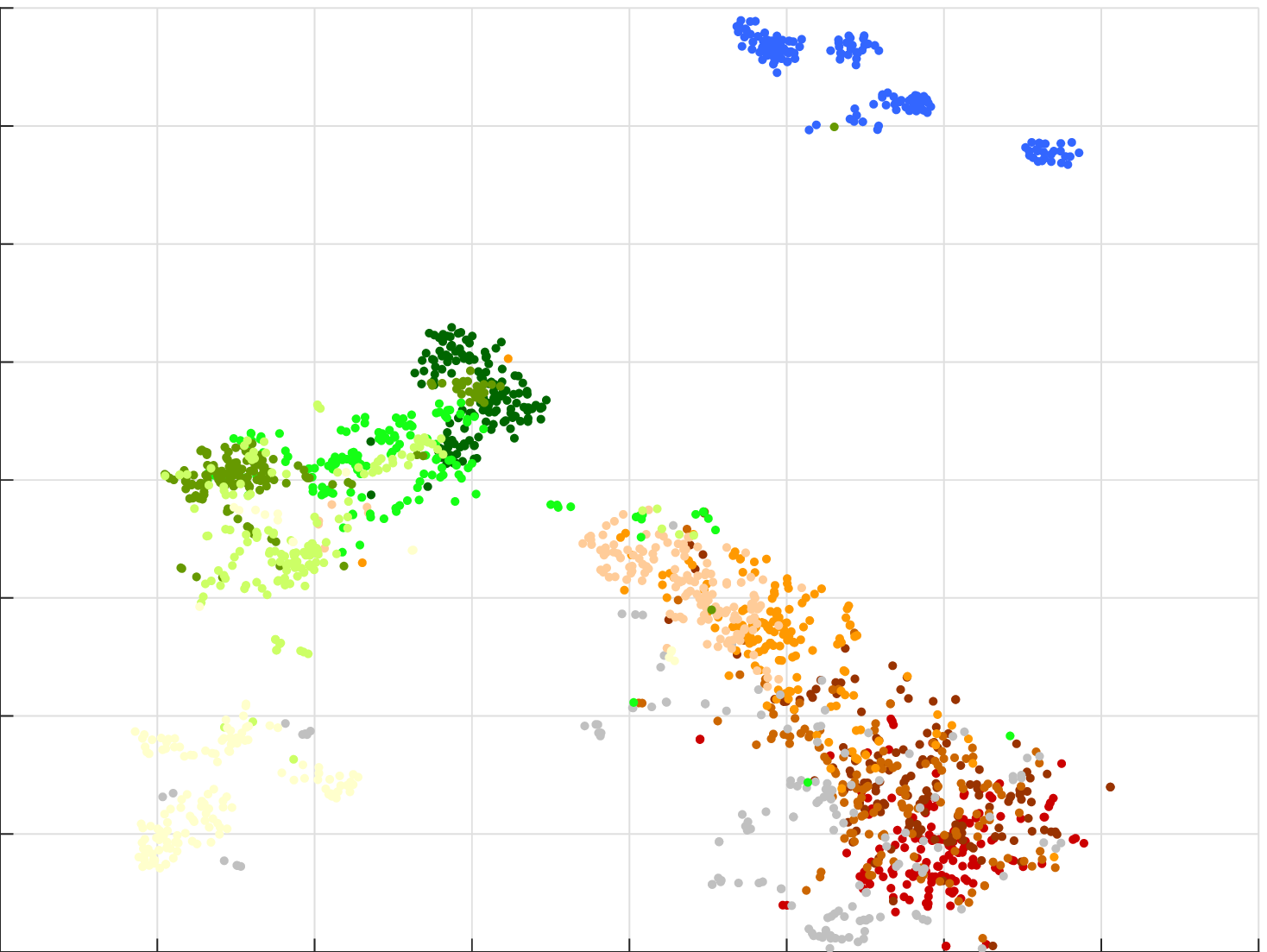}    &
      \includegraphics[width=0.18\textwidth]{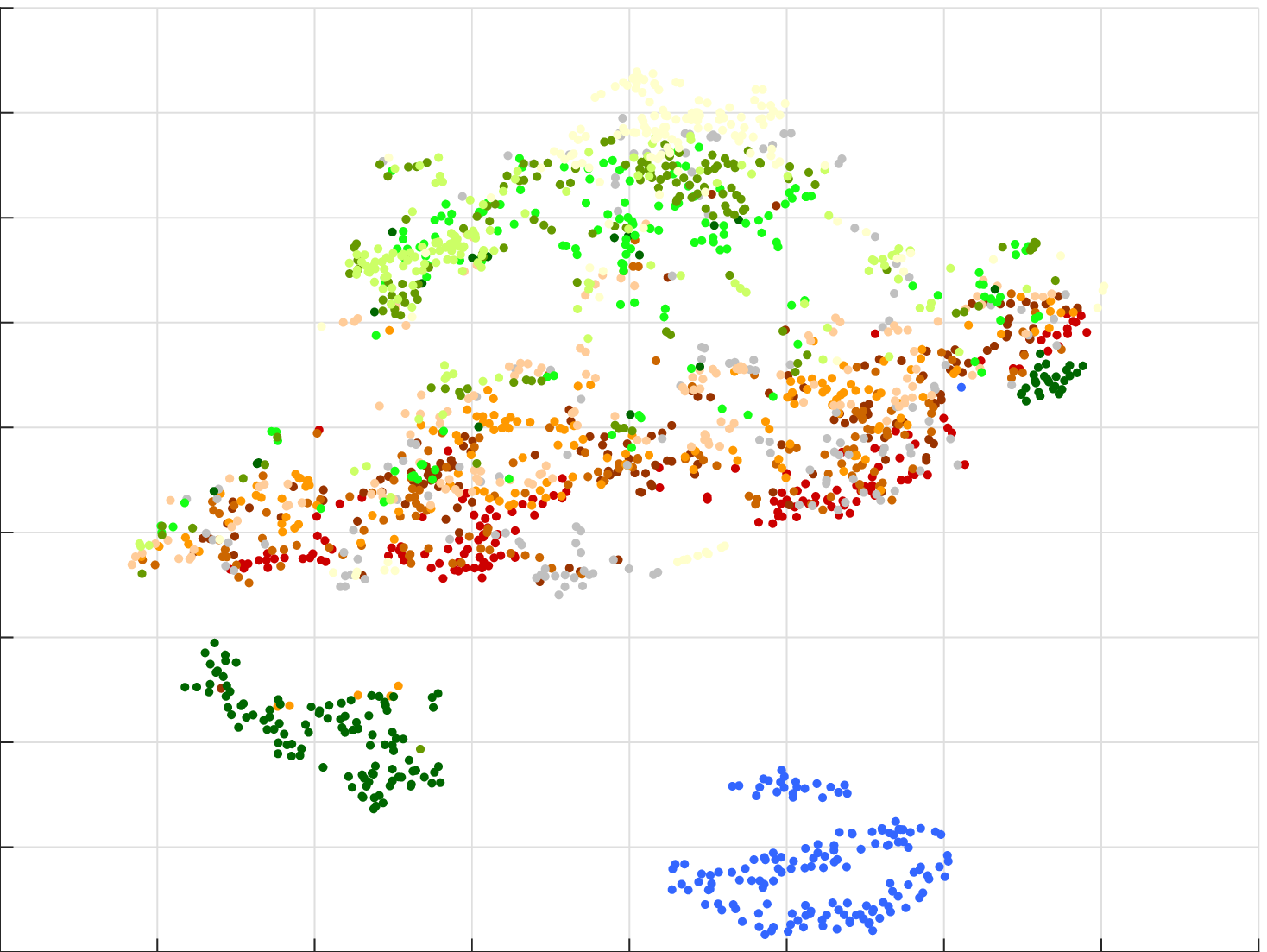}    \\

      PolSAR data space         &
      Optical data space        &
      LPP projected space       &
      LPP-SE projected space    &
      COSPACE projected space   \\
      
      \includegraphics[width=0.18\textwidth]{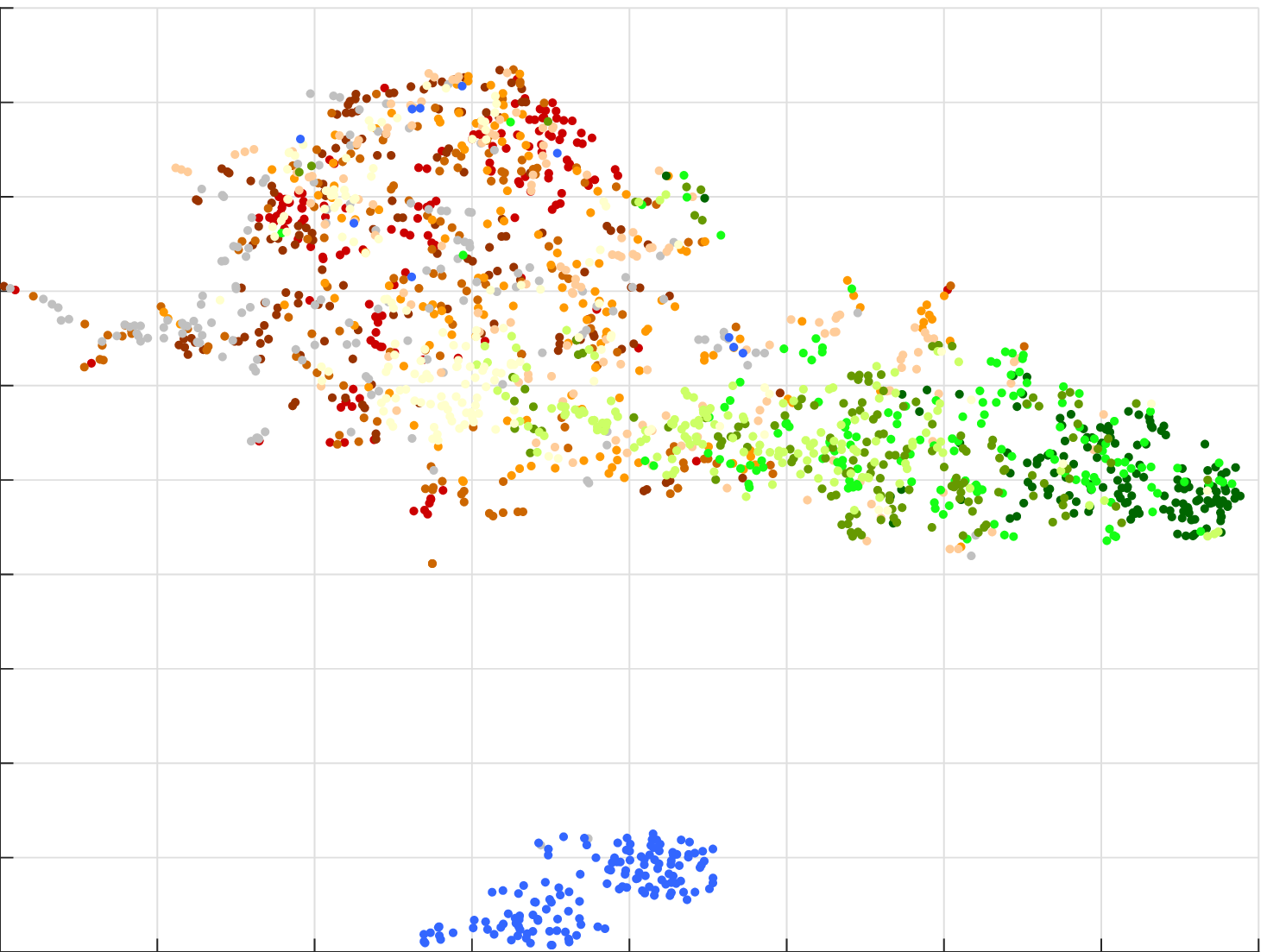}  &
      \includegraphics[width=0.18\textwidth]{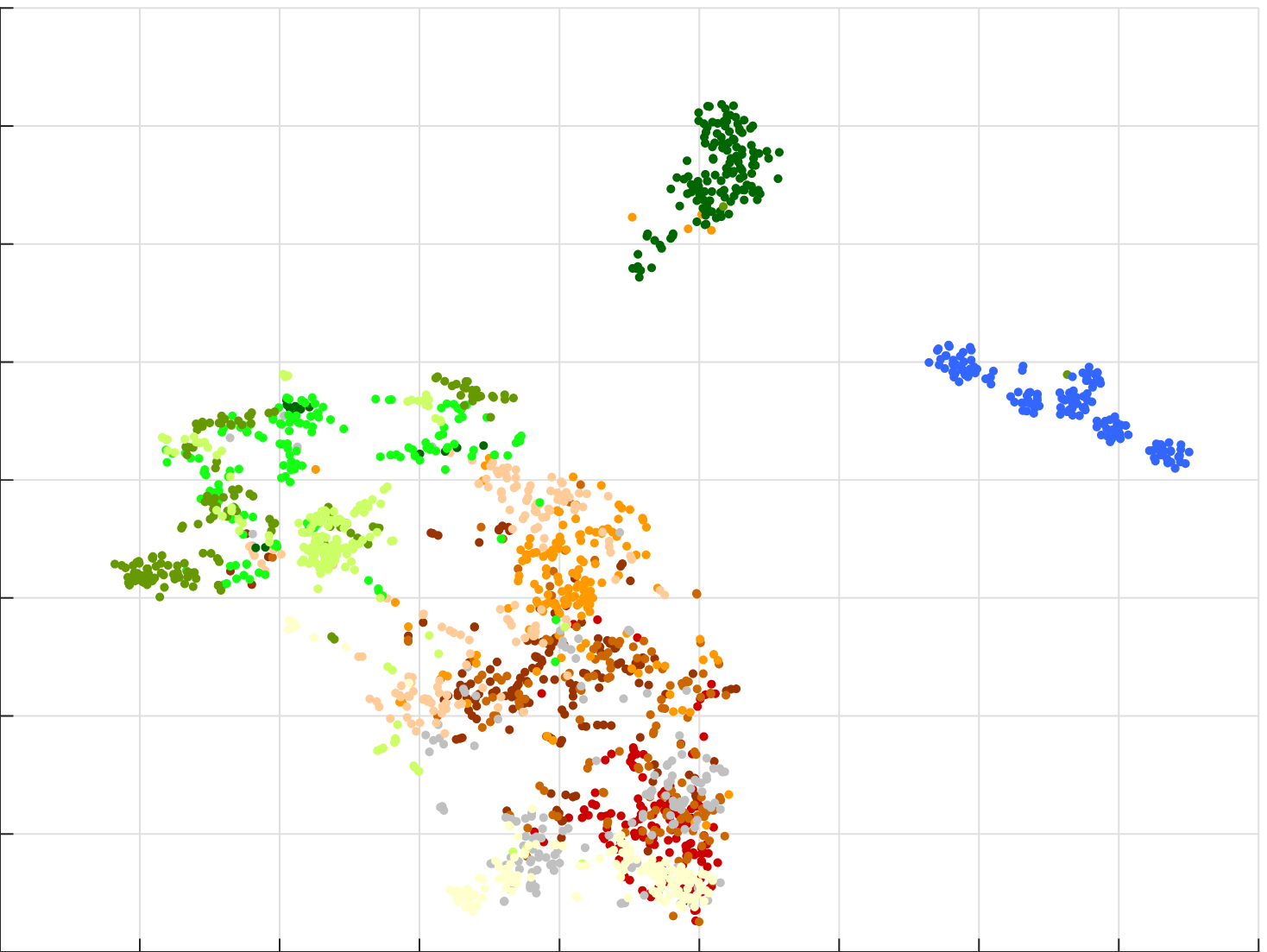}  &
      \includegraphics[width=0.18\textwidth]{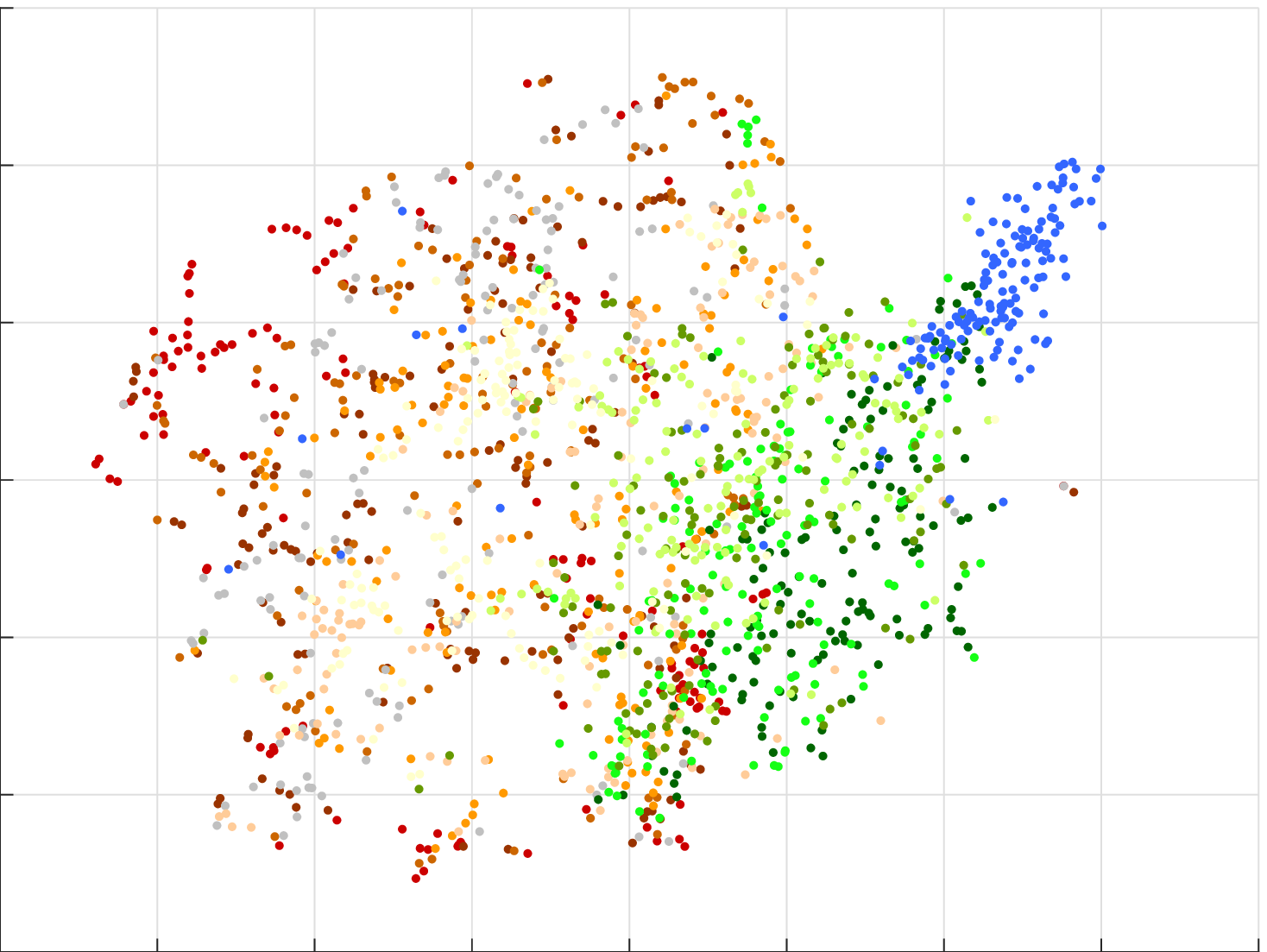}  &
      \includegraphics[width=0.18\textwidth]{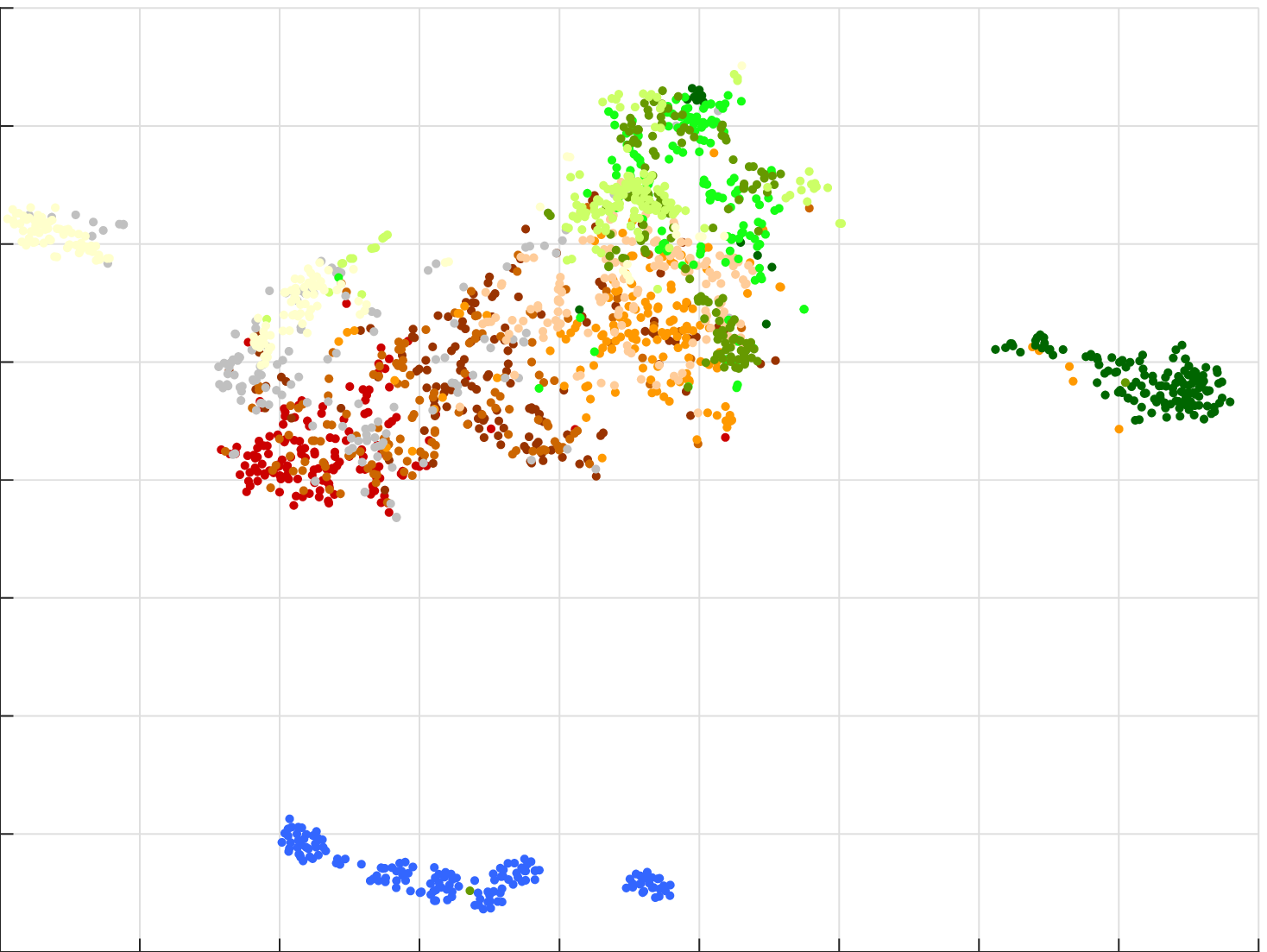}  &
      \includegraphics[width=0.18\textwidth]{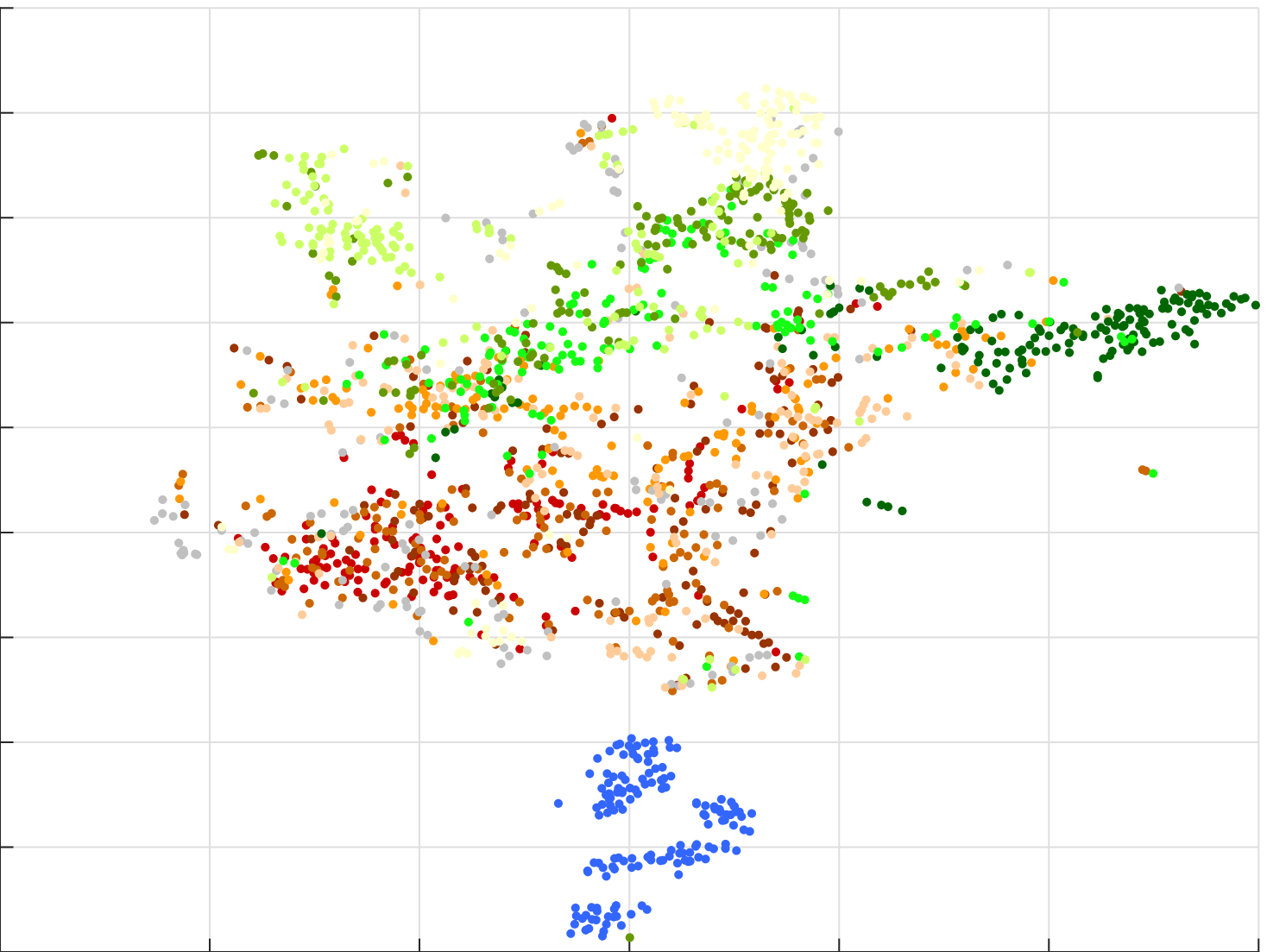}  \\

      SSMA projected & SSMA projected & MIMA projected & MIMA projected & \multirow{2}{*}{LeMA projected space} \\ 
      PolSAR data space & optical data space & PolSAR data space & optical data space &

    \end{tabular}  }      
    \caption{Visualization of the optical data and the PolSAR data of the LCZ data set, using t-SNE \cite{maaten2008visualizing} in their original and projected spaces. The x and y axis are the first and second dimensions resluted from the t-SNE. The first row are: the PolSAR data in the original space, the optical data in the original space, LPP jointly projected space, LPP-SE jointly projected space, and COSPACE projected space, respectively. The second row are: the PolSAR data in SSMA projected space, the optical data in SSMA projected space, the PolSAR data in MIMA projected space, the optical data in MIMA projected space, and LeMA projected space, respectively. }
	\label{fig:LCZVis}
\end{figure*}

This section demonstrates and discusses the experimental results obtained on the LCZ data set. 

\subsubsection{LeMA.}
The most outstanding phenomenon appears in Fig.~\ref{fig:LCZexp} and Table.~\ref{tb:LCZ} is that LeMA outperforms all the other algorithms by 2 to 6\%, which is considered a large margin in this experiment. Since LeMA and COSPACE both accomplish the fusion by using the labeled data and have similar performance in the experiment of LCLU data set, it is very interesting to find out the reason why LeMA not only outperforms COSPACE but also all the other fusion algorithm with a large margin. The difference between COSPACE and LeMA is that, while learning the projections from the labeled data, LeMA, additionally, includes pseudo-label into the learning phase. The pseudo-label are predictions of unlabeled data inferred by a trained classifier. LeMA has a strategy of selecting pseudo-label which have a high probability to be correctly labeled. In our classification evaluation, those correct-prone pseudo-label are also used for training classifiers of fused data. In the case of the experiment on the LCZ data set, by comparing to original 3170 training records and 18205 testing records, there are 1231 additional pseudo-label selected from the test data set which are used for training classifiers. It increases the training data set by 38.83\% and occupies 6.76\% of the testing data for validation. We believe the change in classification setting is the main reason that LeMA performs the best in the experiment of LCZ data set. On the other hand, in the experiment of the LCLU data set (3116 training records and 441778 testing records), LeMA has 721 additional pseudo-label, which increases the training data set by 23.14\% and occupies 0.16\% of the testing data for validation.

\subsubsection{Fusion.} 
According to Table.~\ref{tb:LCZ}, all fusion algorithms, except LeMA, have similar performance to the classification using only the multispectral imagery. Based on the 0.19\% difference between OPT-POL and OPT, we might infer that features extracted from dual-Pol SAR data do not benefit the LCZ classification scheme in terms of overall accuracy. However, the dual-Pol SAR data do benefit classifications of certain classes, namely, scattered trees, low plants, bare soil or sand, and water. By comparing COSPACE, LPP, LPP-SE, SSMA, and MIMA to OPT, mean overall accuracy improve -3.32\%, -1.1\%, -0.72\%, -0.59\%, and 0.49\%, respectively. This infers that few can be done by these fusion algorithms. However, among these fusion algorithms, our propose MIMA outperforms these four former algorithms by 3.81\%, 1.59\%, 1.21\%, and 1.08\%. 

\subsubsection{Data-driven manifold alignment.}
When the fusion is carried out by the {OPT-POL}, essentially, the information given by the label decides the classification boundary. However, in addition to the label, the data-driven manifold alignment involves the topological structures of the data to find the  classification boundary. As shown in Fig.~\ref{fig:LCZexp}, SSMA cannot compete with OPT-POL. This means that data structure is not beneficial with respect to LCZ classification. This is actually reasonable. The LCZ classification scheme describes the contents of an urban local neighborhood relating to the morphological structure, man-made or natural components, and height of structures. However, the topological structure derived from the remote sensing data reveals data structure in terms of its physical meanings, such as covering materials for multispectral images and geometric information for SAR data. Thus, the structure is not directly related to LCZ concepts. On this regard, the data-driven manifold alignment is more appropriate for the LCLU classification, since the information derived in the topological structure is directly related to LCLU classes. 

Despite the challenges that LCZ classes pose, when comparing to OPT-POL, the proposed MIMA slightly improves 1.02\% overall accuracy with the \textit{LSVM}. Comparing to LeMA, the performance of MIMA differ by -3.83\%, -0.79\%, and 0.23\% in terms of overall accuracy with three different classifier. We consider the performance are comparable. Only the -3.83\% indicates a big difference. However, this is because 38.83\% and 6.76\% differences of training and testing records have a huge impact on the classifier \textit{1NN}. With the two other classifiers, even with fewer training samples, the proposed MIMA is able to provide comparable classification accuracy.

\subsubsection{Parameter $\mu$.} 
According to Fig.~\ref{fig:LCZexp} and Table~\ref{tb:LCZ}, trends in terms of $\mu$ show that {SSMA} achieves its best performance when parameter $\mu$ equals one and performances are downgraded by increasing $\mu$ without a pattern. However, MIMA exhibits a pattern that classification accuracy increases as the value of parameter $\mu$ increases. This means that putting higher weights on the topological structure while fusing with MIMA would provide better classification performance in terms of OA. 

\subsubsection{Fusion visualization.} We visualize the fused features of different algorithms by t-SNE, as illustrated in Fig.~\ref{fig:LCZVis}. However, it is difficult to carry out a detailed analysis according to the visualization results. In general, the manifold alignment based fusion provides spaces where classes concentrate well. To our knowledge, the optical data are projected into a more discriminative subspace via the proposed MIMA.

\begin{table}[h]
\caption{The computational cost of algorithms in comparison. The time listed in this table are means of ten repetitions on each algorithm, carried out on the LCZ data set. The unit is reported in second}
\centering
\label{tb:computation}
\resizebox{\linewidth}{!}{
\begin{tabular}{c|cccccc}
\toprule
\toprule
Fusion algorithm & (\textit{D}) COSPACE & (\textit{E}) LeMA & (\textit{F}) LPP & (\textit{G}) LPP-SE & (\textit{H}) SSMA & (\textit{I}) MIMA \\
\midrule
Time (sec) & 324.3175 & 1080.3080 & 4.6141 & 6.0033 & 45.2369 & 129.3108 \\
 \bottomrule
 \bottomrule  
\end{tabular}}
\end{table}

\begin{figure*}
\centering	
    \begin{tabular}{cc}
      \includegraphics[width=0.3\textwidth]{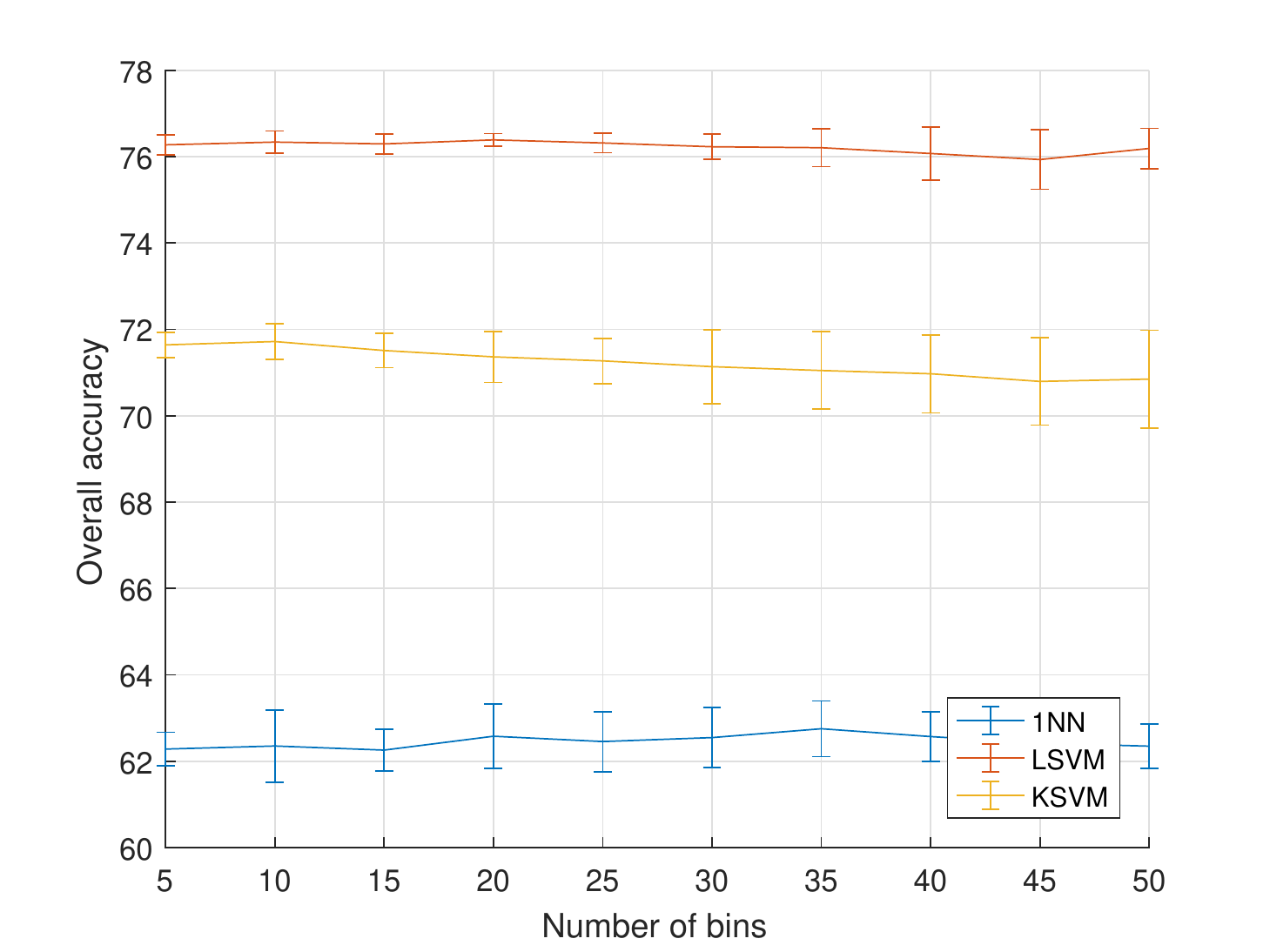}  &
      \includegraphics[width=0.3\textwidth]{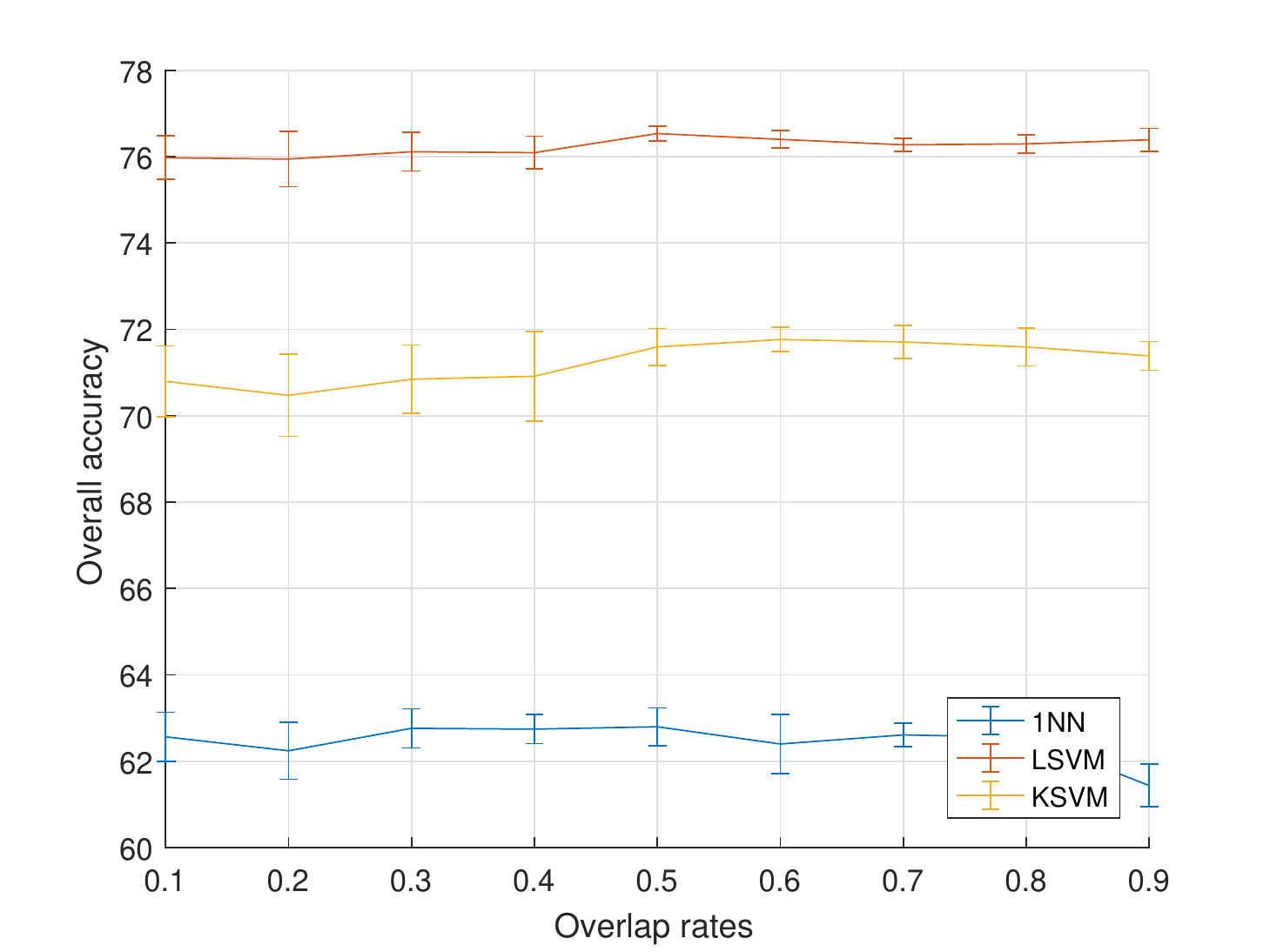}  \\
      \includegraphics[width=0.3\textwidth]{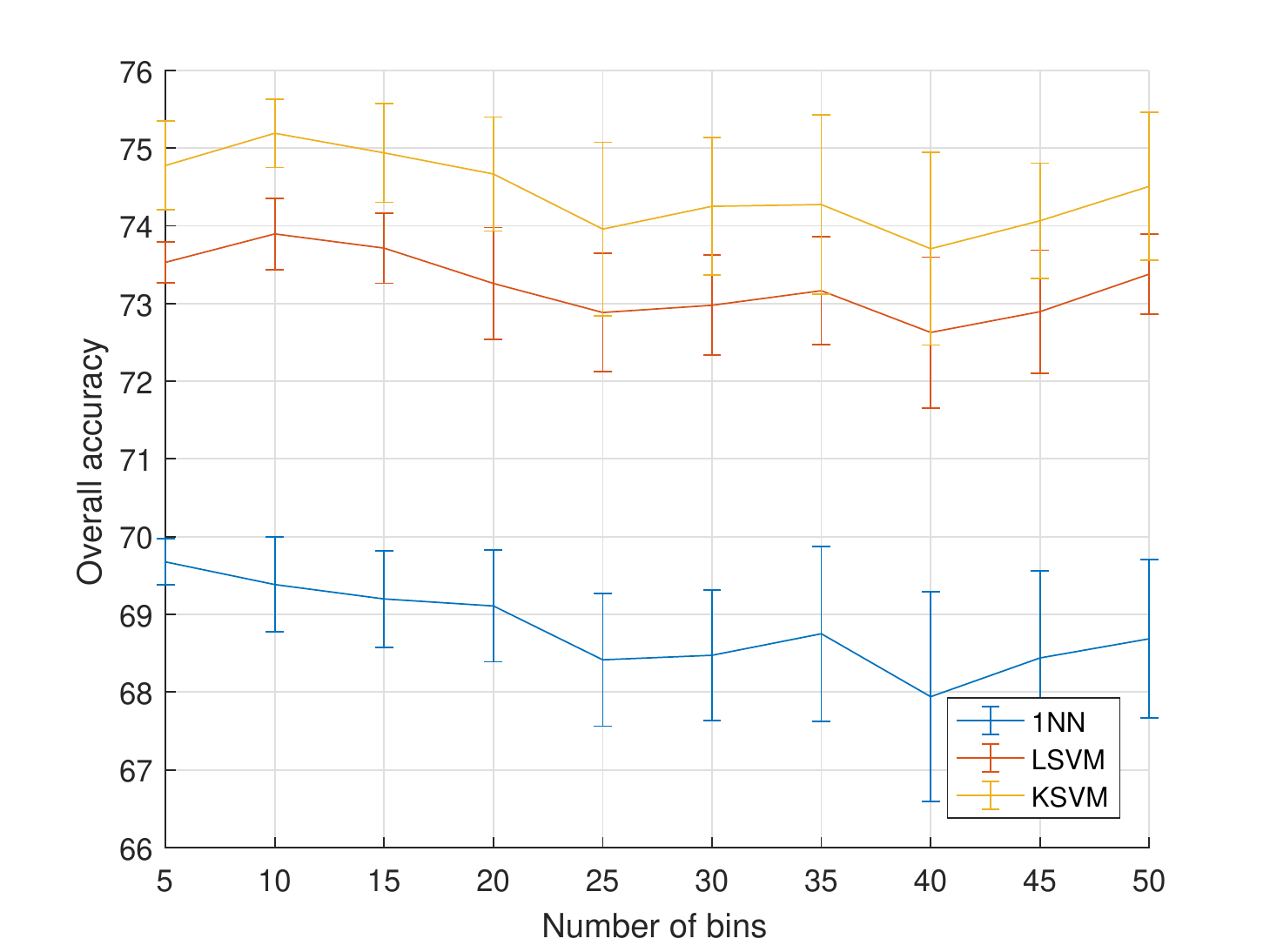}  &
      \includegraphics[width=0.3\textwidth]{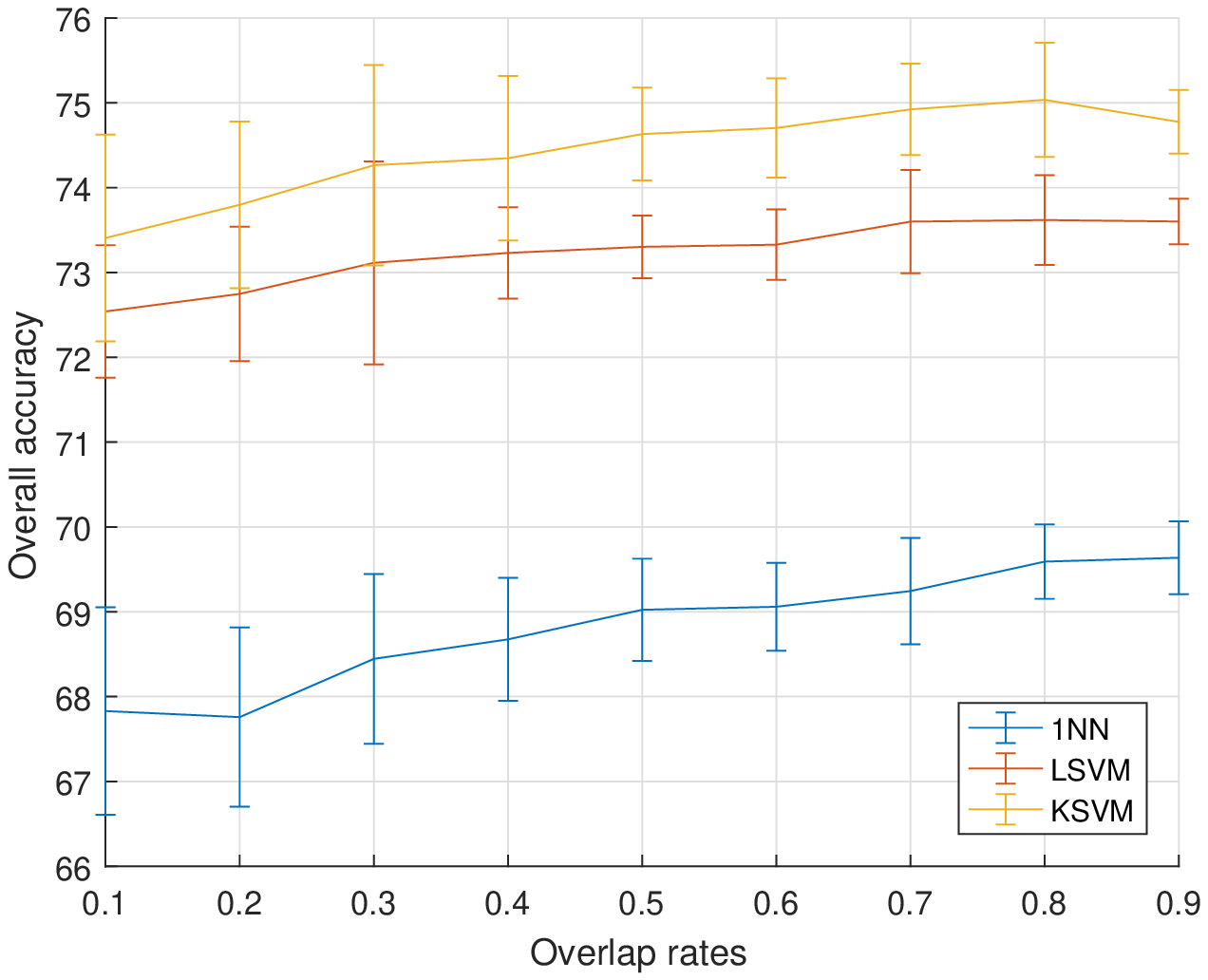}  
    \end{tabular}        
    \caption{Plots of the classification overall accuracies achieved by applying the classifiers: the \textit{ONE-NN}, the \textit{LSVM}, and the \textit{KSVM} on MIMA fused feature, while only values of two parameters varies, the number of data bins and the overlap rate. From left to right: (1) plot of overall accuracies achieved on the LCLU data set; the curve and the error bar represent the mean and the standard deviation, which achieved statistically with varying overlap rates; (2) plot of overall accuracies achieved on the LCLU data set; the curve and the error bar represent the mean and the standard deviation, which achieved statistically with varying number of bins; (3) plot of overall accuracies achieved on the LCZ data set; the curve and the error bar represent the mean and the standard deviation, which achieved statistically with varying overlap rates; (4) plot of overall accuracies achieved on the LCZ data set; the curve and the error bar represent the mean and the standard deviation, which achieved statistically with varying number of bins;} 
	\label{fig:BIN_OVER}
\end{figure*}
\subsection{Computational cost}
In order to show the computational efficiency of algorithms in comparison, experiments of ten repetitions over the LCZ data set had been carried out for every algorithm in comparison. All the experiments are accomplished on a desktop with a processor of Intel Core i7-4790 CPU (3.60 GHz). The mean time of these repetitions are listed in Table.~\ref{tb:computation}.

Our proposed MIMA do suffer from comparably high computational cost, as shown in Table~\ref{tb:computation}. This is due to the high computational cost of the spectral clustering \cite{ng2002spectral} in MIMA. If the algorithm efficiency is of key importance for a targeted application, more studies could be carried out to find a less demanding clustering algorithm as a substitute. However, considering 9170 optical pixels with 34 dimensions and 9170 SAR pixels with 40 dimension are involved in the training of the algorithm, we think two minutes which required by MIMA is still acceptable.

\subsection{Data bins and overlap rates}
\label{sec:MAPPERpara}

As described in the section~\ref{sec:MIMAtheory}, there are two parameters brought to MIMA by the MAPPER. They are: the number of data bins and the overlap rate of adjacent data bins. Among all experiments in the previous sections, the number of data bins is chosen as 5 and the overlap rate is selected as 50\%, based on the experience of medical studies \cite{nicolau2011topology,li2015identification}. However, in this section, the impact of those two parameters are discussed, in terms of the remote sensing data.

Theoretically, the number of data bins has a similar effect to the value $k$ of the \textit{kNN}, which controls the extent of a local neighborhood. Because the local topological structure is derived by the clustering in smaller slices of the data, when a larger number of bins is applied. On the other hand, the overlap rate controls the strength of the connection between adjacent local neighborhoods. Although the theoretical concept is clear, their impacts are really depending on the data set which it works with. 

Fig.~\ref{fig:BIN_OVER} demonstrate the impact of the number of data bins and the overlap rate by using the LCLU data set and the LCZ data set, in terms of classification accuracy. The number of bins is set to values from 5 to 50 with an interval of 5. The overlap rate is set to values from 0.1 to 0.9 with an interval of 0.1. For the sake of simplicity, the parameter $\mu$ is set to 2 and the latent space dimension is set to 50, for the analysis in this section.

According to the upper two plots in Fig.~\ref{fig:BIN_OVER}, regarding the LCLU classification, the number of bins and the overlap rate do not have a significant influence in terms of the overall accuracy. However, based on the bottom two plots in Fig.~\ref{fig:BIN_OVER}, regarding the LCZ classification, one can recommend a large overlap rate around 90\% and the number of data bins around 10. Thus, the decision of both parameters really depends on the data set and the targeted classification scheme. Last but not the least, it also relates to the choices of the filtering function. One more interesting point is that, by comparing the fluctuation of curves in Fig.~\ref{fig:LCLUexp}, Fig.~\ref{fig:LCZexp}, and Fig.~\ref{fig:BIN_OVER}, we can observe that impacts of $\mu$ and $dn$ are much larger than impacts of the number of bins and the overlap rate.

\section{Conclusion}
In this paper, we propose a MAPPER-induced manifold alignment for semi-supervised fusion of optical data and polarimetric SAR data, inspired by the semi-supervised technique and the emerging field of topological data analysis. Specifically, we embed a successful topological data analysis tool, MAPPER, into SSMA, to accomplish heterogeneous data fusion. Furthermore, our modified version of MAPPER functions adaptively to data by improving clustering. The performance of MIMA on fusing optical data and polarimetric SAR data is superior to that of SSMA, LPP, COSPACE, LeMA, and the feature concatenation, with respect to LCLU classification and LCZ classification. SSMA-based method is applied to fuse optical data and SAR data for the first time. This is also the first time that topological data analysis is applied in remote sensing field. 

In the future, further experiments will be conducted to explore the potential of the proposed MIMA by selectively introducing field knowledge of remote sensing data. In this manner, physical meanings of different remote sensing data can be explicitly introduced into data fusion, instead of treating it as a data-driven machine learning problem. We believe an expert knowledge driven MIMA can further improve the fusion performance.


%



\section*{Acknowledgment}

The authors would like to thank the following institutions for providing data sets used in this study: German Research Center for Geosciences, IEEE GRSS IADTC and OpenStreetMap.

\ifCLASSOPTIONcaptionsoff
  \newpage
\fi



\bibliographystyle{IEEEtran}
\bibliography{IEEEabrv,ref}
\end{document}